\documentclass[Afour,sageh,times]{sagej}       
\usepackage{graphicx}                       
\usepackage{graphics}                       
\usepackage{epsfig}                         
\usepackage[tight,footnotesize]{subfigure}  
\usepackage{amssymb,amsmath}
\usepackage{mdwmath}
\usepackage{commath}      	
\usepackage{mathtools}
\usepackage{eqparbox}
\usepackage{bm} 			
\usepackage{amsmath,amsfonts,amssymb}

\newcommand{\ie}{{\em i.e.\ }}

\newcommand{\eg}{{\em e.g.\ }}

\newcommand{\et}{{\em et al.\ }}
\newcommand{\ets}{{\em et al.'s\ }}

\newcommand{\gap}{\mbox{ }}

\newcommand{\BC}{\mbox{$\mathcal B$}}

\newcommand{\FC}{\mbox{$\mathcal F$}}
\newcommand{\GC}{\mbox{$\mathcal G$}}

\newcommand{\MC}{\mbox{$\mathcal M$}}
\newcommand{\NC}{\mbox{$\mathcal N$}}

\newcommand{\RC}{\mbox{$\mathcal R$}}
\newcommand{\SC}{\mbox{$\mathcal S$}}
\newcommand{\TC}{\mbox{$\mathcal T$}}

\newcommand{\VC}{\mbox{$\mathcal V$}}

\newcommand{\XC}{\mbox{$\mathcal X$}}

\newcommand{\beq}{\begin{equation}}
\newcommand{\eeq}{\end{equation}}
\newcommand{\bear}{\begin{eqnarray}}
\newcommand{\bears}{\begin{eqnarray*}}
\newcommand{\eear}{\end{eqnarray}}
\newcommand{\eears}{\end{eqnarray*}}
\newcommand{\bdm}{\begin{displaymath}}
\newcommand{\edm}{\end{displaymath}}
\newcommand{\lba}{\left[\begin{array}}
\newcommand{\ear}{\end{array}\right]}

\usepackage{stfloats}                       
\usepackage{hyperref}
\usepackage[T1]{fontenc} 
\usepackage{enumitem} 
\usepackage[ruled]{algorithm2e}
\usepackage{tabularx}
\usepackage[table]{xcolor}
\usepackage{longtable}
\usepackage{booktabs}       					
\usepackage{multirow}
\usepackage{xcolor,colortbl}               		
\usepackage{array}								

\usepackage{caption} 
\usepackage[title]{appendix}
\newcommand\given[1][]{\:#1\vert\:}
\usepackage{moreverb,url}
\newcommand\BibTeX{{\rmfamily B\kern-.05em \textsc{i\kern-.025em b}\kern-.08em
T\kern-.1667em\lower.7ex\hbox{E}\kern-.125emX}}
\def\volumeyear{2018} 
\begin{document}
\runninghead{Recovering from External Disturbances}
\title{Endowing Robots with Longer-term Autonomy by Recovering from External Disturbances in Manipulation through Grounded Anomaly Classification and Recovery Policies.}
\author{Hongmin Wu*, Shuangqi Luo*, Longxin Chen, Shuangda Duan,\\ Sakmongkon Chumkamon, Dong Liu, Yisheng Guan, and Juan Rojas.
}
\affiliation{*Hongmin Wu and Shuangqi Luo contributed to the present work in equal parts and share first authorship. All authors are with the School of Electromechanical Engineering in Guangdong University of Technology in Guangzhou, China.}
\corrauth{Juan Rojas, Guangdong University of Technology,
Biomimetics and Intelligent Robotics Laboratory,
HEMC, Panyu, GDUT
Guangzhou, Guangdong, 
510006, China.}
\email{juan.rojas@gdut.edu.cn}
\begin{abstract}
Robot manipulation is increasingly poised to interact with humans in co-shared workspaces. Despite increasingly robust manipulation and control algorithms, failure modes continue to exist whenever models do not capture the dynamics of the unstructured environment. To obtain longer-term horizons in robot automation, robots must develop introspection and recovery abilities. 
We contribute a set of recovery policies to deal with anomalies produced by external disturbances as well as anomaly classification through the use of non-parametric statistics with memoized variational inference with scalable adaptation. A recovery critic stands atop of a tightly-integrated, graph-based online motion-generation and introspection system that resolves a wide range of anomalous situations. Policies, skills, and introspection models are learned incrementally and contextually in a task. 
Two task-level recovery policies: re-enactment and adaptation resolve accidental and persistent anomalies respectively. Re-enactment policies model human decision making to re-enact the best skill in the task-graph. Adaptive recoveries leverage human intuition about the task-state to overcome persistent errors. The system is capable of fast and robust anomaly identification and classification during all phases of a task including during the execution of newly learned recovery skills.
The introspection system uses non-parametric priors along with Markov jump linear systems and memoized variational inference with scalable adaptation to learn a model from the data in an incremental way and yield compact interpretable models that enhance classification and identification accuracy. 
Extensive real-robot experimentation with various strenuous anomalous conditions in a co-bot scenario is induced and resolved at different phases of a task and in different combinations. The system executes around-the-clock introspection and recovery and even elicited self-recovery when misclassifications occurred. 
\end{abstract}
\keywords{Recovery policies, long-term automation, anomaly classification, anomaly identification, manipulation}
\maketitle
\section{Introduction}\label{sec:Intro}
\begin{figure}[t]    
	\centering		
		\subfigure{\includegraphics[width=3.25in]{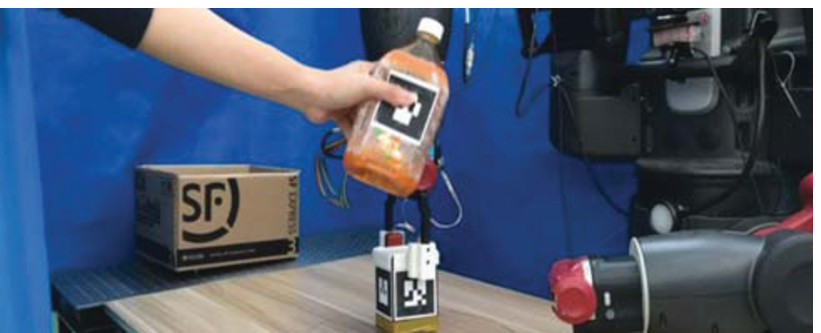}}
\vspace{0.01cm}
\subfigure{\includegraphics[scale=1]{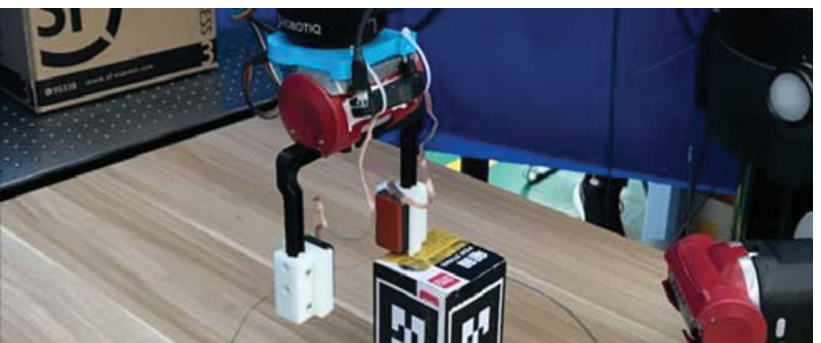}}
		\caption{Anomaly examples when a robot performs a kitting experiment in a shared workspace with a human. Top: an accidental collision between the robot and the human as the human places a new object in the collection area. Bottom: a tool collision as the end-effector prepares to pick an object.}
        \label{fig:external_anomaly_example}        
\end{figure}

As robots enter increased levels of unstructured environments and shared workspaces with humans, unexpected anomalies are anticipated. Even as manipulation and control algorithms become increasingly robust, failure modes continue to exist. Numerous sources of error and possible execution anomalies arise from the complex dynamics found in robots, their interactions with the world and human collaborators, as well as robot's limitations to model the world. Internal errors---resulting from improper modeling of visual, kinematic, or dynamic models ---and limited hardware accuracy can potentially lead to anomalies. It is the external anomalies, however, that are the hardest to account for in unstructured environments. External anomalies may arise from the inability to model sudden accidental collisions (human-robot, robot-world, or robot-object-world), object slips due to inertial dynamics, misgrasps; or even a chain reaction were one anomaly generates other anomalies. Fig. \ref{fig:external_anomaly_example} illustrates two anomaly examples in a kitting experiment. Furthermore, anomalous conditions are hard to model as similar anomalies can occur with wide variability, making it challenging for robots to recognize. In this work, we refer to self-monitoring as (physical) introspection. It includes the ability of a robot to recognize both the nominal and anomalous state conditions it may be in.
Therefore, it is imperative to have a recovery framework that leverages introspection results and incrementally learns to resolve new anomalous situations in unstructured environments.

This work implements a recovery framework to allow the explicit encoding of contextual recovery policies in online collaborative manipulation tasks.
Few papers have studied the development of explicit recovery policies for recovery of anomalous conditions, especially those that are characterized by random or unstructured qualities that are hard to model or anticipate.
Recovery action designs in robots shares similarities with motion generation and skill sequencing notions in robot manipulation 
\citep*{2010RAM-Calinon-LrnReprodGesturesImitation, 2013NC-Ijspeertdynamical-CMP_LrnAttracMdls_MtrBeh, 2013NIPS-Paraschos_Peters-ProMPs, 2016IJML-Abeel-End2EndTraining, 2009JAIR-chernova-InterPolicyLrn_ConfidenceBasedAutonomy, 2010IROS-Grollman_Jenkins-IncrmntlLrn_Subtasks_UnsegDemos, 2012IJRR-Konidaris-LfD_SkillTrees, 2015IJRR-Niekum-LrnGrnddFiniteStateReprUnstrucDems,2018ICRA-Gutierrez-IncrementalTaskMod, 2016ISER-Levine-LearnHandEyeCalib_BotGrasping_DL}.
However, in anomaly recovery, a robot may require to re-attempt a specific skill, whilst at other times it may need to apply a new skill to adapt to new world conditions. It is important to learn recovery behaviors incrementally in response to \textit{specific} anomaly events that may occur unpredictably in a task. For recovery to be useful, a \textit{recovery critic} must resolve the best policy to enact for a given anomaly at a given time.

Anomalies (the deviation of sensor-related signatures from those experienced in nominal executions) have been studied particularly in structured and uni-modal formats \citep*{1998IJRR-Hovland-HMM_ProcessMonitorAsmbly, 2005BotAut-Petterson-ExecutionMonitoringInRoboticsSurvey}. More recently, anomaly identification \citep*{2011ICRA-Stolt-ForceCtrldAsmblyStpBttn,2013IJMA-Rojas-TwrdsSnapSensing,2014ICRA-Rojas-EarlyFC} and classification \citep*{2013IROS-DiLello-BayesianContFaultDetection,2017humanoids-rojas-shdp-var-hmm,2016ICRA-Park-MultiModalMonitoringAnomalyDet_RobotManip,2017IROS-Park-MultiModalAnomalyClassifFeeding,2018AutBot-Park-MultimodAnomDet_AssistiveBots} in unstructured environments have been the subject of growing interest as the need for robots to work in unstructured is greater and more feasible. In \citep*{2013IROS-DiLello-BayesianContFaultDetection}, a non-parametric Bayesian prior was used with Hidden Markov Models (HMMs) and a Gaussian observation model and Gibbs sampling to do anomaly classification of four static anomaly situations. In \citep*{2018ROMAN-Luo-RobustVersatileEventDet}, Markov Jump Linear Systems were used to model latent states through linear dynamical process (the vector auto regressive) for anomaly identification in a pick-and-place task as well as a drawer opening task. A novel threshold that varies according to the execution of the process was designed through the use of gradient of belief state for the HMM gradient but no anomaly classification was conducted. In \citep*{2016ICRA-Park-MultiModalMonitoringAnomalyDet_RobotManip}, anomaly identification was conducted by using a traditional HMM but with a detection threshold that varied according to clusters of execution progress. The same work was improved in \citep*{2018AutBot-Park-MultimodAnomDet_AssistiveBots} and instead introduced two approaches to computed likelihoods as a function of progress. First, they used Gaussian radial basis functions to produce the clusters and associated likelihoods that gauge execution progress and vary the identification threshold. Second, they sought a method to eliminate discontinuities between clusters and opted to use a Gaussian-process regressor to compute the mean and standard deviation of the log-likelihood segments. While this work focused on detecting anomalies caused by a wide variety of sources, it did not implement a multi-class anomaly classifier. Finally in \citep*{2017IROS-Park-MultiModalAnomalyClassifFeeding}, an artificial neural network is used to identify and classify anomalies in the context of robot-assisted feeding and producing competitive results. The network uses input features extracted from an HMM, raw sensory signals, and a convolutional neural network. In our work, we look to improve the performance of the identification and classification of anomalies by using the \textit{sticky} Hierarchical Dirichlet Process (sHDP)-HMMs with memoized variational inference and scalable adaptation to learn more compact and interpretable models of VAR process to enhance the anomaly identification and classification accuracy (Sec. \ref{sec:robot_introspection}). 
Few works design recovery policies that explicitly handle the occurrence of various anomalies at different times in a task. For example, in \citep*{2011IROS-Rodriguez-AbortRetry,2013IROS-Nakamura-ErrorRecTaskStrat,2018ROMAN-Wu-RecovExtDist_StateDep}, the entire task is only re-attempted upon failure. In \citep*{2013ICAR-Chang-BotTaskRecovery_PetriNets_LfD}, Chang \et devised an error recovery system based on Petri Nets learned from demonstration. Error conditions however were defined based on object location: if objects were not located in expected states, an error was triggered. This forced the system to maintain a growing list of expected object locations. The work did not consider other anomaly sources. In \citep*{2015RSS-Kappler-DateDrivenOnlineDecisionMakingManipu}, failure classification was performed for only one perturbation and it was pre-taught. No failure identification was presented nor was there an explicit recovery policy. Instead, a recovery behavior was inserted manually in a specific place in the task and no explicit experimental results quantified recovery versatility and robustness. Likewise the ability to grow recovery behaviors incrementally over time was absent. In \citep*{2015IJRR-Niekum-LrnGrnddFiniteStateReprUnstrucDems}, a system that allows for the incremental addition of skills is taught, but there is no mentions on how anomalies could be explicitly classified. Adaptive behavior was taught for two anomalies that occurred predictably and were characterized by a consistent structure. No explicit recovery policy was presented in this work to handle anomalies. 

The research question that we studied in this work is: ``Given the ability to classify anomalies, to what extent can we extend long-term autonomy when using a simple set of task-level recovery policies that grow over time''. Our work contributes the design of an explicit manipulation anomaly recovery system that is characterized by six attributes: 
\begin{enumerate}
  \item The use of Bayesian non-parametric Hidden Markov Models with memoized variational inference with scalable adaptation for robust online anomaly classification trained with few data. 
  \item The learning of contextual recovery policies that provide unique recovery policies for anomalies at specific locations in the task graph. 
  \item  The establishment of two types of recovery policies that differentiate between anomalies whose causes are accidental (one-off) occurrences and anomalies that are persistent. 
  \item The ability to introspect and perform recoveries reliably whilst the robot already executes a recovery behavior triggered by a previous anomalous conditions.
  \item The integration of a system that combines scalable and encapsulated motion generation, introspection, and recovery. 
  \item The inclusion of an (costly-to-generate) anomaly dataset in a co-bot Kitting experiment that includes relevant multimodal sensory-motor information and RGB video for a wide range of anomalous conditions and recoveries at different parts of the task (details in Extension 2). 
\end{enumerate}
Our contribution builds on top of our previously developed introspection system which could introspect into nominal skills and identify anomalies, but not classify them or recover contextually from them. In this paper, we significantly expanded introspection to deal with the online classification of a challenging set of anomalies and further implemented contextual recovery policies to resolve them. Extensive experimentation showed not only that we can perform globally anomaly classification and contextual recoveries effectively but also that the system self-recovers from erroneous classification and successfully recovers from existing anomalies. 

Manipulation tasks are represented through a hierarchical graph-based representation. Nodes consist of modules that encode skill generation, skill introspection, and skill goal setting. Modules run in parallel allowing for a constant sense-plan-act-introspect (SPAI) paradigm. 
Skill modules are flexible and can execute a users preferred skill generation technique (i.e. HMM motion generation \citep*{2010RAM-Calinon-LrnReprodGesturesImitation}, Dynamic Motion Primitives (DMPs) \citep*{2013NC-Ijspeertdynamical-CMP_LrnAttracMdls_MtrBeh}, Probabilistic Motion Primitives (ProMPs) \citep*{2013NIPS-Paraschos_Peters-ProMPs}, Interaction Probabilistic Movement Primitives (IProMPs) \citep*{2017Humanoids-Longxin-HRC}, or deep reinforcement learning (RL) techniques \citep*{2016IJML-Abeel-End2EndTraining}.
Introspection modules run anomaly identification and classification through Bayesian non-parametric Markov Jump Linear Systems (MJLS) with improved inference techniques for better model representation \citep*{2017humanoids-rojas-shdp-var-hmm,2015NIPS-Hughes-ScalableAdapStateComplex_npHMM}). Anomaly identification is a fault detection service that if flagged, triggers anomaly classification services that cluster signals with broadly similar structure.

Once an anomaly is classified, a recovery policy is executed. Recovery policies include re-enacting or adaptive policies. Re-enacting policies resolve accidental (one-off) anomalies, while adaptive policies resolve persistent anomalies. Re-enactments re-attempt either a current or previous manipulation skill but with new goal parameterizations. Re-enactments are learned from human users by modeling human recovery choices through a multinomial distribution of task nodes. Once learned, new node transitions are introduced in the graph for specific accidental anomalies at specific nodes. For adaptive policies, the robot requires user intervention to provide skill training to overcome a persistent anomaly at a given point in the task graph. Once an adaptive recovery is trained (including both skill generation and introspection models), it is introduced into the graph while retaining previously learned policies from the parent node. The approach fashions a system that incrementally learns anomalies globally and recoveries contextually (Sec. \ref{sec:recovery}).

A co-bot experiment performing kitting tasks is used as a proof-of-concept. A human collaborator places objects in a collection bin that the robot has to package. We hold that tedium and monotony on the human collaborator part result in the introduction of a variety of external disturbances or anomalies to the robot system. 
We demonstrated that we could not only identify anomalies reliably (overall accuracy of 93.09\%) but also classify them in an online fashion (overall accuracy of 96.15\%). And that given simple task-level recovery policies, we could also recover consistently and reliably most of the time. The tight integration achieved in this work enabled robots to continue functioning more than 82\% across all our anomaly scenarios and 95\% in more typical scenarios.

The framework showed interesting functionality including: (i) the ability to introspect and recover from anomalies that occurred \textit{during recovery activities themselves} and (ii) the ability to self-correct. Even in situations where the initial classification and recovery policy where wrong, the system at times quickly self-corrected and completed the task successfully. The current framework has broad applicability to all manipulation domains that suffer from uncertainties in unstructured environments: making industrial and service robots prime candidates for this technology.
Extensions 1-5 include supplemental video, dataset, results and analysis, and robot-agnostic source-code for the co-bot kitting experiment with anomalies and recovery information. The supplemental information is also accessible at \citep*{2018IJRR-Wu-RecoveryFramework_supplementalURL}.
\section{Overview} \label{sec:overview}
In this section we introduce a high-level overview of the system along with relevant notation. A summary of all notation can also be found in Appendix \ref{app:notation_table}.
\begin{figure*}     
	\centering		
		\includegraphics[scale=1.45]{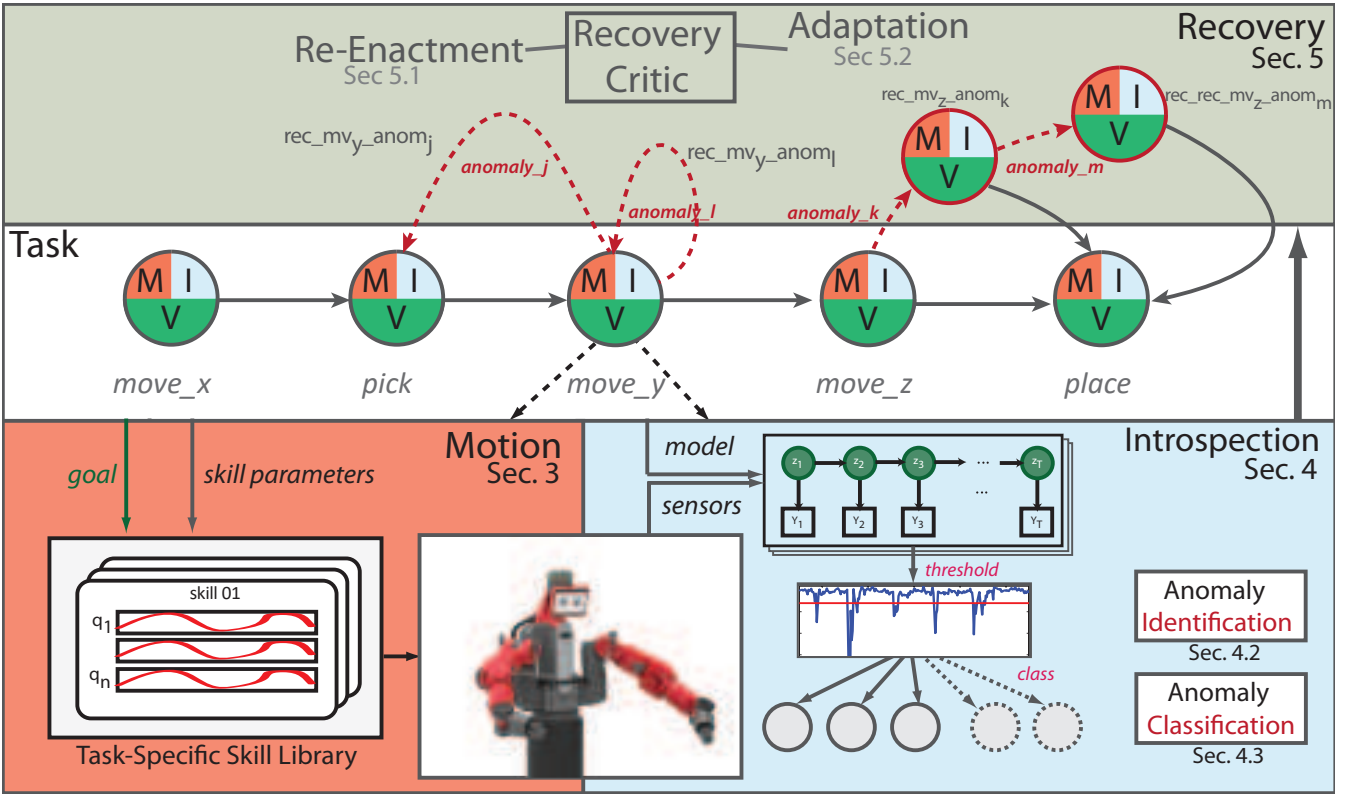}
		\caption{Manipulation tasks are controlled through a graph based scheme consisting of nodes and edges. Each node contains three types of modules: motion, visualization, and introspection; all of which run in parallel. Motion modules use pose goals provided by the visualization module as well as node-specific skill parameters to generate desirable skills. Introspection modules use node-specific models, parameter, and hyper-parameter settings to continually look for anomalies. If identified, the introspection modules further classifies them. A recovery critic then issues a policy for re-enactment or adaptation.}\label{fig:framework}   
\end{figure*}

Directed graphs are a useful tool to manage complexity in manipulation tasks \citep*{2015ICRA-Kroemer-TwrdsLearnHierSkillsMultiPhaseManip,2015IJRR-Niekum-LrnGrnddFiniteStateReprUnstrucDems,2015RSS-Kappler-DateDrivenOnlineDecisionMakingManipu}. Motion comprises structure, not unlike that of grammar, that can be captured as a set of motion primitives and associated sensory-motor perceptions (\citep*{2017IROS-Rojas-BotIntrospecWrenchBasedActionGrammars,2006CAS-LinHeger-TowardsAutomaticSkillEvaluation:DetectionSegmentationRobotASMotion,2006IEEETBiomedEng-Rosen-GeneralizedApproachModelingMIS_DiscreteMarkov}). 

Manipulation tasks are represented as a graph \GC\gap that consists of a sequence of behaviors. 
Behaviors \BC\gap in turn are composed as either simple or compounded actions,  where actions are represented by nodes \NC\gap. Actions are connected by transitions \TC\gap and as such, behaviors too are connected by transitions. A node transition from a node $\NC_s$ to another node $\NC_t$ is denoted as: $\TC_{s,t}=\{s,t \in \NC\}$. The manipulation graph is thus the set of nodes and transitions: $\GC:\{ \NC,\TC \}$. We also introduce a pair of additional definitions for behaviors: (i) behaviors are also referred to as phases in a manipulation task. Phases imply temporal progression, hence given behaviors a temporal context in the accomplishment of a task. (ii) The behaviors with which any task is bootstrapped are also referred to as milestones $\BC=(\BC_1,...,\BC_i)$, which indicate that it will be these behaviors that define key points in the task and will play a significant part in accomplishing the task. 

As we introduce recovery policies, more concretely adaptive policies, we will generate simple adaptive behaviors that are composed essentially of (adaptive) nodes and denominated as $\mathcal{N}_{ij}$. Adaptive nodes will be pushed in-between milestones (Sec. \ref{subsec:adaptive}). The node insertion generates a new graph branch that connects the current behavior to the subsequent milestone (see the $rec\_mv_z\_anom_k$ node in Fig. \ref{fig:framework}). It is also possible to introduce further adaptive nodes $\mathcal{N}_{ijk}$ in existing branches (see the $rec\_rec\_mv_z\_anom_m$ node in Fig. \ref{fig:framework}) if a new adaptation takes places as a result of an anomaly \FC\gap during a recovery skill. In this way, the set of nodes in a task, those within milestone behaviors and those in branched nodes $\NC=\{ \mathcal{N}_i \bigcup \mathcal{N}_{ij} \bigcup \ldots \bigcup \mathcal{N}_{ij...q} \}$, can incrementally grow over time as new capabilities are introduced. 

We now turn our attention to a node's internal functionality. A node does more than simply generate motion. Nodes are composed of of a set of modules which run as parallel processes. Generally speaking, modules can encapsulate a wide range of functions like: skill generation, introspection, visual goal setting (visualization for short), natural language processing, navigation, to name a few. For this work, we restrict node modules to: skill generation $\SC$, visualization $\VC$, and introspection $\MC$. In a given task, skill modules $\SC_m=\{ S_1,...,S_M\}$ perform the necessary motor skills to achieve a task (Sec. \ref{sec:skills}). Visualization modules $\VC_m=\{ V_1,...,V_M\}$ process goal targets for specific motor skills (Sec. \ref{subsec:skill_training}). Introspection models $\MC_m=\{ M_1,...,M_M\}$  aid a robot to understand the types of skills or anomalies that are experienced within a task. In our work, we generate and maintain skill, visual, and anomaly libraries on a per-task basis 
\endnote{In this work we do not explicitly study the re-use of motor skill and introspection models that might repeat across tasks, though this too is an important factor in the re-usability and scalability of these systems}.

The introspection module is in charge of triggering anomaly flags when the system experiences sensory-motor signatures that deviate from those expected in the currently running node. Once an anomaly triggered, the introspection system will provide a classification $\FC_x$ to the anomaly. Classifying anomalies is by nature more challenging than classifying nominal skills as the variability under which anomalies occur is much larger (see Sec. \ref{sec:robot_introspection}). Similarly, acquiring data for failure activities also brings challenges: discovering a set of anomalies in a task is not a straight forward process, deciding on how to discriminate between them is also not trivial. The policy under which anomalies are re-generated can be controversial: should they be induced or expect to occur accidentally. Sec. \ref{subsec:overview_anomaly_collection} further comments on these issues.

Once an anomaly has been classified, recovery actions \RC\gap are necessary. A recovery agent or critic issues one of two types of recovery policies: re-enactment policies $\RC_R$ or adaptive policies $\RC_A$. Re-enactment policies are applied to anomalies that are distinctly accidental (one-off events), while adaptive policies are applied to anomalies that are persistent (i.e. anomalies that occurs repeatedly). Re-enactment policies, re-attempt a previously enacted skill that is selected as a function of the anomaly that occurred. That is, a re-enacting policy issues a transition from the current node $\NC_i$ to a designated goal node $\NC_g$ such that $\RC_R:\TC_{\NC_i,\NC_g}$ (see Sec. \ref{subsec:revertive}). For adaptive policies, the robot requires user intervention to train a motor skill to overcome the persistent fault $\FC_x$. Once an adaptive recovery is trained, it is added into the graph such that: $\RC_A: S_{N_i}|\FC_x \rightarrow N_{ij}(Sm,Mm,Vm)$. In this way, adaptive recoveries are incrementally introduced to the system as persistent anomalies appear (see Sec. \ref{subsec:adaptive}).
\subsection{Experimental Setup} \label{subsec:experimental_setup}
In this section we introduce a co-bot-based Kitting experiment selected to test our anomaly classification and recovery policies. We also present the experimental testbed and manipulated objects. Details regarding external disturbances and data collection techniques are also described.
\subsubsection{Kitting Experiment}\label{subsubsec:kitting_exp_description}
The collaborative kitting experiment consists of a robot and a human co-worker that closely collaborate to place a set of goods in a packaging box. The human co-worker is tasked to place a set of 6 objects on the robot's ``collection bin'' (located in front of the robot) in a one-at-a-time fashion as shown in Fig. \ref{fig:home}. The objects may accumulate in a queue in front of the robot. As soon as the first object is on the table, the robot identifies the object and begins the placing process in the packaging box located to the right of the robot. Thus, the robot picks an object (Fig. \ref{fig:pick}) and transports it towards the box (Fig. \ref{fig:box}), after which, the robot appropriately places it in the box (Fig. \ref{fig:place}). 

\begin{figure}[b]     
	\centering		
		\includegraphics[width=\linewidth]{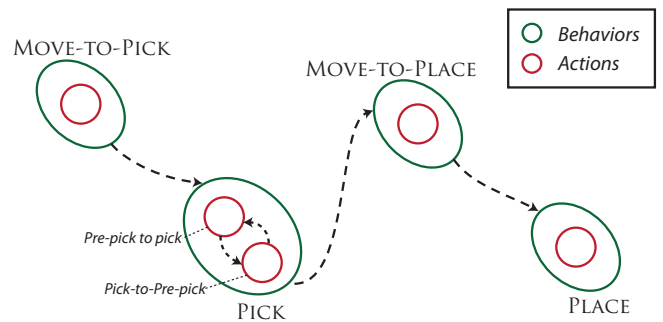}
		\caption{Task graph for kitting experiment composed of 4 behaviors and 5 actions. The pick behavior is compound and consists of two actions. The task is thus bootstrapped with 5 nodes that encompass 5 skills, goals, and introspection models. Modules are not shown explicitly in the node actions for clarity.}
        \label{fig:behs_and_actions}   
\end{figure}

The kitting task is originally bootstrapped with 4 behaviors $\BC\gap$ and 5 actions $\NC\gap$ as shown in Fig. \ref{fig:behs_and_actions}.
All behaviors except pick consists of single actions or nodes. The compound pick behavior consists of two nodes: pre-pick to pick and pick to pre-pick. The task requires that we train 5 actions and as such 5 skills, visualization goals, and  introspection models. However, in the rest of the paper, we will describe the task only in terms of the 4 high-level behaviors for simplicity. Recall that the original graph will grow as adaptive nodes are learned when adaptations are necessary.
\subsubsection{External Disturbances}\label{subsubsec:external_disturbances}
In this section we try to motivate the kinds of external disturbances that may be typical of a collaborative environment in a human-robot collaboration setup in a warehouse-like job as the one described in Sec. \ref{subsubsec:kitting_exp_description}. Despite collaboration, we think that collaborative tasks, kitting in this case, might still result in low-cognitive demands for the human user. The low-cognitive load might lead to monotony which would then cause boredom and attention-loss. In such cases, a human co-worker may be more likely to accidentally collide with the robot or alter the environment in unexpected ways. For example, the user may accidentally collide or unintentionally move a packaging object in ways the robot cannot model or anticipate as it tries to grip the objects. Object shifting (objects to be grasped or event the packaging box) may lead to tool-collisions, failed grasps, or even air grasps (where the object was completely removed). There also exists the possibility that picked objects may at times slip from the robot's tool if the grasp is not optimal; or if upon motion, inertial forces acting on the object cause dynamics that break the grasp. The system may even experience a chain of anomalies: human collisions that lead to object slips that move objects in such a way that lead to air grasps. As part of the discovered anomalies from Sec. \ref{subsec:overview_anomaly_collection}, we introduce the basic anomaly types and their acronym in the interest of brevity: human collisions (HC), tool collisions (TC), object slips (OS), and no-object (NO). Sec. \ref{sec:robot_introspection}, will introduce the introspection methodology used to model robot skills including a description of our Anomaly Identification algorithms in Sec. \ref{subsec:identification}) and Anomaly Classification algorithms in Sec. \ref{subsec:anomaly_classification}. Later, in Sec. \ref{sec:recovery} we introduce our recovery critic policies including Re-enactments (Sec. \ref{subsec:revertive}) and Adaptations (Sec. \ref{subsec:adaptive}).
\subsubsection{The Robot}
A Baxter humanoid robot's right arm is used to pick commonplace objects set before him. The equipment used with the robot is: a 6 DoF Robotiq FT sensor, the standard Baxter electric pinching fingers, and Baxter's left hand camera. Each finger is further equipped with a multimodal tactile sensor composed of: (i) a four by seven taxel matrix that yield absolute pressure values, (ii) a dynamic sensor which provides a single capacitive reading in millivolts (mV) useful to detect tactile events, and (iii) an IMU and gyroscope \citep*{2017ICRA-maslyczyk-CapacitiveSensor}. Baxter's left hand camera is placed flexibly in a region that can capture objects in the collection bin with a resolution of 1280x800 at 1 fps (we are optimizing pose accuracy and lower computational complexity in the system) as seen in Fig. \ref{fig:home}. The use of the left hand camera facilitated calibration and object tracking accuracy. ROS Indigo on Linux 14.04 and a number of workstations are used to control all aspects of the experimentation. Code is available in our supplementary page \citep*{2018IJRR-Wu-RecoveryFramework_supplementalURL}.
\subsubsection{Objects}
\begin{figure*}     
	\centering		
        \subfigure{\label{fig:home}\includegraphics[scale=1]{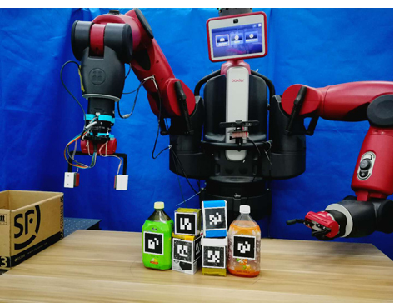}} \hspace{0.01cm}
        \subfigure{\label{fig:pick}\includegraphics[scale=1]{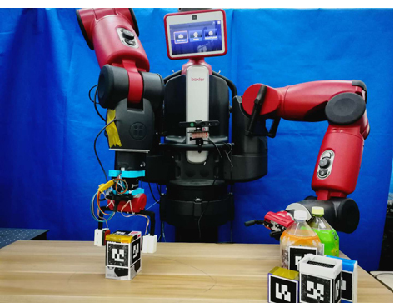}} \hspace{0.01cm}
        \subfigure{\label{fig:box}\includegraphics[scale=1]{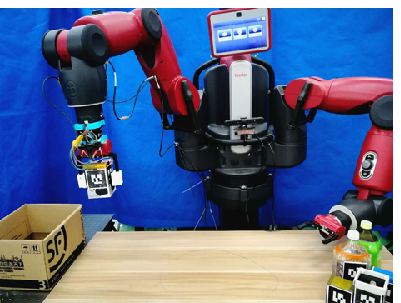}} \hspace{0.01cm}
        \subfigure{\label{fig:place}\includegraphics[height=3cm,width=4cm]{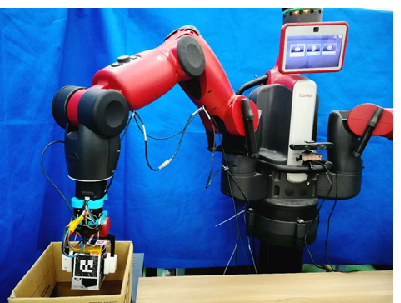}}
		\caption{Snapshots of a kitting experiment. Objects that need to be packaged are placed by a human collaborator before the robot in a collection bin. The shared workspace affords possibilities for accidental contact and unexpected alteration of the environment. The robot is tasked to pick-and-place each on of the objects in a packaging box to its right. The visualization module uses the ALVAR tags to provide a consistent global pose with respect to the base of the robot and the introspection system is continually monitoring for anomalies and their types. If an anomaly is classified, the recovery critic selects from amongst two policies to try to restore the task flow and reach the next milestone in the task. The ultimate result is to recover successfully and in doing-so help the robot achieve longer-term autonomy.}
        \label{fig:experimental_setup}   
\end{figure*}

A set of 6 common household objects consisting of box-liked shapes and bottles were used in our work as shown in Fig. \ref{fig:home}. The objects ranged in weight from 0.0308kg to 1.055kg and in volume from $3.2\mbox{ x }10^{-04}m^3$ to $1\mbox{ x }10^{-03}m^3$. The object's surfaces also varied slightly: some heavier objects had sleeker surfaces that incited object slips---we believe not an unreasonable determination as warehouses contain a wide variety of objects---whilst other objects had rougher surfaces. Across trials, object locations and order was varied to promote generalization.

Alvar tags, with 0.06m sides, were placed around the circumference of the objects for robust visual recognition (ALVAR can handle change in lighting conditions, optical flow-based tracking, and good performance for multi-tag scenarios) regardless of orientation (Fig. \ref{fig:experimental_setup}). 
\subsection{Cataloging Experiments}\label{subsec:overview_anomaly_collection}
In this section we provide brief overviews of the data collection process for skill \SC\gap and introspection \MC\gap modules. Detailed presentations will be found in Sec. \ref{sec:skills} \& \ref{sec:robot_introspection} respectively.
\subsubsection{Motion Skill Training}
In this work, motion skills are encoded through DMPs. DMP training uses one-shot kinesthetic demonstrations to teach five skills each of the four skills needed to bootstrap the behaviors for the kitting task.
\subsubsection{Deducing Anomalies}\label{subsec:deducing_anomalies}
As for the process of discovering what anomalies might exist in a given task, we must express that, undeniably, robot researchers hold a bias towards which anomalies will exist and be discovered in a given task. To this end, in our work, we aim to discover the anomalies in the task by emulating a collaborative kitting task where the human collaborator experiences tedium and monotony and leads to unintentional changes or disturbances in the environment or the robot respectively. 

To this end, we tasked 5 robot researchers to act as a collaborative co-worker in a kitting task with the Baxter humanoid robot under the monotonous conditions already mentioned (Sec. \ref{subsubsec:external_disturbances}). Each user was trained to place the set of six household objects, one-at-a-time, in the collection area. From this exercise we extract two pieces of information: (a) anomaly classification labels (as judged by a human expert) that emerge from the task (those mentioned in Sec. \ref{subsubsec:external_disturbances}, namely HC, TC, OS, NO) and (b) the recording of sensory-motor data surrounding the anomalous event. We do this by considering a window of $\pm2 secs.$ and recording through the use of an online database system\endnote{We use the Redis database for this purpose (https://redis.io/) as rosbags can only be processed offline.}. The sensory-motor data collected at this stage, allows to build basic models of the anomalies further described in Sec. \ref{sec:robot_introspection}. One thing to note for anomaly classification is that in this work we attempted to classify anomalies broadly. Consider for an example a human collision: regardless of user, high or low collision, right or left, even temporal occurrence in the task, all of these are sought to be classified as the same single event of human collision. The same principle applies across the rest of the anomalies. Our approach to classification is much broader than similar works \citep*{2018AutBot-Park-MultimodAnomDet_AssistiveBots} and renders the classification task much more challenging. Coupled with the fact that only a limited number of trials is available for training, the modeling task is challenging. 
\subsubsection{Training and Inducing of Anomalies}
Beyond the original data collection performed in Sec. \ref{subsec:deducing_anomalies}, a second data collection round is conducted to improve training (parameter and hand-designed feature tuning). This round is performed iteratively seeking to maximize optimal performance while protecting against overfitting. The final number of training and testing trials used for anomaly identification and classification are described in Exp. 1 and Exp. 2 respectively.
\subsubsection{Learning Recoveries}
Upon the occurrence of accidental one-off anomalies, re-enactment recovery policies are learned from human users. Exp. 3 is used to learn probability models from human users given specific anomalies (see Sec. \ref{subsubsection:re-enactment_polilcy} for details). Similarly, for persistent anomalies, adaptive recoveries are incrementally trained through kinesthetic teaching. In Exp. 4, 5, and 6 a variety of adaptive skills are learned to address specific and emerging anomalies (see Sec. \ref{subsec:adaptive} for details).
\section{Motor Skills} \label{sec:skills}
In manipulation, motor skills are compact action representations that are extracted from continuous high degree-of-freedom (DoF) robot motions \citep*{2013NC-Ijspeertdynamical-CMP_LrnAttracMdls_MtrBeh,2013NIPS-Paraschos_Peters-ProMPs,2016Arxiv-Meier-ProbabilisticDMPs,2017Humanoids-Longxin-HRC,2010RAM-Calinon-LrnReprodGesturesImitation}. Attractive qualities in motor skill representations include stable dynamics when attractor points (start and goal locations) or temporal scales are changed along with flexible re-use like blending or parallelizing primitives. Techniques from dynamical systems like Dynamical Motion Primitives (DMPs), or from probability like Probabilistic Motion Primitives (ProMPs) are widely used to encode manipulation task information. In this work, we encode motions using DMPs though we can handle any manipulation approach by extracting key parameters into the framework's motion module library.

The DMP framework encodes dynamical systems through a set of nonlinear differential equations whose point attractor system is defined by a nonlinear forcing function, which in turn depends on a canonical system for temporal scaling. For a one DoF point attractor system, the point attractor system is defined as \citep*{2009ICRA-Pastor-LfD_DMP}: 
\begin{equation}
	\tau \dot{v}=K(g-x)-Dv-K(g-x_0)s + Kf(s),
    \label{eqtn:dmp_attractor_system}
\end{equation}
\begin{equation*}
	\tau \dot{x}=v.
    \label{eqtn:dmp_velocity}
\end{equation*}
Eqtn. \ref{eqtn:dmp_attractor_system}, is an extended PD control signal with spring and damping constants $K$ and $D$ respectively, position and velocity $x$ and $v$, goal $g$, scaling $s$, and temporal scaling factor $\tau$.

The scaling term originates from an additional system, called the canonical dynamical system, which controls the system's phase execution:
\begin{equation}
	\tau \dot{s} = -\alpha s,
    \label{eqtn:dmp_canonical_system}
\end{equation}
and where $\alpha$ can be an arbitrary constant. 

The forcing term $f(s)$ is used to alter attractor point dynamics and achieve an arbitrary trajectory (often learned from demonstration \citep*{2009ICRA-Pastor-LfD_DMP}). The forcing term can be defined as a phase-dependent linear combination of basis functions $\psi_i(s)$:
\begin{equation}
	\tau f(s)= \frac{\sum_{i=1}^{N} w_i \psi_i (s)s} {\sum_i \psi_i(s)}.
    \label{eqtn:forcing_function}
\end{equation}
Gaussian distributions with mean $c_i$ and variance $h_i$ were used as basis functions: $\psi(s)=exp(-h_i (s-c_i)^2)$. The forcing function is the linear combination of basis functions with variable weights $w_i$ and normalization constant $\sum_i \psi_i(s)$. Phase $s$ monotonically decreases from 1 to 0 to control phase progress by activating Gaussian distributions centered at $c_i$. The diminishing phase value guarantees the vanishing of the forcing term leaving the simpler point attractor dynamics to converge to the target. Spatio-temporal scaling is possible through the $(g-x)$ term in Eqtn. \ref{eqtn:dmp_attractor_system} performs spatial scaling enabling the system to adjust to varying goals. Finally, system speed-up (or slow-down) is possible through the $\tau$ variable in Eqtn. \ref{eqtn:dmp_velocity} as well.
\subsection{Learning from Demonstration}\label{subsec:skill_training}
Forcing term weights are learned from demonstration. Using kinesthetic teaching $x(t), \dot{x}(t), \ddot{x}(t)$ with duration $T$ are extracted as in \citep*{2015IJRR-Niekum-LrnGrnddFiniteStateReprUnstrucDems}. The target forcing term is computed by rearranging Eqtn. \ref{eqtn:dmp_attractor_system}, integrating it with Eqtn. \ref{eqtn:dmp_canonical_system}, and substituting appropriate values to convert from time-mode to phase-mode as shown in Eqtn. \ref{eqtn:forcing_target}.
\begin{equation}
	f_{target}(s)=\frac{ -K(g-x(x))+D\dot{x}(s)+\tau\ddot{x}(s) } {g-x_0}.
    \label{eqtn:forcing_target}
\end{equation}
Next, the goal is set to $g=x(T)$ and $\tau$ is selected such that a DMP reaches 95\% convergence at $t=T$ before using standard linear regression to compute the weights $w_i$. Such procedure yields a baseline controller that can be improved by reinforcement learning \citep*{2005RR-schaal-LearningMPs} though this is not done in this work.

Motor skills are trained as individual skills (more robust methodologies \citep*{2010IROS-Grollman_Jenkins-IncrmntlLrn_Subtasks_UnsegDemos, 2012IJRR-Konidaris-LfD_SkillTrees, 2015IJRR-Niekum-LrnGrnddFiniteStateReprUnstrucDems} were not used here) for each phase of the task. Cartesian position and XYZ Euler representations are used to encode the attractor dynamics. 

With respect to introspection models, we leverage sensory-motor signatures to learn the structure of sensory responses to motion data \citep*{Rojas_n_Peters:2005:ABBI,2015RSS-Kappler-DateDrivenOnlineDecisionMakingManipu}. Our observations consist of a 6 DoF end-effector twist and wrench respectively, a 7 DoF pose (using quaternions as orientation), and 56 tactile values (each finger has 4-by-7 taxels). All observations were hand-processed into features as detailed in Sec. \ref{subsec:signal_processing}. 
All object poses are acquired using AR codes through the ROS ALVAR framework \endnote{\url{http://wiki.ros.org/ar_track_alvar}.}. 

As previously mentioned in Sec. \ref{subsubsec:kitting_exp_description}, we use kinesthetic teaching to train five simple skills: move-to-pick, pre-pick-to-pick, pick=to-pre-pick, move-to-box, and place. We ensure that skills are executed in such a way that no occlusion occurs. Skills are executed at least 7 times to obtain sensor information of nominal skills which is used by the introspection models to first implement anomaly identification (as described in Sec. \ref{sec:robot_introspection} and also seen in \citep*{2017humanoids-rojas-shdp-var-hmm}). Once DMP and introspection models are trained, they are stored in their corresponding libraries. Then, a behavior graph is constructed where nodes contain appropriate ID types that are handled by the system to enact necessary models during task execution. As for transitions, \textit{nominal nodes} currently transition to only one other node, so no explicit transition classification is enacted. For anomalies however, transitions to different nodes will depend on the anomaly classification (Sec. \ref{subsec:anomaly_classification}) and the re-enactment policy of our critic critic (Sec. \ref{sec:recovery}). 
\subsection{Goal Setting}\label{subsec:goal_setting}
For task execution, the Visualization module is responsible for selecting appropriate goal targets to enacted skills. While the goal is that the visualization module uses task affordances to select appropriate target goals in a skill, currently goal targets are pre-specified according to the nature of the skill. Pre-pick nodes use the Alvar code pose of objects in queue order from right-to-left. Pick nodes are set to the pose of actively actively tracked objects. The move-to-box skill uses the centroid location of the flat plane of the box. Place skills use packaging box locations set according to the number of objects already picked in the task. 

Additionally, we highlight that though the skill set used in this work is simple, the space of possible anomalies is significant and is this work's main focus. To this end, in our experimentation, we test strenuous anomalous conditions that could emerge in unstructured environments. (Sec. \ref{sec:experiments}).
\section{Robot Introspection} \label{sec:robot_introspection}
\begin{figure*}     
	\centering		
		\includegraphics[width=\linewidth]{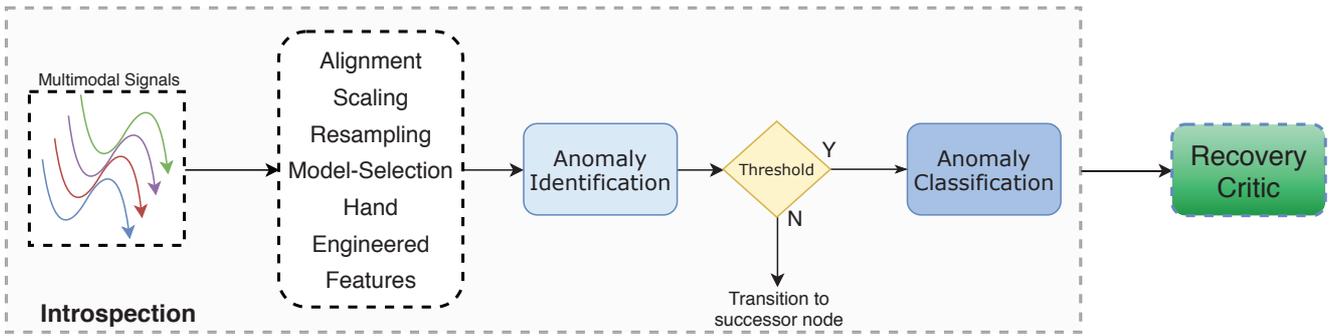}
		\caption{The robot introspection module flow diagram. A continuous anomaly detector analyzes whether a sensory-motor signal deviation occurs compared to nominal actions. If so, an anomaly flag is triggered and a classification label provided (Sec. \ref{sec:robot_introspection}). After introspection, control is directed by the recovery critic (Sec. \ref{sec:recovery}).}
        \label{fig:anomaly_flow_chart}   
\end{figure*}

Robot introspection is a precursor to policy recovery. A non-parametric Bayesian MJLS system is used for anomaly identification and classification. This section will first introduce the Bayesian non-parametric model and then present the specific techniques used for anomaly identification and classification. Fig. \ref{fig:anomaly_flow_chart} summarizes the introspection system flow.
\subsection{Bayesian non-parametric Hidden Markov Modeling}\label{subsec:shdp-var-hmm}
Robot introspection uses Bayesian non-parametric Markov Jump Linear Systems (MJLS) and memoized variational inference with scalable adaptation as the modeling mechanism. A non-parametric Bayesian HMM, namely the the sticky Hierarchical Dirichlet Process HMM can be used to learn a VAR process (sHDP-VAR-HMM). Such an approach enables us to both learn the model complexity (also the mode or number of latent states) directly from the data. The VAR switching process allows to model mode-specific observations through linear dynamics \citep*{2010-Fox-BNP_LearningMarkovSwitchingProcesses}. sHDP-VAR-HMMs have been successfully used to model dynamical signals like bee dancing, stock market behavior, human activity models, nominal robot manipulation signals amongst others \citep*{2010-Fox-BNP_LearningMarkovSwitchingProcesses,2013IROS-DiLello-BayesianContFaultDetection,2015IJRR-Niekum-LrnGrnddFiniteStateReprUnstrucDems,2017humanoids-rojas-shdp-var-hmm}. Recent advances in variational inference allow to process large datasets incrementally and optimize the creation and removal of states yielding highly optimized models that are simpler, more compact, more interpretable, and better aligned to ground truth state segmentations \citep*{2015NIPS-Hughes-ScalableAdapStateComplex_npHMM}. In this section we first describe the standard Hidden Markov Model, followed by the sHDP-VAR-HMM, followed by variational inference concepts.

\subsubsection{Hidden Markov Models}\label{subsubsec:introspec_hmm}
\hfill\\
HMMs are a doubly stochastic and generative process used to make inference on temporal data. The underlying stochastic process contains a finite and fixed number of latent states or modes $z_t$ which generate observations $\XC =\{ x_t\}_{t=1}^{N}$ through mode-specific emission distributions $b(z_t)$. These modes are not directly observable and represents sub-skills in a given task node. Transition distributions, encoded in transition matrix $\pi_{ji}$, control the probability of transitioning across modes over time. Given the initial mode distribution $\pi_0$ and a set of observations, the Baum-Welch algorithm is used to infer model parameters $\Pi=(\pi,b)$. HMMs assume a fixed number of latent states as well as mode-specific conditionally independent observations. Such assumptions limit the expressive power of HMMs as they are unable to derive natural groupings and model complex dynamical phenomena.
\subsubsection{The sHDP-VAR-HMM}\label{subsubsec:introspec_shdp_hmm}
\hfill\\
Bayesian non-parametric priors extend HMM models to learn latent complexity from data as well as the transition distribution of an HMM \citep*{2010-Fox-BNP_LearningMarkovSwitchingProcesses,2014AOAS-Fox-JointModel_MutlipleTimeSeries_BetaProcess_MotionCapture,2015NIPS-Hughes-ScalableAdapStateComplex_npHMM,2017humanoids-rojas-shdp-var-hmm}. This section introduces key concepts of the sHDP-VAR-HMM (although for an extended presentation see \citep*{2009PhD-Fox-BNP}). To allow for a flexible number of latent states, priors on probability measures $G_j$ that have an unbounded number of support points $\theta_k$ can be used. Dirichlet Process's (DP) are known for their clustering properties (\ie the Chinese restaurant process) across countably infinite modes $\theta_k$ and provides a distribution over the support points according to Eqtn \ref{eqtn:g0}.
\begin{eqnarray}
	G_0=\sum_{k=1}^{\infty} \beta_k \delta_{\theta_k}, \nonumber		\\
    & \theta_k \sim H, \mbox{and}	\\
    & \beta | \gamma \sim GEM(\gamma). 					& \nonumber        \label{eqtn:g0}
\end{eqnarray}
Here, $H$ is a base distribution, and $\beta_k$ are weights sampled via a stick-breaking process generally represented as $GEM(\gamma)$. The DP allows to sample observations without explicitly constructing an infinite probability measure $G_0 \sim DP(\gamma, H)$. Instead, it is possible to use the DP as a prior for the set of HMM transition probability measures $G_j$. However, this construction as it stands, would consistently generate independent HMM modes between transition steps. The goal is to define the probability measures $G_j$ on a common base of support points and let $G_j$ produce a variation on the global discrete measure $G_0$.

So, through a Bayesian hierarchical specification $G_j \sim DP(\alpha,G_0)$, where $G_0$ which itself draws from $DP(\gamma,H)$, it can be shown that the probability measures are:
\begin{eqnarray}
    G_j = \sum_{k-1}^\infty \pi_{jk} \delta_{\theta_k} 	& \pi_j | \alpha,\beta \sim DP(\alpha,\beta).
\end{eqnarray}
The HDP-HMM, in this form, does not yet differentiate self-transitions from moves between distinct latent states and allows for fast switching dynamics between them and causing significant posterior uncertainty. For this reason, a ``sticky'' self-transition bias parameter is introduced that favors self-transitions \citep*{2010-Fox-BNP_LearningMarkovSwitchingProcesses}.

As for observation models, the sHDP-HMM can be used to learn VAR processes, which are useful to model complex phenomena. The transition distribution is defined as in the sHDP-HMM case, however, instead of independent observations, each mode now has conditionally linear dynamics, where the observations are a linear combination of the past $r$ mode-dependent observations with additive white noise. In our case, we consider the first-order ($r=1$) auto-regressive Gaussian likelihoods that is the observations are a noisy linear combination of the previous observation plus additive white noise $e$, with observation $x_{t}$, can be defined as 

\begin{eqnarray}
\begin{aligned}
    \mbox{Observation Dynamics: } e_t & \sim \mathcal{N}(0,\Sigma_k) \\
    	x_t &= A_{k} x_{t-1} + e_t(z_t = k) \\
    \mbox{Mode Dynamics: }
    	z_t^{(i)} & \sim \pi_{z_{t-1}^{(i)}}^{(i)}.
        \label{eqtn:gen_proces_shdp_var_hmm}
\end{aligned}
\end{eqnarray}
Where, each state $k$ is composed of time-invariant regression matrix coefficients $\boldsymbol{A}$ and a covariance matrix $\boldsymbol{\Sigma}$ are necessary. The generative process for the resulting HDP-AR-HMM is then found in Eqtn \ref{eqtn:gen_proces_shdp_var_hmm}.

Both $\boldsymbol{A}$ and $\boldsymbol{\Sigma}$ for specific latent states are both uncertain, they need to be learned. The parameters $\boldsymbol{\theta} = \{\boldsymbol{A}, \boldsymbol{\Sigma}\}$ are approximated for each state by defining a conjugate prior distribution on them. Particularly, a Matrix Normal Inverse Wishart (MNIW) is used as a conjugate prior distribution when both $\boldsymbol{A}$ and $\boldsymbol{\Sigma}$ are uncertain. If only the covariance is uncertain, the conjugate prior is defined as $d-$dimensional Inverse Wishart (IW) distribution with covariance parameter $\Delta$, a symmetric positive definite scale matrix and $\nu$ the degrees of freedom as in Eqtn. \ref{eqtn:covariance_iw}.
\begin{equation}
	 \boldsymbol{\Sigma} \sim \mathcal{IW}(\nu,\Delta)
     \label{eqtn:covariance_iw}
\end{equation}
The full definition of this joint prior is found in \citep*{2015NIPS-Hughes-ScalableAdapStateComplex_npHMM} and defined as $\mathcal{NIW}(\kappa,\boldsymbol{\vartheta},\nu,\Delta)$. For the \textit{IW}, the first moment of the distribution is:
\begin{equation}
	 \mathbb{E}[\boldsymbol{\Sigma}]=\frac{\nu\Delta}{\nu-d-1}.
     \label{eqtn:iw_first_moment}
\end{equation}
where, $\nu$, is the degrees of freedom. The expectation of the covariance, for $N$ exemplars of data $\XC_N$ for a given skill and a sequence with length $T_n$, is defined as:
\begin{equation}
	\mathbb{E}[\boldsymbol{\Sigma}] = s_F \sum_{n=1}^N\sum_{t=1}^{T_n}(x_t - \overline{x})(x_t - \overline{x})^T. 
    \label{eqtn:expectation_of_covariance}
\end{equation}
Then, to determine the matrix $\boldsymbol{A}$ of regression coefficients, we use the matrix-normal inverse wishart (MNIW) distribution, which places a conditionally matrix-normal prior on A (for a given latent state) such that:
\begin{equation}
	\boldsymbol{A} | \boldsymbol{\Sigma} \sim 
    \mathcal{MN}(\boldsymbol{A};\boldsymbol{M},\boldsymbol{\Sigma},\boldsymbol{K})
\end{equation}
The matrix normal is computed once $\boldsymbol{\Sigma}$ is available, where the covariance $\boldsymbol{\Sigma}$ represents the covariance across the rows, while $\boldsymbol{K}$ represents the covariance across the columns. 

By using the model over a set of multi-modal exemplar data $\XC_n$, the sHDP-AR-HMM can discover and model shared behaviors in the anomaly data across exemplars, even from a few examples. This model does assume however that all exemplars share the same (latent) modes and that modes switch amongst themselves in the same way). It is also possible to use a beta-process prior \citep*{2010-Fox-BNP_LearningMarkovSwitchingProcesses} to avoid this limitation, but this has not yet been implemented for online performance. Pseudo-code for the generation of skill models using the sHDP-VAR-HMM is outlined in Algorithm \ref{alg:shdpHMM}.
\begin{algorithm}
  \SetAlgoLined
  \KwIn{\\
    $N_c$: Number of sequences for class $ c \in \mathcal{C}$; \\
    $\{\XC_n\}_{n=1}^{N_c}$: Dataset with $N_c$ sequences, each of length $T_c$; \\
    $N_{i}$: Number of the maximum iteration for learning;\\
    $N_{r}$: Number of runs for the whole learning; \\
    $random\_state$: The random number generator;\\
    $k\_splits$: Number of folds;\\
    $a,b,d,e$: Hyper-prior for concentration parameters; \\
    $\nu,\Delta,V,M,s_F$: Hyper-prior for the MNIW distribution; \\
    $\kappa$: The self-transition bias; \\
    $K$: The truncation active states.\\
  }
  \KwResult{HDP-HMM models for each class}
  $\{\XC_n\}_{train}, \{\XC_n\}_{test}$ = \qquad\qquad KFold\_split($\{\XC_n\}_n^{N_c}, k\_splits, random\_state$) \\
  \For{$k$ in $k\_splits$}{
    \For{$i$ in $N_r$}{
          \For{$n$ in $N_i$}{
              \uIf{not converged}{$\Theta_\pi,\ loss$ = HDPHMM($\{\XC_n\}_{train}, a,b,$
                              \qquad\qquad\qquad\qquad$ d,e,\nu,\Delta,V,M,s_F,\kappa,K$)
                              }
              \Else{HDP-HMM with $\Theta_\pi$ and $loss$}
          }
          $\{loss\}_i^{N_r} \leftarrow loss$ \\
          \uIf{$i == N_r$}{
          $\Theta_\pi \leftarrow $ with the minimum loss value}  
     }
     $\mathcal{L}_{test\_mean} = \frac{\sum_{i=1}^{N_{test}}p(\{\XC_n\}_{i}|\Theta_\pi)}{N_{test}} , i\in N_{test}$ \\
     $\{\mathcal{L}_{test\_mean}\}_k^{k\_splits} \leftarrow \mathcal{L}_{test\_mean}$ \\
     \uIf{$k == k\_splits$}{
     \KwRet{$\Theta_\pi$} with the maximum $\mathcal{L}_{test\_mean}$}
    }
  \caption{sHDP-VAR-HMM Models for Classification}
  \label{alg:shdpHMM}
\end{algorithm}\\
\subsubsection{Memoized Variational Inference with Scalable Adaptation} \label{subsec:introspec_var_inference}
\hfill \\
Prior to the work in \citep*{2015NIPS-Hughes-ScalableAdapStateComplex_npHMM}, inference algorithms for HMMs and HDP-HMMs have not efficiently learned from large datasets nor have they effectively explored data segmentations with varying number of states. Inference algorithms can be trapped at local optima near their initialization points. Stochastic optimization methods, which are unable to update the number of modes after execution, are particularly vulnerable to data segmentation and exploration and local optima \citep*{2014ICML-johnson-StochVarInf_BayesianTS,2014NIPS-Foti-StochVarInf_HMMs}. These methods may yield states that become irrelevant and should be removed. Recently, algorithms that add and remove states via split and merge moves have been designed for non-parametric priors like HDP and BP algorithms \citep*{2014AOAS-Fox-JointModel_MutlipleTimeSeries_BetaProcess_MotionCapture, 2014NIPS-Chang-ParalSampHDPs}. However, these Monte Carlo proposals suffer from scalability as they must use the entire dataset and also require that all sequences fit in memory. 

Hughe's \et memoized variational inference algorithm with scalable adaptation uses birth proposals to create new states and merge and delete moves to remove poor predicting states; however, adaptations are validated through a global variational bound \citep*{2015NIPS-Hughes-ScalableAdapStateComplex_npHMM}. The algorithm caches sufficient statistics and parallelizes local inference steps to efficiently process sequence subsets at each time step to allow for rapid adaptation of the state space cardinality. The inference algorithm outputs all around better models---more compact and interpretable---to infer the sHDP-HMM's posterior distribution leading to better classification results. Please refer to \citep*{2015NIPS-Hughes-ScalableAdapStateComplex_npHMM} for complete details of the algorithm and to \citep*{bnpy} for the open-source code.
\subsection{Anomaly Identification} \label{subsec:identification}
\begin{figure}     
	\centering
		\includegraphics[width=\linewidth]{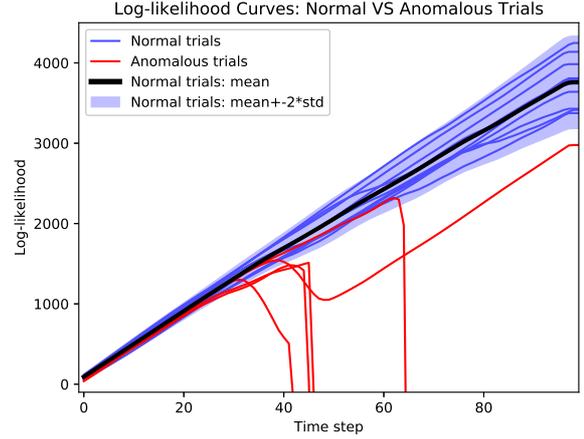}
		\caption{Illustration of cumulative log-likelihood signals between nominal (expected) signals and anomalous ones.}
	\label{fig:log-likehood_anomalies}        
\end{figure}

Anomaly identification continuously monitors robot behavior to identify unexpected behaviors during skill execution and even during recovery phases. Recovery phases are challenging as they usually begin in anomalous states and are more likely to trigger false-positives \citep*{2018ROMAN-Wu-RecovExtDist_StateDep}. Different metrics for anomaly identification have been suggested in \citep*{2016ICRA-Park-MultiModalMonitoringAnomalyDet_RobotManip,2017humanoids-rojas-shdp-var-hmm,2017IROS-Park-MultiModalAnomalyClassifFeeding,2018ROMAN-Wu-RecovExtDist_StateDep}. Most of these techniques use the maximum cumulative log-likelihood value of the observations given a model. In \citep*{2018ROMAN-Luo-RobustVersatileEventDet}, it was shown that such metrics  performance is limited during recovery stages. For instance, Fig. \ref{fig:log-likehood_anomalies} contrasts nominal (expected) log-likelihood signals from anomalous ones. 
In \citep*{2018ROMAN-Luo-RobustVersatileEventDet}, we presented a metric based on the the natural logarithm of the HMM filtered belief state (from hereon referred to as the ``forward gradient'' measure) $\nabla L$. Given an HMM model $\Pi$ and an incoming time series $x_{1:t}$, the natural logarithm of the filtered belief state (see 17.4.1 \citep*{murphy2012machine}) associated with the forward model for latent state $i$ can be represented according to Eqtn \ref{eq:sum_of_log_alpha}.
\begin{align}
	L_t&=\log{\sum_{i=1}^{N}{\alpha_i}(t)}=\log{\sum_{i=1}^{N}{\exp({\log{{\alpha_i}(t)}})}}.
    \label{eq:sum_of_log_alpha}
\end{align}
The forward term can be computed iteratively from the previous time-step result as seen in Eqtn. \ref{eq:sum_of_log_alpha} we have:
\begin{equation}
\log{{\alpha_i}(t)}=
\log{{b_i}(y_{t})}+\log{\sum_{j=1}^{N}{\exp({\log{{{\alpha_j}(t-1)}+\log{A_{ji}}}})}}
\label{eq:log_alpha}
\end{equation}
From Theorem 1 in \citep*{2018ROMAN-Luo-RobustVersatileEventDet}, we established that for an incremental time series $Y$, a good HMM model outputs an incremental Viterbi path that \textbf{\textit{stably expands}} on the previous one. The stable expansion of the Viterbi path is as follows: given a Viterbi path "11223" for an input $x[1\mbox{:}t]$, then the path at $x[1\mbox{:}t+1]$ becomes "11223*", where * is the newly appended hidden state. From this theorem we derived a corollary that established that the forward gradient $L$-curve depends on the latest emission probability of the HMM model, which in-turn depends on the latest observation. The key point is the generation of stable and robust large positive-valued gradients when observations are generated by a its true latent state. 

\begin{figure}     
	\centering
		\includegraphics[width=\linewidth]{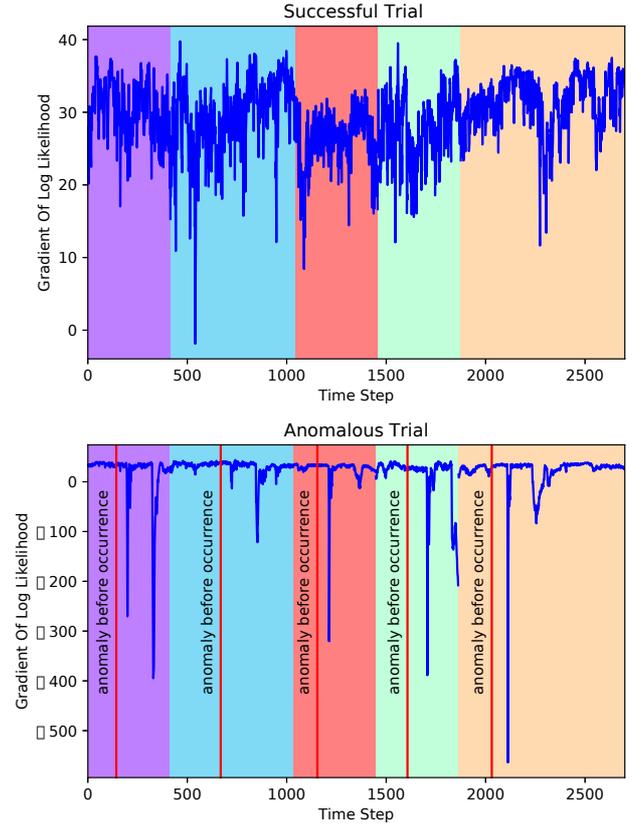}
		\caption{The log-likelihood gradient $\nabla L$ for 5 motor skills $s$ (colored backgrounds) in a task $\BC_e$. Top plot shows a nominal task whose $\nabla L$ is steadily positive (ranges from 10-45 units). Bottom plot shows a trial that experienced one anomaly per skill execution (caused by human collisions to a robot arm). Anomalies occurred shortly after the red vertical lines seen in each skill (marked with "anomaly before occurrence"). When an anomaly occurs, the gradient becomes negative (ranging from $-100s \leq \nabla L\leq  -1000s$), providing distinctive data compared to nominal cases.}
	\label{fig:gradient_comparison_between_success_and_anomaly_trials}        
\end{figure}

Given this fact, anomaly detection using the forward gradient is derived as follows:
given an HMM model $\Pi_s$ (Sec. \ref{subsec:shdp-var-hmm}) representing a certain skill $s$. Let there be $n$ trials of time series exemplar data $\XC_i$ for $i \in \{1, \cdots, n \}$ collected from nominal executions of skills $s \in \SC$; then anomaly detection in a new time series $x$ can be derived as:
\begin{align}
\nabla L_{max} 		&= \max_{i \in \{ 1, \cdots, n \} } (\max_{t \in \{ 1, \cdots, T_i \}}(\nabla L^{\Pi_s}_{t}(\XC_i)),\nonumber \\
\nabla L_{min} 		&= \min_{i\in\{1, \cdots, n\}}(\min_{t\in\{1, \cdots, T_i\}}(\nabla L^{\Pi_s}_{t}(\XC_i)),\\
\nabla L_{range} 	&= \nabla L_{max}-\nabla L_{min}, \nonumber
\end{align} 
where \(T_i\) is the time length of trial \(\XC_i\) and \(\nabla L^{\Pi_s}_{t}(\XC_i)\) is the forward gradient output by model $\Pi_s$ at time $t$ computed using time series $x_i$. Then, use the following test to trigger an anomaly for Y:
\begin{align}
	\nabla L^{m}_{t}(Y)<\nabla L_{min}-\frac{\nabla L_{range}}{2}.
\end{align}
The metric was shown to yield accurate, robust (precision and recall), and fast anomaly identification, even in recovery stages. Fig. \ref{fig:gradient_comparison_between_success_and_anomaly_trials} illustrates the identification performance of the forward gradient approach. Information regarding, parameters values, models, and training and testing are presented in Sec. \ref{sec:experiments}, whilst anomaly Identification results are found in Exp. 1.
\subsection{Anomaly Classification} \label{subsec:anomaly_classification}
The anomaly classification service is triggered once an anomaly is identified. A system can possibly address a wide variety of types of anomalies including low-level hardware anomalies: sensor and actuator noise or breakage; mid-level software contingencies like: logic errors or run-time exceptions; high-level misrepresentations: poor modeling of the robot, the world, their interactions, or external disturbances in the environment \citep*{2005BotAut-Petterson-ExecutionMonitoringInRoboticsSurvey}). In \citep*{2017IROS-Park-MultiModalAnomalyClassifFeeding}, Park \et identified both the anomaly class and the cause. In this work, we deal with anomalies caused by external disturbances generated either by intrusive human behavior or resulting from poor modeling or anticipatory ability on the robot's end. As introduced in Sec. \ref{subsubsec:external_disturbances}, four anomaly classes emerged in the cataloging experiments of the kitting task: (accidental) human-collisions (HC) in a shared-workplace; tool collisions (TC) with adjacent objects in the collection bin or the environment; object slips (OS) caused by inertia or external disturbances; and the unexpected movement of objects that led to missed grasps; otherwise described as ``No Object'' (NO). Compared to anomaly identification, classification is a more challenging problem as one must, not only trigger a binary flag, but have a multi-class classifier affected by unique dynamics of anomalous events: (i) the conditions under which individual anomalies occur can experience a diverse set of dynamics: collisions can happen at different locations, in different directions, and with difference forces. (ii)  anomalies may trigger subsequent anomalous events, for example, an HC may trigger an OS. The system must handle the onset of two temporally-near anomalies making it challenging to discern, and (iii) classification becomes increasingly complex as more adaptation nodes occur downstream since the amount of variations in experienced sensory-motor signatures, poses, and physical interactions increase (the implications for recovery are further discussed in Sec. \ref{sec:recovery}).

Just as with anomaly identification, the sHDP-VAR-HMM was used. Given $M$ trained models for $M$ robot skills, 3-fold cross validation is used along with the standard forward-backward algorithm to compute the \textit{expected cumulative likelihood} of a sequence of observations within the analysis windows (our standard is $\pm$ 2 secs.) $\mathbb{E} \left[ log \mbox{ } P(\XC_i \given \Pi_m) \right]$ for each trained model $m \in M$. Given a test trial $x$, the cumulative log-likelihood is computed for test trial observations conditioned on all available trained skill model parameters $log \mbox{ } P(x_{m_1:m_t} \given \Pi)_m^M$ at a rate of 200Hz. The process is repeated when a new skill is started. Given the phase in the manipulation graph $m_c$, we can index the correct log-likelihood $\mathbb{I}(\Pi_m=m_c)$ and see if its probability density of the test trial given the correct model is greater than the rest for the last observation point: 
\begin{eqnarray}
	log \mbox{ } P(x_{m_1:m_t} \given \Pi_{correct}) > log \mbox{ } P(x_{m_1:m_t} \given \Pi_m), \\ \nonumber
	\quad \forall m(m \in M \land m \neq m_c).
    \label{eqtn:state_classification_condition}
\end{eqnarray}
Further information regarding, parameters values, models, and training and testing are presented in Sec. \ref{sec:experiments}. Anomaly classification results are detailed in Exp. 2.
\section{Anomaly Recovery} \label{sec:recovery}
After classification, the recovery critic implements recovery through re-enacting or adaptive policies as shown in Fig. \ref{fig:framework}. Re-enacting policies re-execute a skill (possibly the current skill or a previous skill) as designated by the  policy (Sec. \ref{subsec:revertive}). Adaptive policies resolve persistent errors by training adaptive skills that leverage human understanding into the complex set of world-object-robot relations (see Sec. \ref{subsec:adaptive}). The recovery critic runs, not only during all normal phases of the task, but also significantly, during recoveries of anomalous events. To illustrate, refer to Fig. \ref{fig:framework}, where it is seen that for node \textit{move\_z}, a persistent anomaly \textit{anomaly\_k} led to the creation of an adaptive skill found in node $rec\_mv_z\_anom_k$. Then, during the execution of this adaptive skill, a new persistent anomaly \textit{anomaly\_m} entered the system. Our framework identifies it and assigns a new adaptation encoded in node $rec\_rec\_mv_z\_anom_k$ that enables the system to reach the next milestone.

\begin{figure*}     
	\centering		
		\includegraphics[width=\linewidth]{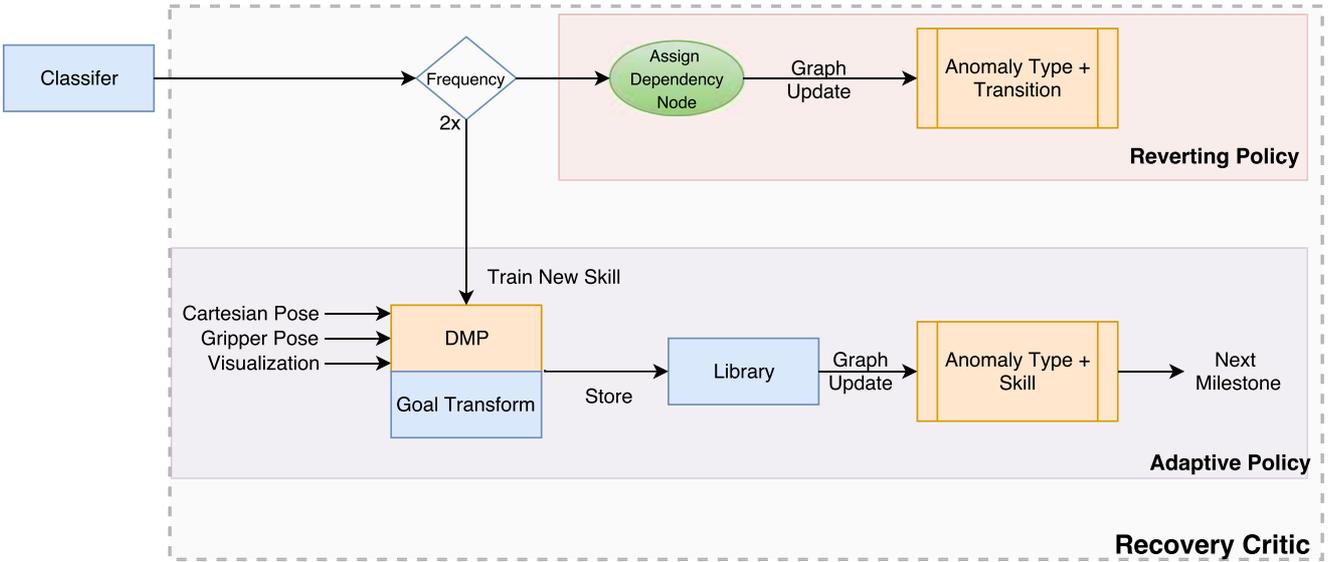}
		\caption{After classification, the recovery critic triggers a re-enacting or an adaptive policy according to the nature of the anomaly: persistent or one-off (accidental). Re-enacting policies model human decision making probabilistically (Sec. \ref{subsec:revertive}). Adaptive policies train a new skill and transform goal to reach a next phase in the task. The new skill is stored and the task graph updated (Sec. \ref{subsec:adaptive}).}
        \label{fig:recovery_flow_chart} 
\end{figure*}

Implemented recoveries, whether re-enacting or adaptive, are strictly coupled to the specific anomalies (or anomaly labels) that caused them. Recoveries themselves are globally unique and thus emerge contextually in the task (not so with anomalies). To illustrate, consider that the same anomaly may show up at different points in a task, e.g. a tool collision may happen as we try to pick an object; as we move to the packaging box; or as we place the object in the box. However, the recoveries associated with these anomalies are unique. That is, the recovery skill needed when experiencing a collision during the pick phase may be different from the one used when hitting the box (it may be possible that the same recoveries skills repeat, but we have not  explicitly studied how to leverage repeated recoveries in this work). An overview of the recovery framework is summarized in Fig. \ref{fig:recovery_flow_chart}.
\subsection{Re-Enacting Policies}\label{subsec:revertive}
Re-Enacting policies resolve accidental one-off anomalies. All anomalies are considered accidental by default, and only when they cannot be resolved through re-enactment are they considered persistent. The premise is that accidental events are resolved through the re-enactment of re-parameterized skills. The key question is to identify which skill needs be re-enacted? A few works have used a policy where either the entire task is repeated from the beginning or fixed points in the task are selected \textit{a priori} \citep*{2013IROS-Nakamura-ErrorRecTaskStrat,2018ROMAN-Wu-RecovExtDist_StateDep,2011IROS-Rodriguez-AbortRetry}. In this work, we learn more efficient skill selection mechanisms. 

Given a current milestone $\NC_j$, for each new accidental anomaly $\FC_y$, a new re-enactment (transition) $\RC_R$ is inserted into the graph as follows:
\begin{equation}
 \RC_R: \TC_{\NC_j,\NC_j^{*}} | \FC_y,\NC_j
\end{equation}
\noindent where $*$ is the target node and it is selected according to the policy introduced in Sec. \ref{subsubsection:re-enactment_polilcy}.

In the kitting experiment, consider an object slip anomaly during node 3 when the robot is moving towards the box. Instead of returning home, the robot can re-enact a re-parameterized version of the pick skill. Fig. \ref{fig:framework} illustrates the concept, consider node \textit{move\_y}. When \textit{anomaly\_j} occurs, the recovery critic assigns re-enactment $rec\_mv_y\_anom_j$ which transitions to the previous pick node. Or, back in the kitting experiment, consider an accidental human collision that bumps the robot arm whilst executing the \textit{move\_to\_box} skill. Provided built-in safety procedures, once the temporary accidental contact concludes, the robot could re-enact the current skill. Note that nodes contain skills that are inherently reactive. The starting and goal poses of a skill can be set without altering the skill's properties. A re-enactment of the current skill with a re-parameterized starting pose would be enough to complete that task phase and reach the next  milestone. Fig. \ref{fig:framework} also illustrates the concept for \textit{move\_y} and anomaly \textit{anomaly\_l}. The critic here assigns re-enactment $rec\_mv_y\_anom_l$ which is a self-transition. 
In effect, re-enactment goal nodes are chosen \textbf{in relation to the nature of the anomaly type}.

\subsubsection{The Re-Enactment Policy}\label{subsubsection:re-enactment_polilcy}
Re-enactment goal nodes are assigned through multinomial distributions that model human-user goal node selections given a current node and a specific anomaly. Five human users studying a robotics master's degree were trained to understand the graph topology of the task, possible transitions, skill  execution, goal parameterization, anomaly types, and legal node selections/transitions for re-enactment. Each user examined 5 trials of induced anomalies on a per-node, per-anomaly basis, yielding independent multinomial distributions to determine re-enactment policies. For instance, if at node 2, three anomaly types occur, then there will be three multinomial distributions modeling the policy. For each multinomial, let $\bm{N}=(N_1,...,N_K)$ be a random vector where $N_j$ is the number of times a node $j$ is selected as a re-enactment target node. Then $\bm{N}$ has the following pmf:
\begin{equation}
	\mathrm{Mu} (\bm{N}|n,\theta) =
	\left( \substack{n\\  N_1...N_K} \right) 
    \prod_{j=1}^{K} \theta_j^{N_j}
\end{equation}
\noindent
where, $\theta_j$ is the probability that node $N_j$ is selected. The results are shown in Table \ref{tbl:multinomial_counts}. 
\begin{table}[tb]
\centering
\caption{Human user selections of re-enactment target nodes given a current node execution and a specific anomaly.}
\label{tbl:multinomial_counts}
\begin{tabular}{cccc}
\toprule 
Anomaly & Source Node & Target Node & Count \\ 
\midrule
\multirow{4}{*}{HC}     & Node 1 & Node 1 & 25         \\
        				& Node 2 & Node 2 & 30         \\
        				& Node 3 & Node 3 & 25         \\
                        & Node 4 & Node 4 & 25         \\
\midrule
\multirow{2}{*}{TC}     & Node 2 & Node 1 & 25         \\
        				& Node 3 & Node 3 & 5          \\
\midrule        
\multirow{3}{*}{OS}     & \multirow{2}{*}{Node 2} 	& Node 2 & 20         \\
        				&        					& Node 1 &  5         \\
        				& Node 3 					& Node 2 & 25         \\
\midrule        
\multirow{2}{*}{NO}     & \multirow{2}{*}{Node 2} 	& Node 2 & 24         \\
				        &        					& Node 1 & 1          \\                       
\end{tabular}
\end{table}
The multinomial provides an indirect way to represent human intuition about the complex set of relations that exist between the robot (and its limbs), the relevant objects of the task at hand, and the interactions that the robot and the objects have with the world. Additionally, the multinomial also encode a person's internal belief about the utility of a choice, his/her own learning ability (within a trial and across trials), and the person's risk propensity or aversion in decision making\endnote{Utility theory seeks to model complex decision-making processes using notions of expected value, expected utility, risk and learning models, and more \citep*{1998NAP-Book-ModelBeh}}. For instance, OSs that occurred during the picking skill (node 2), were assigned two different types of re-enactment target nodes: to re-execute the same pick skill with 80\% probability and to execute the previous move-to-pick (node 1) with 20\% probability. The choice of returning to node 1 represents a more conservative belief or risk averse selection on the user's part. 
\subsubsection{Re-Enactment Target Nodes}\label{subsubsec:re-enactment target nodes}
Goals for re-enacted target nodes are set by the visualization module. The starting pose is simply the current pose at the time of anomaly, while the goal pose is set as originally described in Sec. \ref{subsec:goal_setting}. 

\subsubsection{Training Re-Enactments}\label{subsubsec:training_re-enactments}
Re-enactment policies designed in this section were trained during the cataloging experiments of Sec. \ref{subsec:overview_anomaly_collection}. Success rates for re-enactment policies given accidental anomalies are reported in Exp.'s 3-6 under a variety of different conditions. 

\subsection{Adaptive Policies}\label{subsec:adaptive}
Adaptive policies are used to resolve persistent anomalies. 
Persistent anomalies are classified as such when a re-enactment policy fails to resolve a given anomaly twice consecutively. This phenomena indicates that re-enactment is unable to solve the condition and that the task requires explicit adjustments to finish the task successfully.
In this work, we rely on human intuition and expertise to provide the necessary adaptation skill to solve the persistent task anomaly. 

\subsubsection{Kinesthetic Teaching}\label{subsubsection:adaptive_kinesth_teach}
Our system is designed to pause automatically when two consecutive re-enactment policies occur for the same node-anomaly pair in the graph. The system then awaits for the user to initiate kinesthetic teaching (through the push of a system button) and encode the adaptive skill. The system also, at this time, records all relevant sensory-motor data necessary (until the end of kinesthetic teaching) to train a new introspection model for the current nominal (adaptive) skill.

\subsubsection{Graph Integration}\label{subsubsection:adaptive_graph_integration}
Given a current milestone $\NC_i$, for each new persistent anomaly $\FC_x$, a new adaptive recovery node $\RC_A:\NC_{ij}$ is inserted into the graph as a new branch in-between milestones, where the target node transition $*$ is inherited from the parent node in the graph in accordance to Eqtn. \ref{eqtn:adaptive_recovery}.
\begin{gather}
 \label{eqtn:adaptive_recovery}
 \RC_A: S_{\NC_i}| \FC_x \rightarrow \NC_{ij}\\
 T_{\NC_{i},\NC_{ij},\NC_{j}}  							\nonumber
\end{gather}
Fig. \ref{fig:framework} illustrates the concept, consider how in node $move\_z$, persistent anomaly $anomaly\_k$ is resolved using adaptive skill $rec\_mv_z\_anom_k$ as a new branch between milestones $move\_z$ and $place$.

For cases in which an anomaly $\FC_{xx}$ occurs during an adaptive node $\NC_{ij}$, a new adaptive node is created in a new branching layer:
\begin{gather}
 \RC_A: S_{\NC_ij}| \FC_{xx} \rightarrow \NC_{ijk}\\
 T_{N_{ij}, N_{ijk}, N_{ik}}. 							\nonumber
\end{gather}
Branches always transition to the ensuing milestone, not matter the branching level. In this work, we have assumed that a single adaptive skill is sufficient to restore the nominal functioning of the task. It is plausible to sequence skills to achieve more complex manipulations.

\subsubsection{Setting Adaptive Node Goals}
As described in Sec. \ref{subsec:goal_setting}, skill goals are set by the Visualization module of a node. However, for adaptations, when human users introduce additional manipulation, they are also introducing a transformation on the goal pose of the parent skill with respect to the base frame. Adaptive skills then compute the transformation of the last time step in kinesthetic teaching with respect to the goal of the parent node.  During online testing, the Visualization module computes the real-time goal of the parent node, whilst the adaptive skill transforms that goal to achieve task generalization during adaptation.

\subsubsection{Inheriting Re-Enactment Policies}
Whenever we push a new adaptive node into the graph, that adaptive node is set to inherit the same re-enactment policies available to its predecessor. This is important so as to avoid the need to re-train re-enactments in new adaptation nodes.

\subsubsection{Training}
Cataloging experiments were used to capture sufficient data to create robust nominal skill introspection models for adaptive anomalies. These models are then used by our Anomaly identification routine in Sec. \ref{subsec:identification}, to identify anomalies that may occur during such adaptations. Anomaly Identification performance is presented in Exp. 1, whilst the success rates for adaptive policies presented in this section are reported in Exp.'s 4-6 under a variety of different conditions that elucidate system performance. 
\section{Experiments and Results} \label{sec:experiments}
\newcounter{Ex} 
\stepcounter{Ex} 
Seven experiments are setup to test the accuracy, robustness, and reactivity of anomaly identification and anomaly classification as well as the efficacy and versatility of our recovery policies under different situations. Exp. 1 \& 2, present accuracy and robustness results for Anomaly Identification\stepcounter{Ex} and Anomaly Classification respectively. Exp. 3-6 examine the recovery policy efficacy and versatility. Exp. 3 measures the robustness of re-enacting recovery policies. Exp. 4 tests the robustness of adaptation policies. Exp. 5 analyzes the robustness when both recovery policies co-exist in the same task. Exp. 6 tests the system's ability to introspect anomalies and recover from them whilst the system is executing an undergoing a recovery action due to a previous anomaly in the system. Finally, Exp. 7 analyzes the reactivity of our anomaly classification algorithm.
\subsection{Kitting Experiment Setup}
As stated in Sec. \ref{subsec:experimental_setup}, the Baxter robot is set-up to perform a kitting experiment in conjunction with a human co-worker. The human is responsible for placing objects in the collection bin and the robot is responsible for the packaging. The space is shared between the robot and the human is shared rendering it possible for the human to provoke anomalies in the system: including both accidental and persistent anomalies. 

Three computers are used to run the experiment: Baxter's internal computer, which runs Gentoo Base System 2.2 and an Intel(R) Core i7-3770 CPU@3.40GHz, 4GB-RAM, x64-based processor. The internal computer is used to run a ROS joint trajectory server as well as the camera on the left arm. The other two computers run Linux 14.04 with ROS Indigo. One computer has an Intel(R) Core i5-3470 CPU@3.20GHZ, 6GB-RAM, x64-based processor and runs alvar recognition, the moveit service, and time-series pre-processing for all sensory-motor data. The second workstation, runs an Intel Xeon i7-6820HQ CPU@2.70GHz(3.60GHz Turbo), 8MB-RMA, x64-based processor and is in charge of running anomaly identification and anomaly classification online which is implemented with BNPY \citep*{bnpy}, with a ROS-wrapper. 

Our graph implementation uses a hybrid approach. Base nodes for the kitting experiment are currently implemented through ROS-SMACH. The non-adaptive nodes however are designed through an internal procedural representation which is detailed in Appendix A. Diagrammatic representations and code are accessible through our supplementary materials page \citep*{2018IJRR-Wu-RecoveryFramework_supplementalURL}.
\subsection{Human Subject Training}
In Exp.'s 3-6, five different human subjects, under consent, took part in the experiment as human collaborators. They were trained to place consumer goods, one-at-a-time, in the collection bin of the robot. We ask human subjects to assume they are multi-tasking and experiencing loss of attention. The loss of attention can lead (as discovered by the cataloging experiments in Sec. \ref{subsec:overview_anomaly_collection}) to a number of anomalous events including: (i) HCs, (ii) TCs, (iii) OSs, and (IV) NOs---wall collisions (WC) are introduced in Exp. 4 but these result not from human induction but from different object shape properties. HCs may occur when the robot picks up objects from the collection bin and the human collaborator places new ones. TCs may occur when humans inadvertently place objects near each other such that when the robot attempts to pick an object, one of its fingers collides with the adjacent object (see Fig. \ref{fig:adapt_fingers}). OSs may occur after human collisions that rattle the gripper and cause heavier or smoother objects to fall. NO anomalies may occur when a human accidentally collides or removes an object that the robot intended to pick up. 
\subsection{Signal Processing}\label{subsec:signal_processing}
Regarding the signals used in these experiments, we originally considered a 6 DoF end-effector twist and wrench respectively, a 7 DoF pose (using quaternions as orientation), and 56 taxel values (each finger has a 4-by-7 grid). A variety of human-engineered pre-processing techniques were tested for these signals. The final selection of pre-processing features for these signals was decided during the validation stage of experimentation and will be reported individually for Anomaly Identification and Classification in Exp.'s 1 and 2 respectively. 

All signals were scaled, resampled, and aligned.
With respect to scaling, signals were modified to lie in a range of $-1\leq y_i \leq 1$ by computing the absolute value of the maximum signals during training. 
With respect to resampling, given that different observation signals have different publishing rates (wrench: 1000Hz, tactile: 1000Hz, pose and twist: 100Hz) a re-sampling rate is used to acquire a single time-point at which to model the observations. Our code relies primarily on python and ROS. Rospy nodes inherently use Python's multi-threading class to handle multiple publishers and subscribers. The class, however, lacks real-time performance support and we have only achieved re-sampling rates of up to 50Hz. 
Alignment takes places by syncing the timestamps from the varying ROS topics. 
\subsection{sHDP-AR-HMM Parameters \& Hyperparameters} 
Given that both anomaly identification and classification are based on the same model, we present a base-model to introduce parameter settings that are broadly shared across the methodologies. Whenever particular differences exist from the base-model, they will be explained within specific experiments. 

For the observation model, we use a first-order vector autoregressive with regression matrix coefficients $\boldsymbol{A}$ and covariance matrix $\boldsymbol{\Sigma}$ for specific latent states. Since both of these dynamic parameters are uncertain, they need to be learned. The MNIW is an appropriate prior distribution when both the mean and the covariance are uncertain \citep*{2015NIPS-Hughes-ScalableAdapStateComplex_npHMM}. 

We begin by determining the covariance $\Sigma$ through the use of the IW distribution $NIW$. For this computation, we must define the first moment of the distribution according to Eqtn. \ref{eqtn:iw_first_moment}. Here, we set $\nu$, the degrees of freedom, is set to to the sum of the number of dimensions + 2: $\nu = d + 2$. This setting ensures the conjugate MNIW prior has a valid mean (see Sec. 4.5.1 in \citep*{2012MIT-Murphy-ML_ProbPerspective}). As for the computation of the expectation of the covariance in Eqtn. \ref{eqtn:expectation_of_covariance}, the scalar $s_F$ is set to $1.0$ and multiplied by the scatter matrix (also the empirical covariance). This setting is motivated by the fact that the covariance is computed from polling all of the data and it tends to overestimate latent-state-specific covariances. A value slightly less than or equal to 1 of the constant in the scatter matrix mitigates the overestimation. 

Then, to determine the matrix $\boldsymbol{A}$ of regression coefficients, the matrix normal of the MNIW uses a mean matrix $\boldsymbol{M}$ set to the zeros matrix $\boldsymbol{M} = \mathbf{0}_d$, of size $d \times d$. We do so to let the new observation be primarily be determined by the signal noise. 

For the covariance $\boldsymbol{K}$ across the columns an identity matrix is used such that $\boldsymbol{K} = 1.0*\mathbf{I}_d$ with the same dimension as $\Sigma.$

For the concentration parameter $\alpha$ of the HDP prior, a $Gamma(a, b)$ distribution with values $a = 0.5, b = 5$ is used. For the self-transition parameter $\mu$ a weakly informative $Beta(c,d)$ prior distribution is used with values $c=1, d=10$. 

For the sticky HMM transition distribution, another $\kappa$ (the degree of self-transition bias)  is set to 50. The number of maximum iterations for the Split-Merge Monte Carlo method is set to 1000. Finally, the truncation (maximum) number for latent states is empirically set to $K=10$ for both anomaly identification and classification. 
\subsection{Classification Modalities}
As part of Exp.'s 3-6, we present success rate metrics as a function of two distinct classification system modalities: 
\begin{enumerate}[label=\roman*]
\item perfect anomaly classification (independent system)
\item imperfect classification (combined system)
\end{enumerate}
The perfect anomaly classification modality implies that recoveries are only attempted when true positives classifications are produced by the system. In doing so, we can treat the entire system as three independent sub-systems: anomaly identification (AD), anomaly classification (AC), and the recovery (REC) system. By separating the sub-systems we can study their effectiveness independently from the other systems. The imperfect classification modality on the other hand studies the success rates of recoveries in the presence of misclassifications. This leads us to treating the entire system as a function of two subsystems: AD and AC/REC. Such separation let's us study some interesting phenomena that emerged from the REC system and is detailed in each of the experiments.
\newcounter{Experiment} 
\stepcounter{Experiment} 
\subsection*{Experiment \theExperiment: Anomaly Identification}\label{subsec:anomaly_id_exp}
In Exp. 1, we evaluate the performance of the anomaly identification system across the entire set of experiments. Specific context analysis will be presented within Exp.'s 3-6. We have expanded our previous work on anomaly identification by learning to flag anomalies caused by a larger number of classes. A larger class set (including new skills that are learned through adaptive policies) implies more challenging accuracy, precision, and recall performance in the system. Furthermore, since the anomaly identification system is the first to be triggered, it is critical that identification is done accurately; otherwise the system will suffer increasingly from upstream errors. In this section we present the identification accuracy of the system as well as the robustness through accuracy, Precision and Recall metrics.

The anomaly identification system used the sHDP-VAR-HMM technique (Sec. \ref{subsec:shdp-var-hmm}) to create class models for both the original nominal skills introduced in Sec. \ref{subsec:experimental_setup} (we will call these non-adaptive nodes), but also and very importantly for new adaptation skills that are learned when persistent anomalies take place (we will call these adaptive nodes). In particular, the adaptive skills of Exp. 4a,b,c, and 6a,b. Once the nominal models are trained, the forward gradient measure (Sec. \ref{subsec:identification}) is used for anomaly identification.
Upon the collection of offline data for training from the inducing experiments described in Sec. \ref{subsec:overview_anomaly_collection}, a scoring heuristic was implemented over 5-fold cross-validation that allowed us to select from a variety of hand-engineered multi-modal signal features and parameter values. Different combination of features were tested for specific sets of parameter values. Scoring in the form of accuracy, precision, and recall metrics was computed for each combination. The highest scoring model was selected. The highest score resulted in the following combination:
\begin{itemize}
\item End-effector force $F$, torque $\tau$, linear velocity $\nu$, and angular velocity $\omega$ such that: $[F, \tau, \nu, \omega] \in \mathbb{R}^3$.
\item $l^2$-norm of the above signals; namely: $[|F|, |\tau|, |\nu|, |\omega|] \in \mathbb{R}^1$.
\item The maximum standard deviation $\sigma$ computed for each of the 28 taxels in a tactile sensor for the left and right fingers; namely, $\max_{\sigma} [\sigma_{l}, \sigma_{r}] \in \mathbb{R}^1$.
\end{itemize}

To build anomaly identification models for both non-adaptive and adaptive skills, a fixed number of 7 trials was used. Non-adaptive skills consisted of the move-to-pick, pick, move-to-box, and place skills and adaptive skills are those captured in Exp.'s 4a,b,c and 6a,b respectively. 

Macro accuracy, precision, and recall metrics are extracted by testing whether we can identify anomalies (HCs, TCs, OSs, NOs, or WCs) given some domain (nodes or sub-experiments).

\begin{figure}[t]
	\centering
        \includegraphics[width=\linewidth]{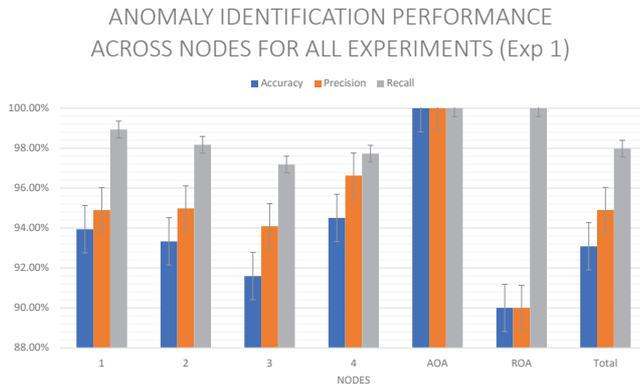}
    \caption{Summary of accuracy, precision, and recall metrics for anomaly identification across all experiments on a per-node basis, including recovery over recovery runs in Exp. 6, and a total summary of performance.} 
    \label{fig:exp1_AD_summary_all_exps}
\end{figure}

In this section, we present a summary of the results for anomaly identification for Exp.'s 3-6 (Exp. 2 presents anomaly classification results). Fig. \ref{fig:exp1_AD_summary_all_exps} charts the summary across nodes 1-4 as well as new adaptive nodes that are particularly generated when anomalies occur during recoveries as seen in Exp. 6. In Exp. 6, we analyzed two scenarios: Adaptations over Adaptations (AOA) and Re-enactments over Adaptations (ROA) which are discussed in detail there. All results and their analysis can be found in Extension 3. 
\\ \\ \textbf{Results}\\
Our anomaly identification accuracy for the totality of all experimental (766) trials was of 93.09\% (see Extension 3 for details). The precision was 94.09\% and the recall 97.98\%. These results show very strong accuracy and performance which is critical to avoid the aforementioned downstream errors. In terms of performance across nodes, the experiments revealed very similar performance throughout the task with an average accuracy of 93.34\%. This implies that anomaly identification performance did not improve or decline as the manipulation graph traversed the nodes- rendering the identification consistent and reliable. The system also showed perfect accuracy and robustness for occasions in which persistent anomalies occurred during recoveries (AOA-Exp. 6). For times where accidental anomalies occurred during recoveries (ROA) the accuracy and precision was strong at 90\% with no false-negatives. 
\stepcounter{Experiment}
\subsection*{Experiment \theExperiment: Anomaly Classification}\label{subsec:anomaly_orig_classif_exp}
\begin{figure}[t]
	\centering
        \includegraphics[width=\linewidth]{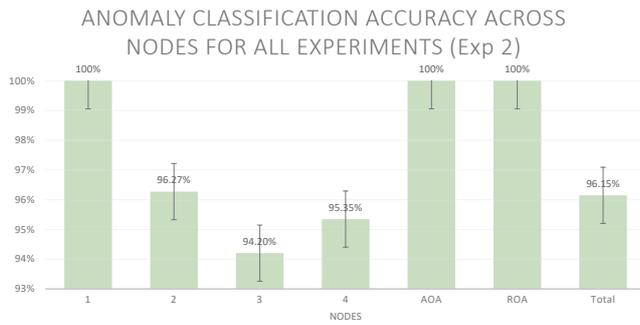}
    \caption{Anomaly classification accuracy for all experiments across nodes 1-4 as well as adaptive nodes AOA and ROA created for anomalies under executing recoveries.} 
    \label{fig:exp2_AC_summary_all_exps}
\end{figure}

\begin{figure}[b]
	\centering
        \includegraphics[width=\linewidth]{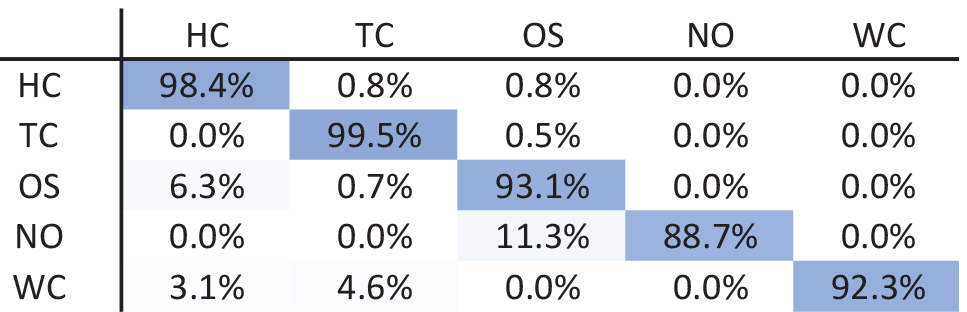}
    \caption{Confusion matrix computed in Exp. 2 for all occurring anomaly classes in the Kitting experiment across all experiments.} 
    \label{fig:exp2_AC_confusion_summary_all_exps}
\end{figure}

\begin{figure}[b]
	\centering
        \includegraphics[width=\linewidth]{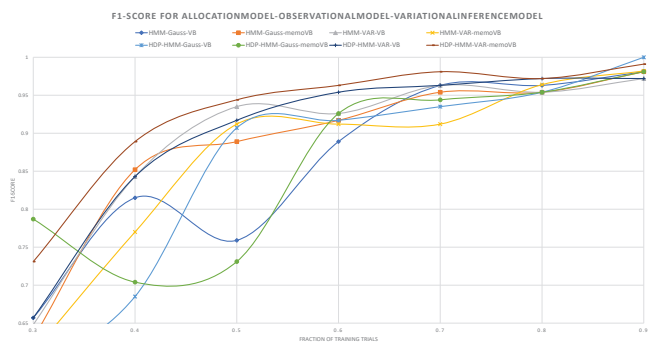}
    \caption{F1-score metric performance comparison for variational inference algorithms across allocation and observation models.} 
    \label{fig:exp2_var_inf_comp}
\end{figure}

After anomaly identification, it is important to understand the performance and robustness of the anomaly classifier.
The anomaly classification also uses the sHDP-VAR-HMM with memoized variational inference (Sec. \ref{subsec:shdp-var-hmm}) along with the same features and training style used in anomaly identification. The model is trained to classify anomalies caused by human collisions (HCs), tool collisions (TCs), object slips (OSs), no objects (NOs), and wall collisions (WCs) introduced in Exp. 4. For training, we used the following number of trials for the aforementioned classes: HC-18, TC-17, OS-18, NO-15, and WC-17. We have not yet implemented an unsupervised learning method that automatically generate new anomaly labels based on previously unseen data (determined through a confidence metric), but we have contemplated this work (see Sec. \ref{sec:discussion}). Anomaly classification is only triggered if anomaly identification experiences a true-positive. Once the classification procedure is called no true-negatives or false-negatives exist in the system. Only true or false positives. For this reason, classification will be measured in terms of accuracy across nodes or confusion matrices for a particular experiment. In Exp. 2, we present a summary of the corresponding information for Exp.'s 3-6. Anomaly classification accuracy across nodes (including the AOA and ROA nodes introduced in Exp 1) is presented in Fig. \ref{fig:exp2_AC_summary_all_exps}. 
A confusion matrix was also computed for classification for all experiments and shown as a figure in Fig. \ref{fig:exp2_AC_confusion_summary_all_exps}. 
Furthermore, we used the F1-score metric to compare the performance variational inference algorithms across allocation and observation models. The models used for this comparison are listed bellow:
\begin{itemize}[noitemsep]
  \item Variational Inference Models: Memoized Variational Inference with Scalable Adaptation (MemoVB) and Variational Coordinate Ascent (VB). 
  \item Allocation Models: HMM and sHDP-HMM
  \item Observation Models: Gaussian (Gauss) and the Vector-Autoregressive (VAR)
\end{itemize}
As for the variational inference algorithms, we compare the algorithm used in this paper; namely, memorized variational inference with scalable adaptation with variational coordinate ascent under different allocation and observation models. Stochastic variational inference was contemplated but not used as the algorithm did not converge after 1000 iterations. Gibbs sampling was also not used as it was not available as part of online BNPY \citep*{bnpy}. The comparisons are also conducted as a function of the number of total training trials. The same number of total training trials was used as  mentioned at the beginning of this experiment. Fig. \ref{fig:exp2_var_inf_comp} shows the comparative performance of the inference methods. 
\\ \\\textbf{Results}\\
Our anomaly classification accuracy for the totality of all experimental (719 trials) data was of 96.15\% (see Extension 3 for details). Interestingly, the accuracy of our anomaly classifier was overall more accurate than our anomaly identification routine. Extensive experimentation has been carried out. General trends are reported here, whilst specific experimental details are presented within each experimental section. For the non-adaptive nodes, node 1 had perfect classification accuracy. Nodes 2-4 ranged from 94.20\% to 96.27\%. This indicates very similar performance over task-time and that the classifier was robust in detecting a varying range of challenges (see each experiment for specific details). The performance during already executing recovery actions was of 100\%. Although the number of trials for this section was 19, the data suggests strong classification performance even as the robot is adapting to anomalies. In terms of the confusion matrix in Fig. \ref{fig:exp2_AC_confusion_summary_all_exps}, accuracy ranged from 88.7\% to 98.4\% for NO and HC respectively. The 2nd poorest classification was that of WC. WC were more challenging as the collision sometimes occurred against the gripper but in other occasions against the held object. OS came next with 93.1\%, OS classification was challenged primarily by the tactile sensor noise experienced and explained later in Exp. 3. 

With regards to variational inference performance, Fig. \ref{fig:exp2_var_inf_comp} shows how the sHDP-VAR-HMM with Memoized Variational Inference with Scalable Adaptation generally outperformed the rest of the combinations except for a couple of instances. In fact, in around 88.3\% of the fraction of training trials our algorithm outperformed all others. The exceptions occurred roughly for the fraction 0.3-0.33 of the total training trials, where the sHDP-HMM-Gauss-MemoVB initially outperformed our algorithm 0.787 to 0.731. Similarly, for the fraction 0.87-0.90 of the total training trials, the the sHDP-HMM-Gauss-VB outperformed our algorithm by 0.9\%. Note that results will vary slightly across experimental runs as trial data is selected randomly and the probabilistic framework we is unable to fix the random seed value across runs.  
\stepcounter{Experiment}
\subsection*{Experiment \theExperiment: Testing Re-enactment}\label{subsec:exp_rev}
\begin{figure*}[tb]
	\centering
    \subfigure{
    	\label{fig:exp3_AD_Nodes}
        \includegraphics[width=0.475\linewidth]{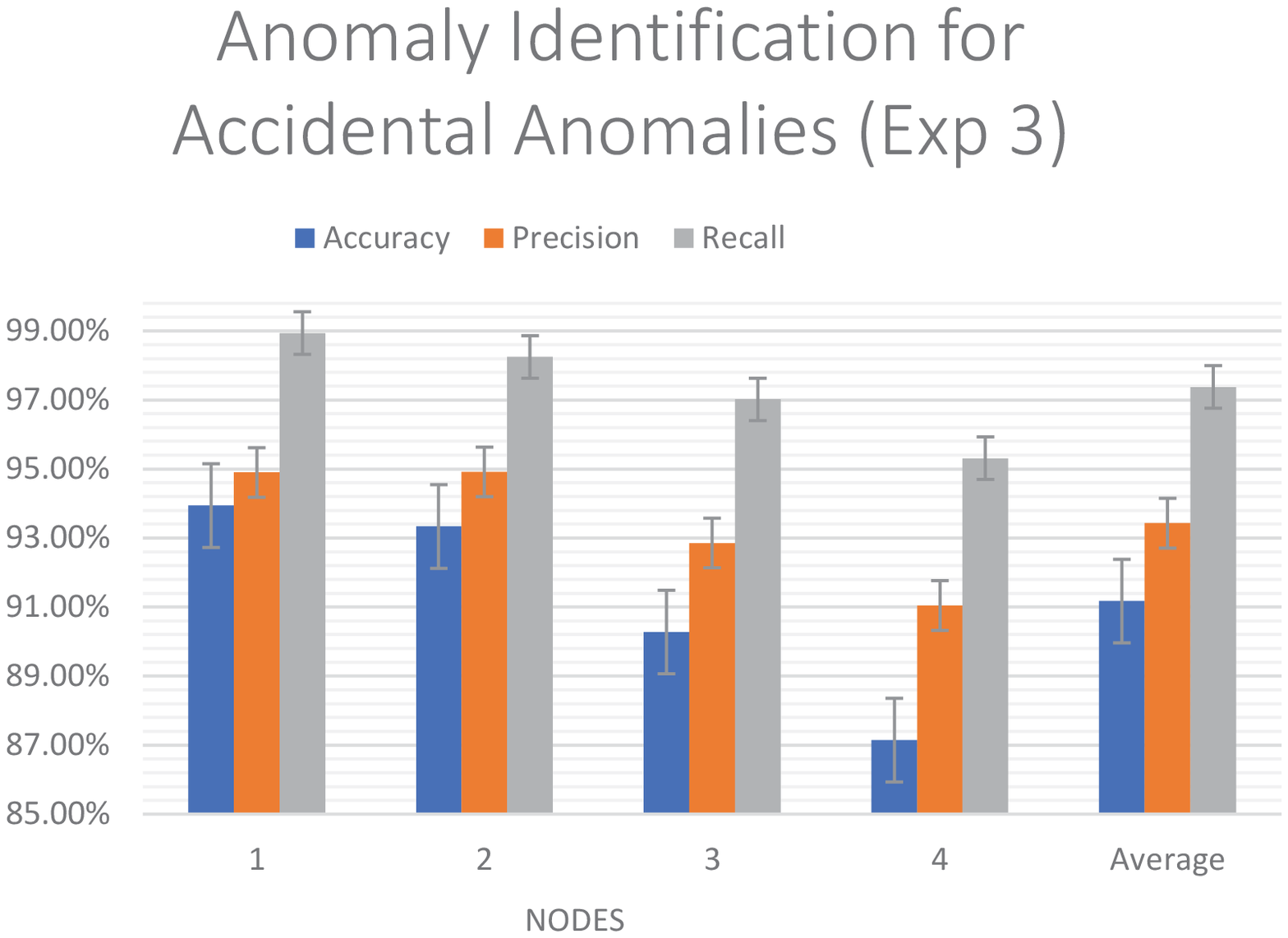}}
    \hspace{0.01cm}
    \subfigure{
    	\label{fig:exp3_AC_Nodes}
        \includegraphics[width=0.475\linewidth]{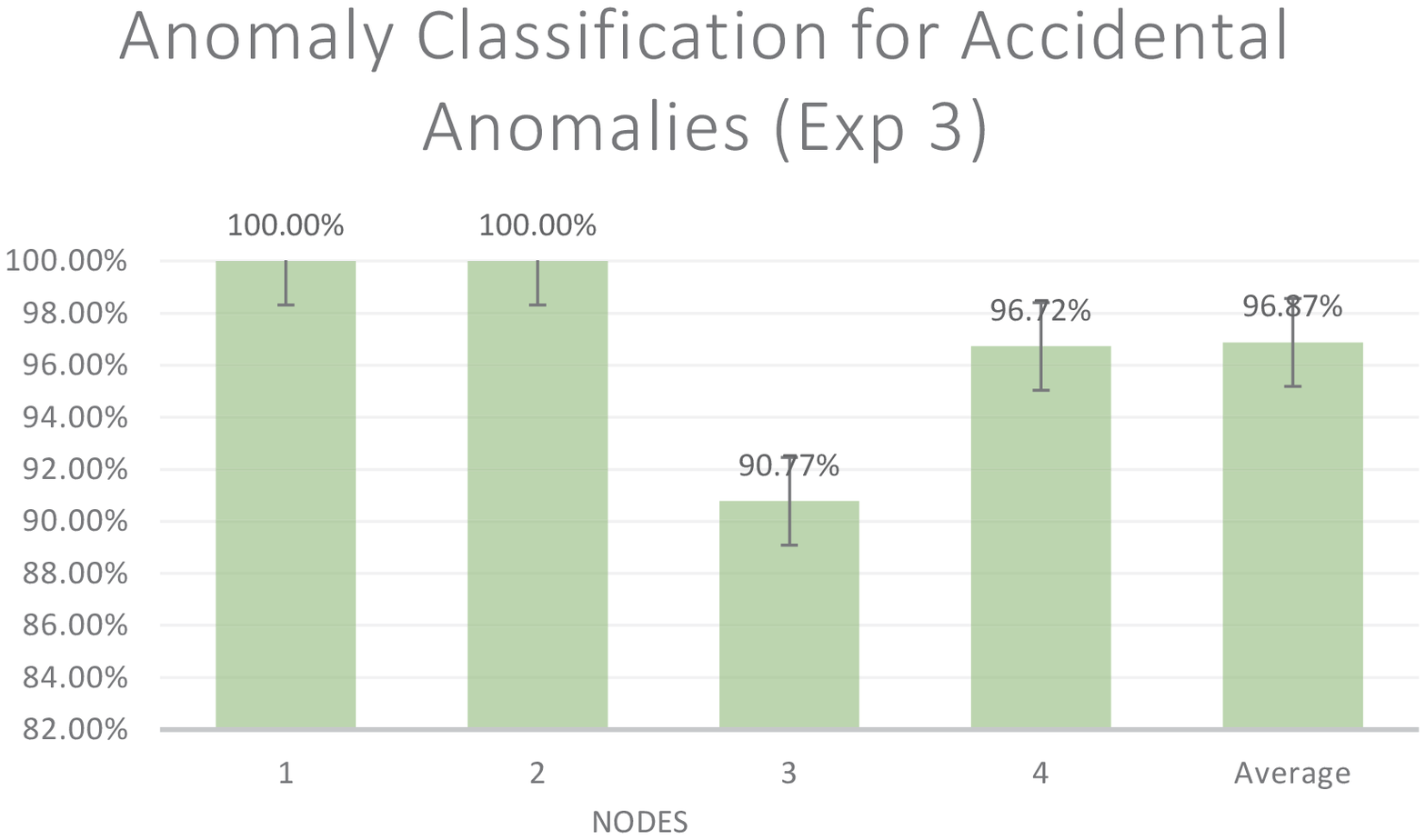}} 
    \caption{Accuracy, precision, and recall metrics for the anomaly identification and accuracy metrics for the anomaly classification system on a per-node basis for accidental anomalies (left and right respectively).} 
    \label{fig:exp3_acc_prec_recall}
\end{figure*}

\begin{figure}[bt]
	\centering
         \includegraphics[width=\linewidth]{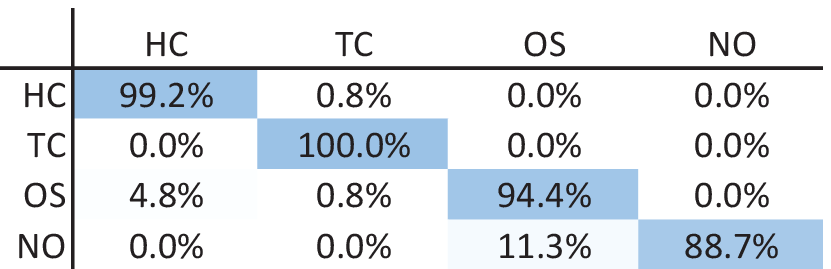}
    \caption{An anomaly classification confusion matrix for accidental anomalies in Exp. 3.}
    \label{fig:exp3_confusion}
\end{figure}

Experiment \theExperiment $\mbox{}$ analyzes the accuracy and robustness of the anomaly identification, anomaly classification and recovery critic for accidental anomalies. We study the recovery critic's ability to re-enact reliably at different phases of the task. To this end, accidental anomalies were induced at specific graph phases as listed below:
\begin{itemize}[noitemsep]
\item[] Node 1: HC 
\item[] Node 2: HC, TC, OS, NO
\item[] Node 3: HC, OS
\item[] Node 4: HC
\end{itemize}
The results for anomaly identification and anomaly classification for Exp. 3 are shown in Figs. \ref{fig:exp3_AD_Nodes} \& \ref{fig:exp3_AC_Nodes}.  
A confusion matrix for classification accuracy is shown in Fig. \ref{fig:exp3_confusion}. 

For the re-enactment recovery system, 60 recoveries were attempted (10 trials per object for 6 objects and induced by 5 trained users) on a per-node basis (4 total) under our two classification modalities: (i) perfect classification and (ii) imperfect classification.

The result of the re-enactment policy for modality (i) is shown in Table \ref{tbl:exp1_SuccRate-Ind} and for modality (ii) in Table \ref{tbl:exp1_SuccRate-Combined}.
\begin{table}[tbh]
\centering
  \caption{Recovery Success Rate with Perfect Classification---modality (i) across nodes and anomaly classes.}
  \label{tbl:exp1_SuccRate-Ind}
    \begin{tabular}{rccccc}
    \toprule
    Node                                             & HC       & TC       	& OS       	& NO       & Average  \\
    \midrule
    1                                                & 100\% 	&          	&          	&          	& 100\% \\
    2                                                & 100\%    & 100\%		& 100\% 	& 100\% 	& 100\% \\
    3                                                & 100\% 	&          	& 90.0\%  	&          	& 95.0\%  \\
    4                                                & 100\% 	&          	&          	&         	 & 100\% \\
    \midrule
    Total                                            &          &          	&          	&          	& 98.75\%  \\ 
    \end{tabular}
\end{table}
\begin{table}[tbh]
\centering
  \caption{Recovery Success Rate with Imperfect Classification---modality (ii) across nodes and anomaly classes.}
  \label{tbl:exp1_SuccRate-Combined}
    \begin{tabular}{rccccc}
    \toprule
    Node  & HC      & TC      	& OS      	& NO      & Average \\
    \midrule
    1     & 95.00\% &         	&         	&         & 95.0\% \\
    2     & 85.00\% & 98.3\%	& 88.3\% 	& 80.00\% & 87.9\% \\
    3     & 91.7\% &         	& 95.0\% 	&         & 93.3\% \\
    4     & 95.0\% &         	&         	&         & 95.0\% \\
    \midrule
    Total &         &         	&         	&         & 92.8\%
    \end{tabular}
\end{table}\\ 
\\ \textbf{Results}\\
For anomaly identification, a total of 574 trials were used for testing (103, 265, 144, and 62 for nodes 1 to 4). An average accuracy of 91.16\%; a maximum of 93.94\% and a minimum of 87.14\% in nodes 1 and 4 respectively. For precision we had an average of 93.42\%; a maximum of 94.90\% and a minimum of 91.04\% in nodes 1 and 4 respectively. For recall we had an average of 97.37\% with a maximum of 98.94\% and minimum of 95.31\% in nodes 1 and 4 respectively. The reason node 4 may experience lower robustness might be due to the fact that more variations exist over time (\eg poses may vary in ways that modify the previously experienced dynamics during training).

For anomaly classification, a total of 516 trials were used for testing (93, 242, 122, and 59 for nodes 1 to 4)with an average accuracy of 96.87\%. Nodes 1 and 2 were classified perfectly, followed by 4, and struggled the most with node 3 at an accuracy of 90.77\%.
The confusion matrix for anomaly classes shows perfect or near perfect classification for TC and HC respectively and struggled more with OS and NO. OS detection suffered primarily form noise in our tactile sensor. We believe a large portion of the noise came from false contacts in the electronics in the tactile sensor. Whilst we attempted to rigidly fix the sensor's electronics, there was still wiggle during anomalous events. With regards to NOs, we were surprised with the lower classification rate. We believe that the tactile sensor's noise was also the culprit. We wanted to use the infra-red sensor on the robot's wrist as an additional observation source, however, the force-torque sensor set-up blocked the IR signal and prevented its use.

With regards to re-enactment recovery, we present success rates for both classification modalities. Under perfect classification, we re-enacted and completed the task successfully on average 98.75\% across all nodes (see Table \ref{tbl:exp1_SuccRate-Ind}). Some failures occurred in Node 3 as an OS occurred. After the OS, the object reached a location outside the field of view of the camera and prevented the system from computing the object pose. We should note that there were 11 other trials where system failures occurred (these were not marked as recovery failures). There were two main causes for the system failures: (a) challenging pick poses resulted in tactile sensor cables constraining the gripper and (b) an electricity overload in the system that rendered parts of the robot to a halt. 

Under imperfect classification, we expected a lower performance, and obtained an average recovery completion of 92.81\% across all nodes (see Table \ref{tbl:exp1_SuccRate-Combined}). The highest rates were obtained in node 1 and 4 under HC anomalies with 95\%, recovery, TC anomalies in node 2 with 98.3\% recovery, and OS anomalies in node 3 at 95\% recovery. The picking skill was the most problematic to resolve in the presence of HCs and NOs. 

With regards to overall system trends we observe: very competitive anomaly detection at an average of 91.16\% and very high anomaly classification (one of our contributions) at 96.87\%. For re-enactment under independent systems we see that re-enactments can resolve almost all accidental anomalies at 98.75\%. 

\begin{figure}[tb]
	\centering
         \includegraphics[width=\linewidth]{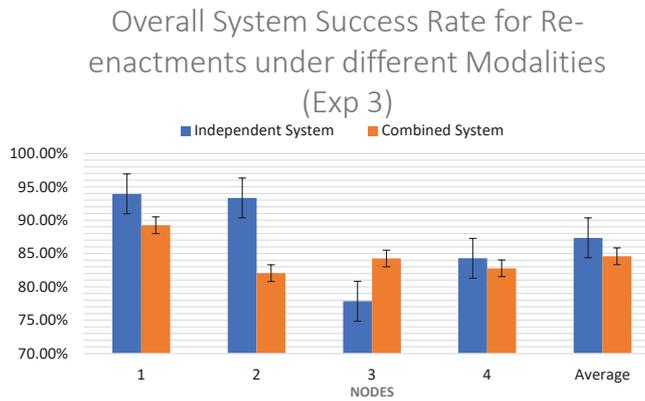}
    \caption{Overall system success rate as a function of modality. Modality (i) considers the contributions of three independent systems: identification (AD), classification (AC), and recovery (REC). Modality (ii) considers the contributions of an independent AD system with a combined AC/REC system. The figure shows an interesting phenomena during node 3: the 2nd modality performed better than the 1st as wrong classifications were corrected downstream and coupled with correct recovery policies.}
    \label{fig:exp3_overall_success_rate}
\end{figure}

One last but very interesting development was evident when we computed the performance of the \textit{entire system} under the two classification modalities as seen in Fig. \ref{fig:exp3_overall_success_rate}. 
Note, interestingly, that for one out of the four nodes---node 3---the overall success rate of the combined system was higher than that of the independent system. This implies that system completed the task successfully more times under imperfect classification than with perfect classification. The specific reason for this phenomena is that soon after a misclassification takes place; the introspection system detects that the robot is still in an anomalous state and triggers a new anomaly flag and issues a new round of classification. This time the correct policy is issued and resolves the anomalous situation. One example is when an OS was misclassified as an HC. The HC triggers a re-enactment, but the robot is not grasping the object. At a later time step, the introspection system flags another anomaly and classifies it as an NO. This time a pick re-enactment is issued and enables the robot to successfully complete the task. 
\stepcounter{Experiment}
\subsection*{Experiment \theExperiment: Testing Adaptation}\label{subsec:exp_adapt_simple}
Experiment \theExperiment $\mbox{ }$ analyzes the robustness of the anomaly identification, classification, and adaptive recovery policy in the face of persistent anomalies. We analyze adaptation robustness by testing three scenarios with an increasing number of persistent anomalies (and thus adaptations). The three sub-experiments test robustness under the following conditions:
\begin{itemize}[noitemsep]
\item[] \theExperiment a: one adaptation at a single phase (two examples).
\item[] \theExperiment b: a 2nd adaptation introduced at a new phase.
\item[] \theExperiment c: a 3rd adaptation introduced at a new phase.
\end{itemize}
For this experiment we run a total of 20 trials per persistent anomaly (4 objects with 5 trial runs per anomaly). A new anomaly class---Wall Collision---was discovered in these experiments and labeled (WC). We analyze whether adaptive policies work robustly independent of the number of adaptations that occur previously in the system and also whether or not the policies generalize across objects. Object locations and order are varied and randomized across trials. Sub-experimental details are given in three distinct sections below. Results are jointly presented and analyzed at the end of this section for succinctness. 
\\ \\
\textbf{Experiment \theExperiment a: Adaptation at Distinct Single Nodes}\label{subsubsec:exp_adapt_simple}
\\
In Experiment \theExperiment a, we analyze the robustness of the framework to properly identify, classify, and recover from persistent anomalies in single instances using adaptive recoveries. As described in Sec. \ref{subsec:adaptive}, when the same anomaly occurs twice consecutively in the same node, the anomaly is considered persistent and an adaptive skill is learned from a user demonstration to recover and transition to the succeeding milestone in the task. 

For this experiment, we tested two distinct persistent anomalies at independent phases of the task (node location is indicated by @\# followed by anomaly type):
\begin{itemize}[noitemsep]
\item [\theExperiment a.1]: @2TC (pick).
\item [\theExperiment a.2]: @3WC (move-to-box).
\end{itemize}

\begin{figure}[tb]
	\centering
    \subfigure{
    	\label{fig:block_pick}
        \includegraphics[width=0.475\linewidth]{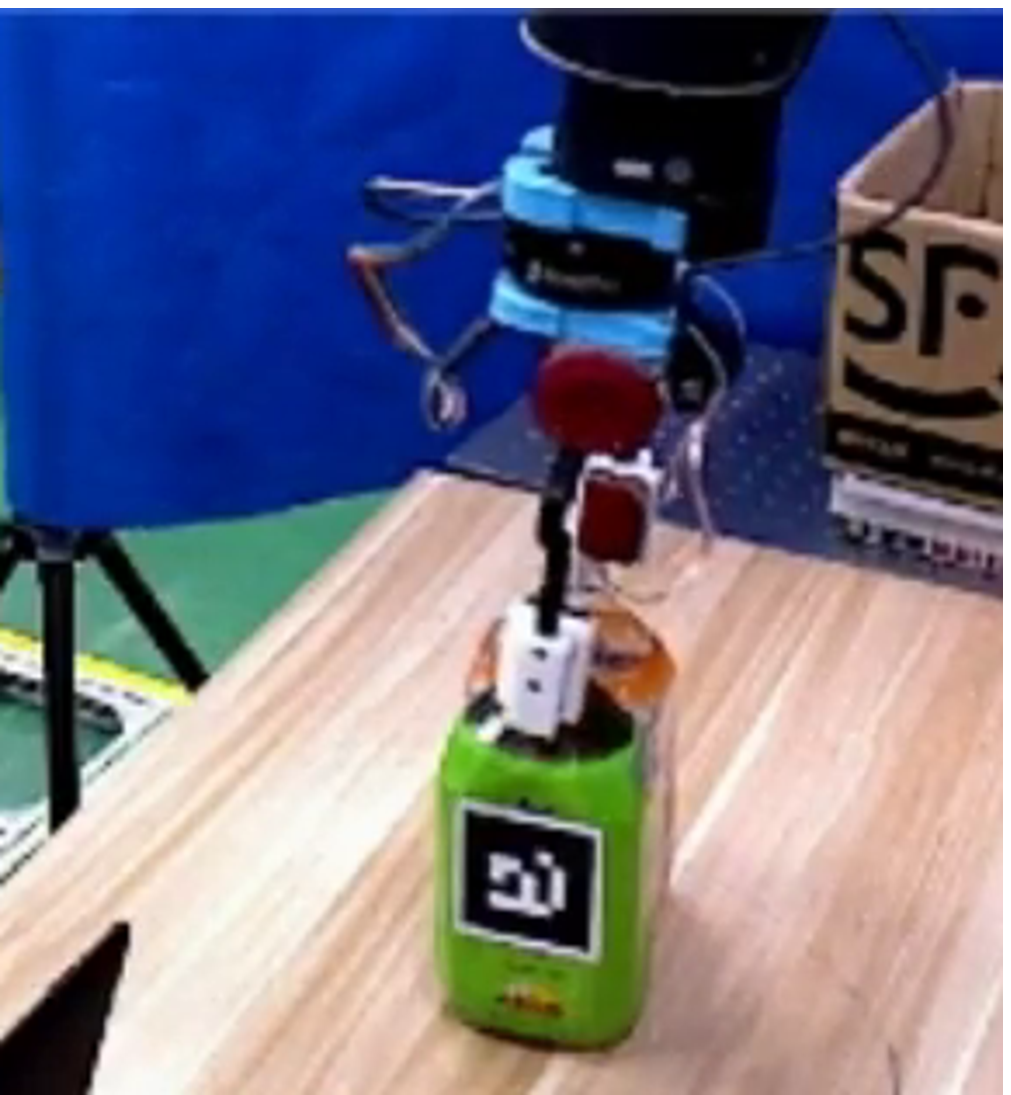}}
    \hspace{0.01cm}
    \subfigure{
    	\label{fig:adapt_fingers}
        \includegraphics[width=0.45\linewidth]{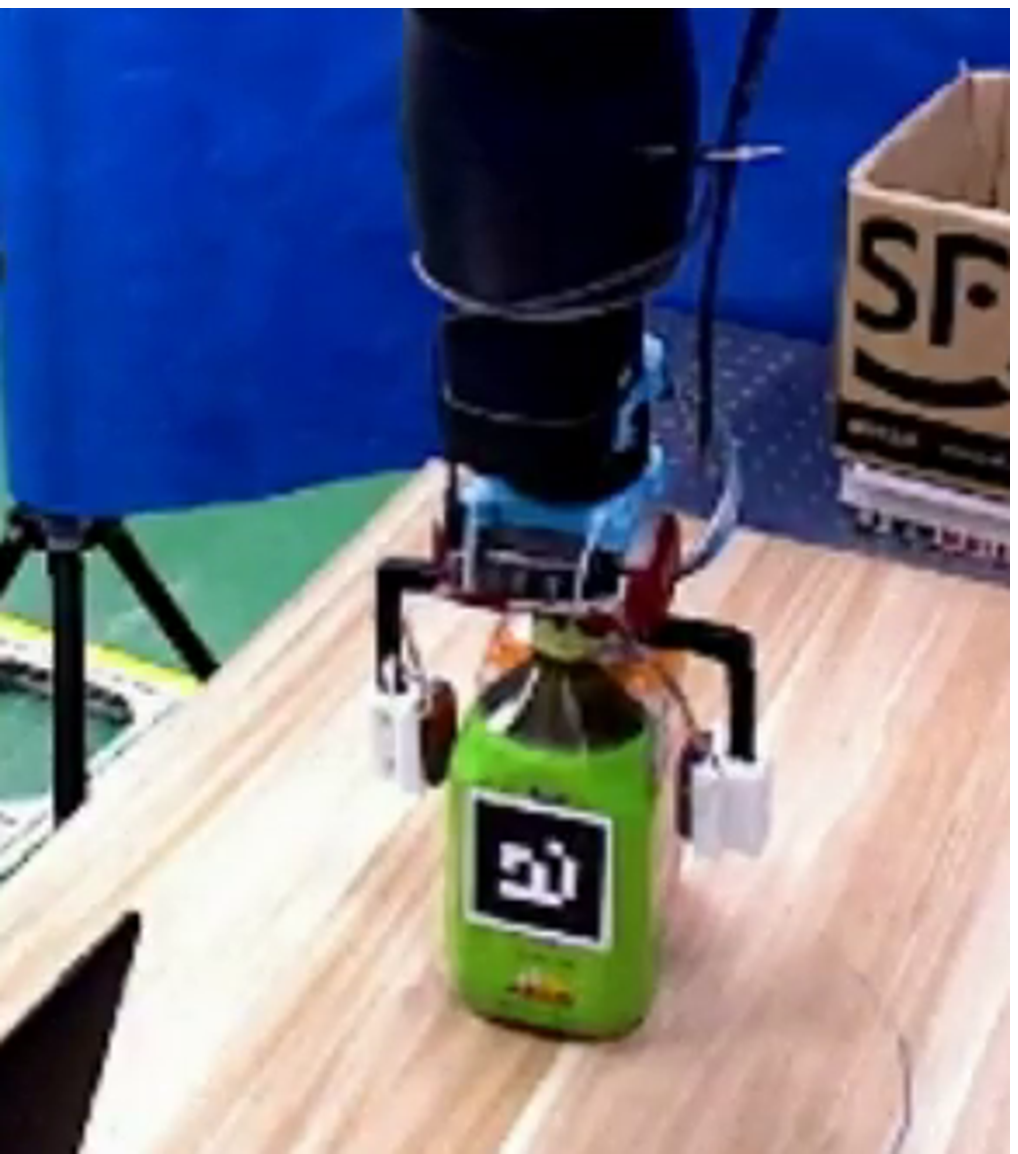}} 
    \caption{Persistent Pick Anomaly. On the left: the proximity of an adjacent object consistently precludes the proper gripping of a target object leading to a persistent tool collision. On the right: the execution of the learned adaptive skill which rotates the wrist and clears the fingers for the pick skill.}
    \label{fig:finger_collision}
\end{figure}

Tool collisions (TC): occurred when two objects were placed by a human operator too close to each other. In such conditions, when the pick skill in node 2 is executed, one of the robot's fingers collides with the neighboring object and prevents a proper pick as illustrated in Fig. \ref{fig:finger_collision}. Re-enactments do not resolve the situation so help from a user is elicited to overcome the persistent condition. The taught adaptive skill rotates the robot wrist about the approach axis and clears the fingers from the obstruction. 

\begin{figure}[tb]
	\centering
    \subfigure{
    	\label{fig:wc_001}
        \includegraphics[width=0.305\linewidth, height = 2in]{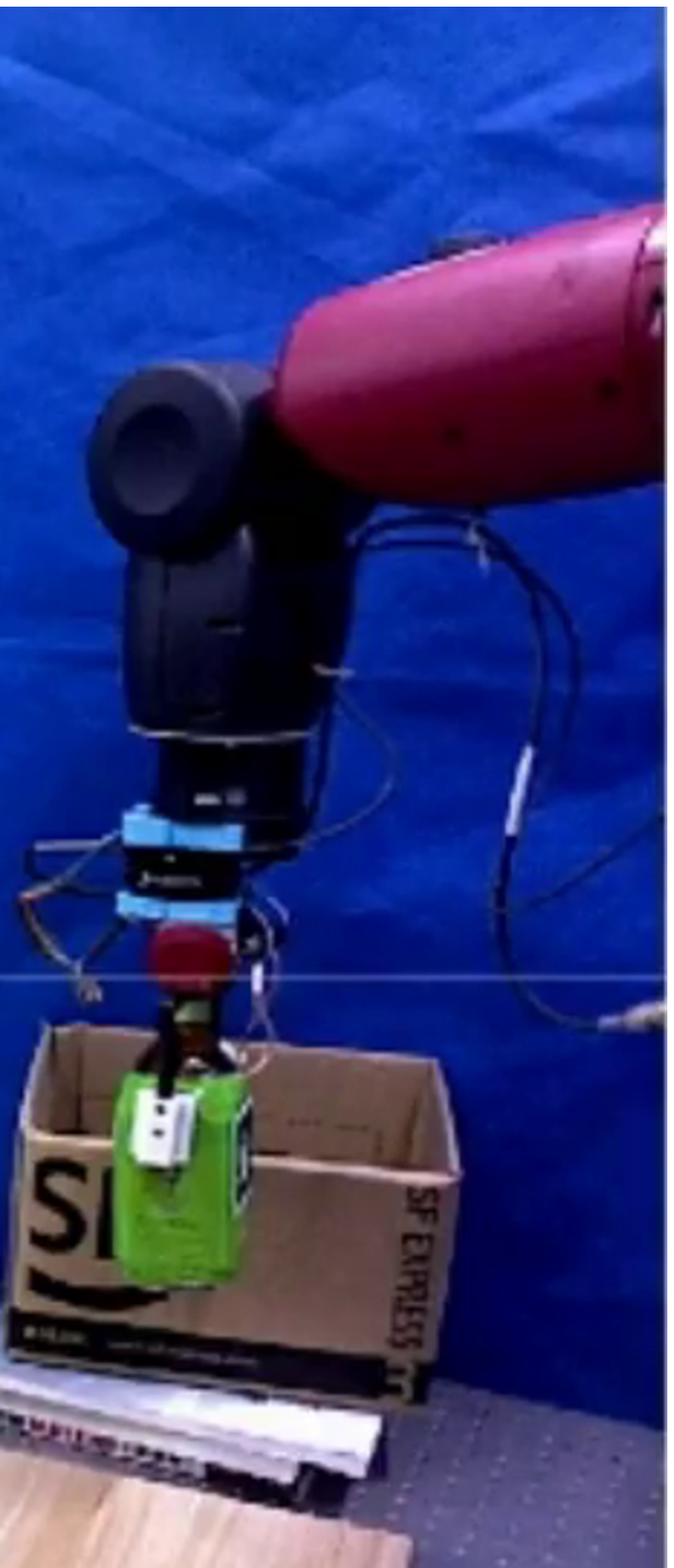}}
    \hspace{0.01cm}
    \subfigure{
    	\label{fig:wc_002}
        \includegraphics[width=0.27\linewidth, height = 2in]{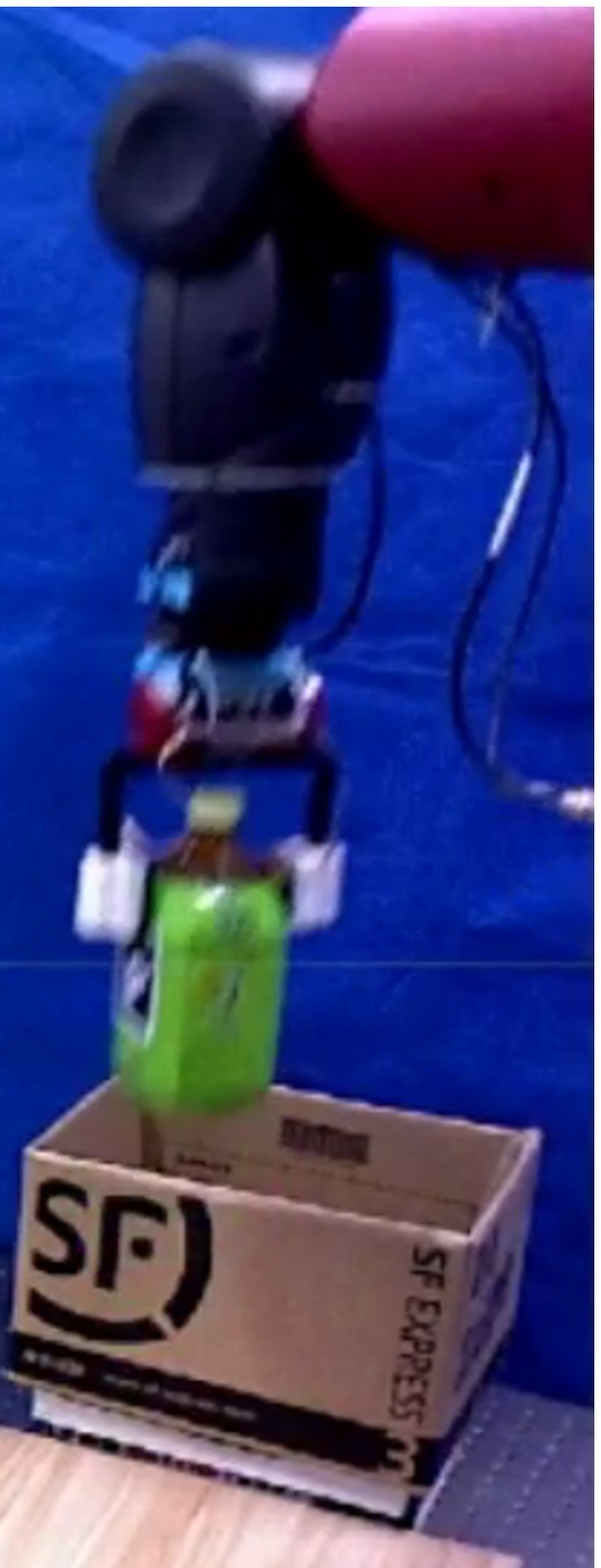}} 
    \hspace{0.01cm}
    \subfigure{
    	\label{fig:wc_003}
        \includegraphics[width=0.275\linewidth, height = 2in]{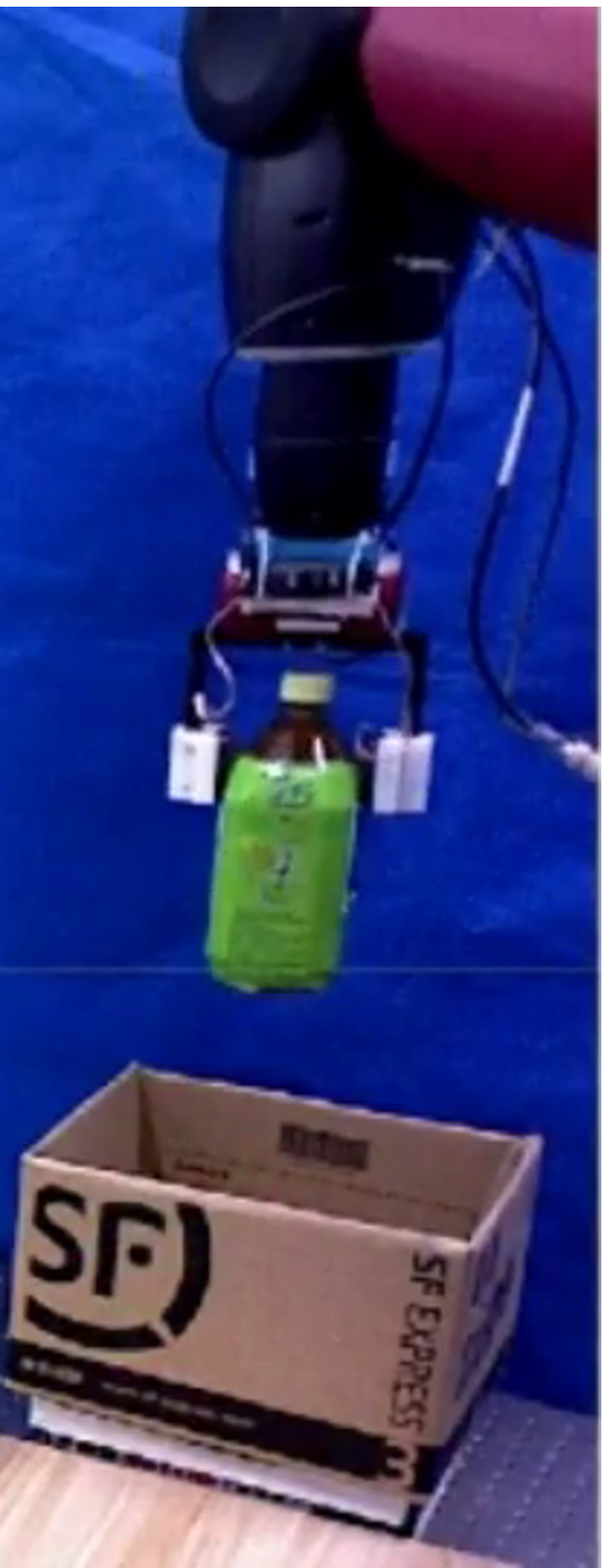}} 
    \caption{Persistent Wall Collision. From left to right: Trained adaptive skill overcomes a wall collision. The new skill executes a clearing mechanism that lifts the object above the edge of the wall before placing the item in the package.}
    \label{fig:wall_collision}
\end{figure}

\begin{figure}[b]
	\centering
    \subfigure{
    	\label{fig:bc_001}
        \includegraphics[width=0.45\linewidth]{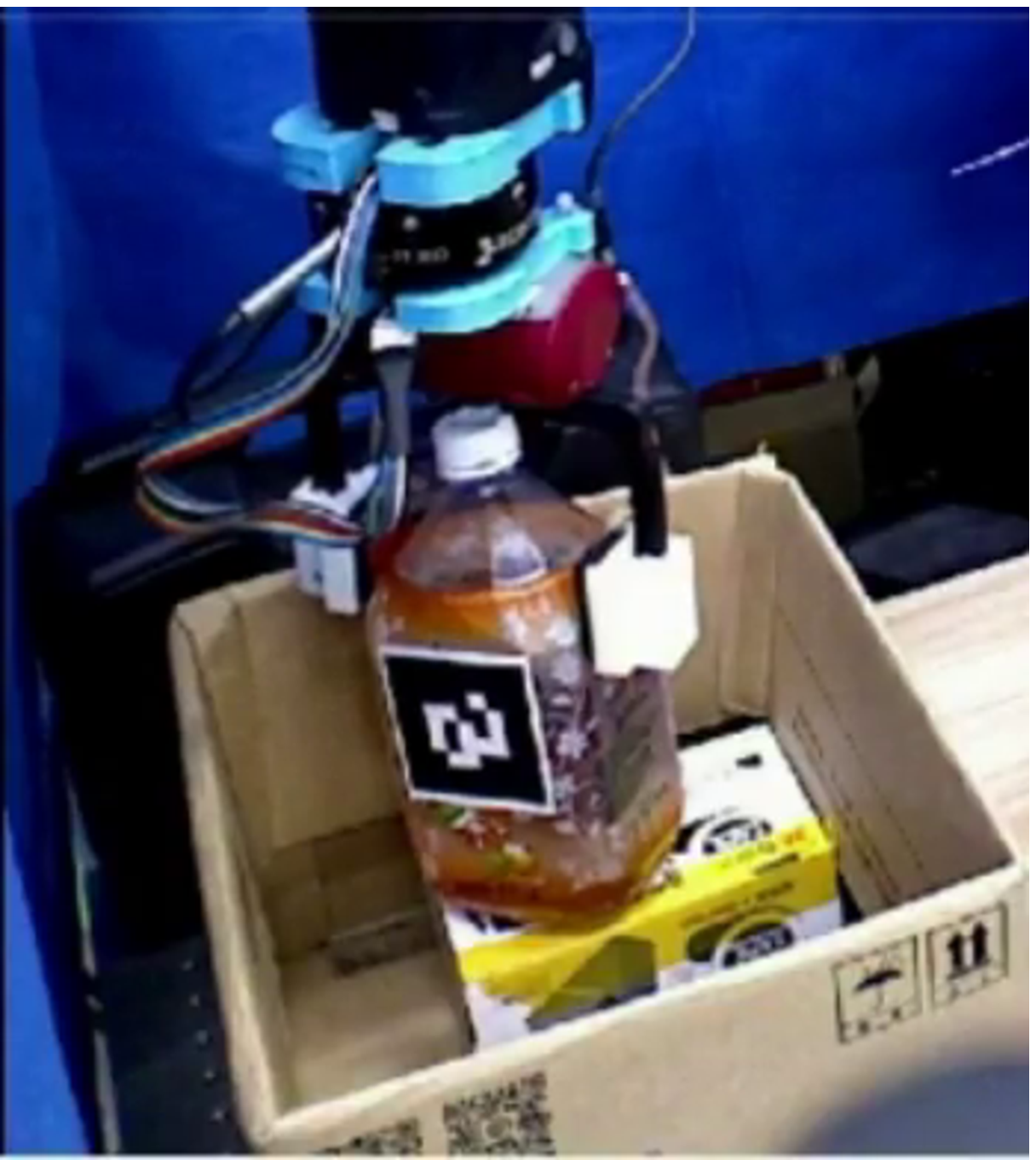}}
    \hspace{0.01cm}
    \subfigure{
    	\label{fig:bc_002}
        \includegraphics[width=0.44\linewidth]{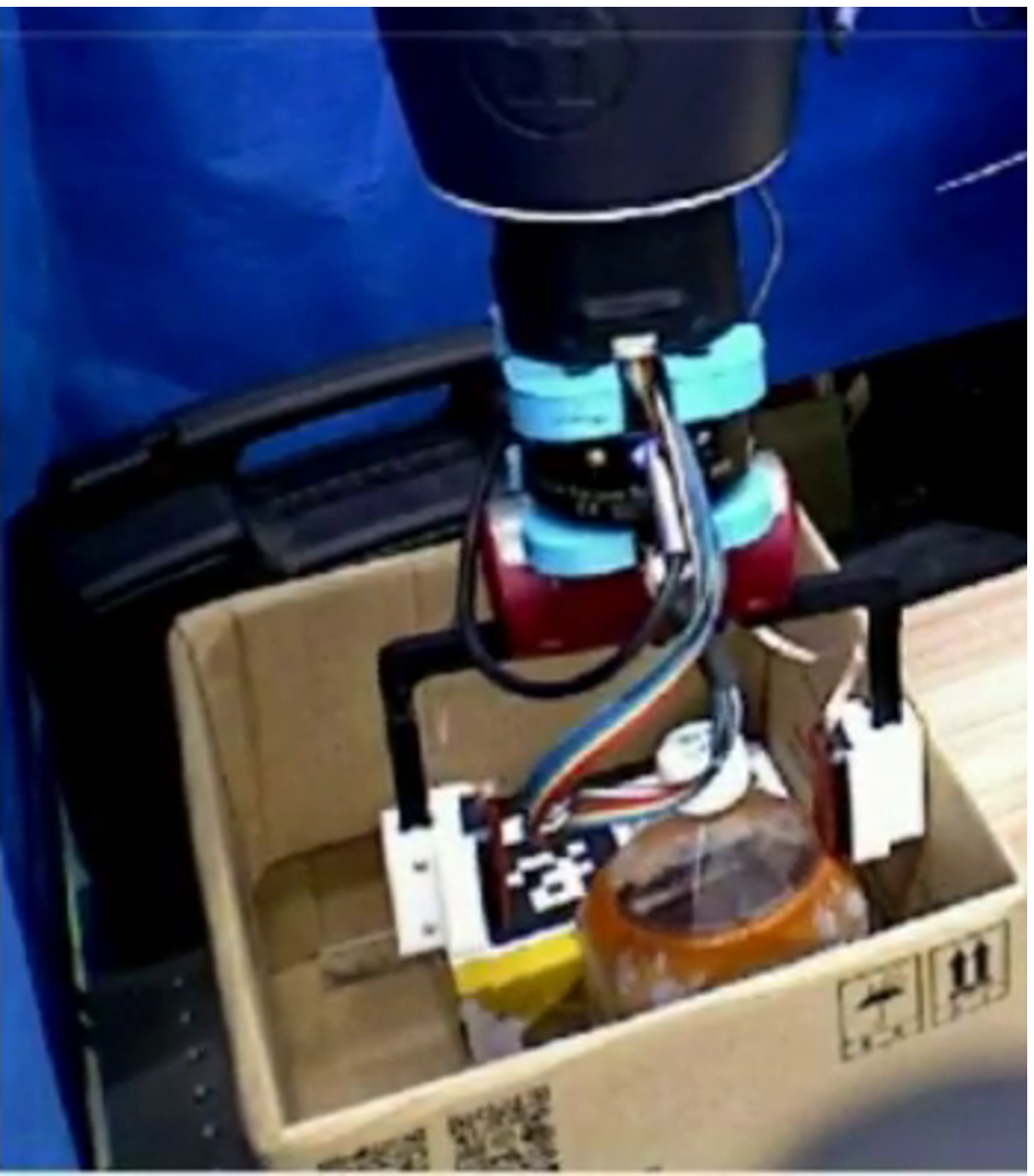}} 
    \caption{As part of the last phase of the task, the robot attempts to place an object in the package only to find an existing object at the target location. A re-enactment does not solve the anomaly, so an adaptive move is taught and a new goal provided by the visualization module.}
    \label{fig:box_placement}
\end{figure}

\begin{figure*}[tbh]
	\centering
    \subfigure{
    	\label{fig:exp4_AD}
        \includegraphics[width=0.475\linewidth]{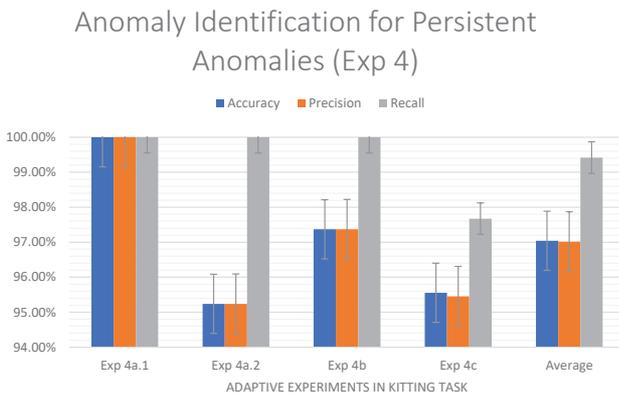}}
    \hspace{0.01cm}
    \subfigure{
    	\label{fig:exp4_AC}
        \includegraphics[width=0.475\linewidth]{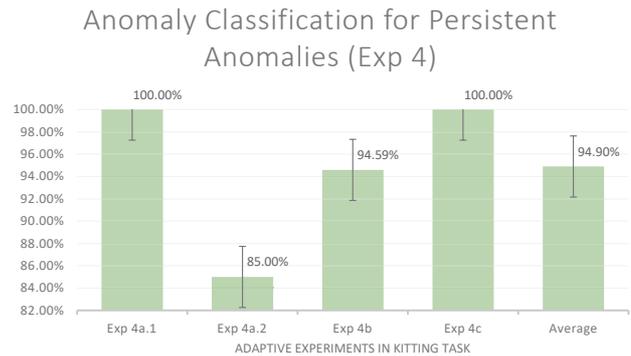}} 
    \caption{Accuracy, precision and recall metrics for the anomaly identification system and accuracy metrics for the classification system on a per-(sub)experiment basis for persistent anomalies (left and right respectively).} 
    \label{fig:exp4_prec_recall}
\end{figure*}

Wall Collisions (WC): in this (second) example, no tool collision occurs at node 2, however a persistent collision occurs at node 3 as the robot moves the picked object to the packaging box. The wall collision is a variant of  of a tool collision. Tool collisions were narrowly defined as collisions that occur on vertical downward motions. In this case, the collision occurs with a lateral motion and the contact can be either tool-wall (of the packaging box) or object-wall. The reason for such anomaly is that the original move-to-box skill was trained on an object of a given height and later, a taller object was picked and the object did not clear the wall using the original skill (see Sec. \ref{sec:discussion} for a discussion on motion adaptation based on shape properties). Re-enactment does not resolve the anomaly; so an adaptive skill which executes a clearing motion is taught. The execution is shown in Fig. \ref{fig:wall_collision}.
\\ \\ \textbf{Experiment \theExperiment b: Incremental Growth for Two Adaptations. }\label{subsubsec:exp_adapt_2}
\\
In Experiment \theExperiment b, we analyze system robustness when two adaptive skills are learned incrementally for different phases of the task. It is important to ensure that the performance of the system is not compromised as more adaptations are introduced into the task graph. In this experiment, we integrate the adaptive recoveries learned in Exp. 3a and induce both persistent anomalies in the same experiment in an incremental fashion at different phases of the task: 
\begin{itemize}
\item [\theExperiment b]: @2TC,@3WC.
\end{itemize}
In this way, the robot first responds by rotating its wrist to clear the persistent obstruction during the pick; and later upon collision with the wall, the robot responds by lifting its arm and clearing the box wall before placing the good in the package. 
\\ \\ \textbf{Experiment \theExperiment c: Incremental Growth for Three Adaptations. }\label{subsubsec:exp_adapt_3}
\\
Finally in Experiment \theExperiment c, we analyze system robustness when we integrate the third adaptation. The next persistent anomaly occurs in node 4 as the robot places an object in the packaging box. The last anomaly results when, upon executing the placing skill, an object already in the box obstructs the placement of our currently held object. So the final sequence of anomalies at varying phase locations is:
\begin{itemize}
\item [\theExperiment c]: @2TC, @3WC, @4TC.
\end{itemize}
The Visualization module is in charge of allotting unique placement goals for all objects in a box, such that they all have a unique space within the package. However, it is possible that upon placement of an object, the latter falls and shifts to a different location in the box causing a tool collision. The adaptive skill teaches a simple displacement motion whose goal is parameterized by the visualization module to a clear location. Fig. \ref{fig:box_placement} shows such process.
\\ \\
\textbf{Results}\\
We now summarize the results for Experiment \theExperiment a,b,c. 
For anomaly identification, a total of 124 trials were used for testing (20, 21, 38, and 45 for experiments a.1, a.2, b, and c). For anomaly identification, we had an average accuracy of 97.04\%, an average precision of 97.02\% and an average recall of 99.42\% across the three sub-experiments. Very strong performance was achieved all around and charted in Fig. \ref{fig:exp4_AD}. 

For anomaly classification, a total of 121 trials were used for testing (20, 20, 37, and and 44 for experiments a.1, a.2, b, and c) with an average accuracy of 94.09\%. Experiment \theExperiment a.2 had the worst performance at 85.0\%, followed by Experiment \theExperiment b at 94.59\%, and perfect classification in Experiment \theExperiment c. 

\begin{figure}[b]
	\centering
         \includegraphics[width=\linewidth]{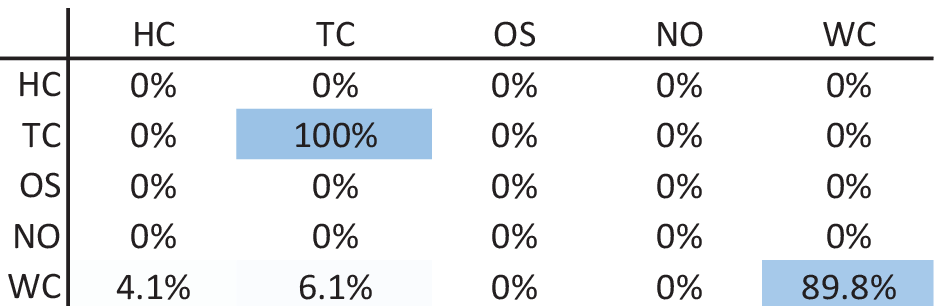}
    \caption{An anomaly classification confusion matrix for persistent anomalies TC and WC in Exp 4.}
    \label{fig:exp4_confusion}
\end{figure}

A confusion matrix was also computed for classification accuracy and shown as a figure in Fig. \ref{fig:exp4_confusion}. TC and WC are the core classes, whilst HC appears as a result of misclassification.
Across all sub-experiments we were able to identify TCs in Exp.'s \theExperiment a and \theExperiment c with 100\% accuracy. Wall collisions were slightly less accurate at 89.80\%. Wall collisions were harder to classify given that those collisions occurred under two different scenarios: at times the gripper collided with the box and at other times the held object made the collision. Hence, the multi-modal signals contained variations that degraded the classification performance.

\begin{figure}[b]
	\centering
         \includegraphics[width=\linewidth]{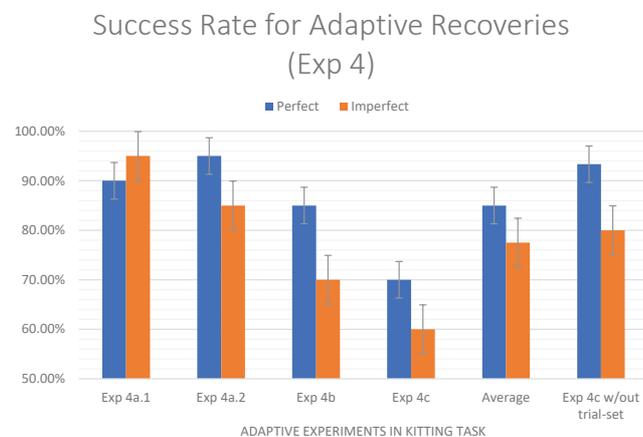}
    \caption{Success rates for Adaptive recoveries under two modalities: (i) perfect classification and (ii) imperfect classification. }
    \label{fig:exp4_success_rates}
\end{figure}

With respect to adaptive recoveries, Fig. \ref{fig:exp4_success_rates} presents success rates under our two classification modalities. 
As expected, the success rates under perfect classification generally were higher than those with imperfect classification with an average across sub-experiments of 85.0\% and 77.5\% respectively. The exception was Experiment \theExperiment a.1, where the imperfect classification modality achieved 95.0\% success rates \textit{v.s.} 90.0\% for modality (i). The failures under modality (i) were due to manipulation system errors. In one trial, during the move-to-box node, the object's collision with the packaging box moved the latter and the place action failed. Our system is limited by not actively tracking objects of interest and rationalizing relationships between them (see Sec. \ref{sec:discussion} for more comments on this).

The results also reveal that one object-set of trials in Experiment \theExperiment c had difficulties. Under perfect classification, an adaptive behavior rotated the gripped object and cause a collision with objects leading to an irrecoverable situation. For imperfect classification, there was a set of trials that led to 0 completions. Failure occurred during the adaptation to the persistent wall collision in node 3 as the system moved to the box. The culprit was the inability of the system to adapt its motion when an object with different shape attributes (height) was used compared to the one used during user demonstrations. This result points to a weakness in the system's ability to generalize adaptations when object shapes vary drastically from training as no spatial reasoning is yet embedded in the system. If each of those two trial-sets were not considered, the average success rate would be to 90.83\% and 82.50\% for perfect and imperfect classification modalities respectively.

\begin{figure}[b]
	\centering
         \includegraphics[width=\linewidth]{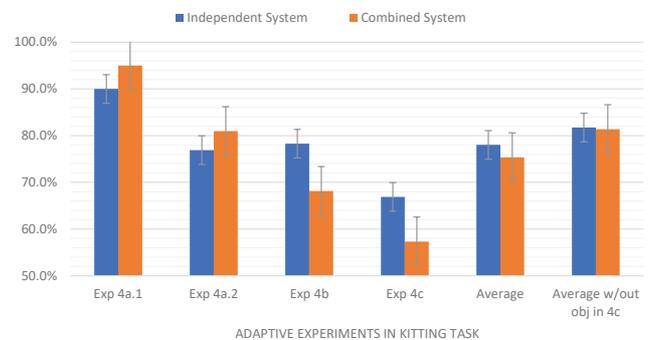}
    \caption{Overall system success rate as a function of modality for adaptive recoveries. Modality 1 considers perfect classification and modality 2 considers imperfect classification. It is surprising that some experiments with imperfect classification outperformed those with perfect classification in success rate. Wrong classifications were corrected downstream and coupled with correct recovery policies.}
    \label{fig:exp4_overall_success_rates}
\end{figure}

With respect to overall system performance, we again compare the performance between modalities. We achieved an average success rate of 78.02\% and 75.36\% for both modalities respectively. Figure \ref{fig:exp4_overall_success_rates} charts the results over sub-experiments and modalities. 

As with Exp. 3, we again see the interesting phenomena that for Experiment \theExperiment a.1,  modality (ii) achieved higher success rates than modality (i). It supports the premise that even when there are misclassifications in the system, the task can be completed as the system some time later correctly detects, classifies, and recovers from existing anomalies. 
\stepcounter{Experiment}
\subsection*{Experiment \theExperiment: Test Re-enactment and Adaptation}\label{subsec:exp_rev_adapt}
\begin{figure*}[t]
	\centering
    \subfigure{
    	\label{fig:exp5_AD}
        \includegraphics[width=0.475\linewidth]{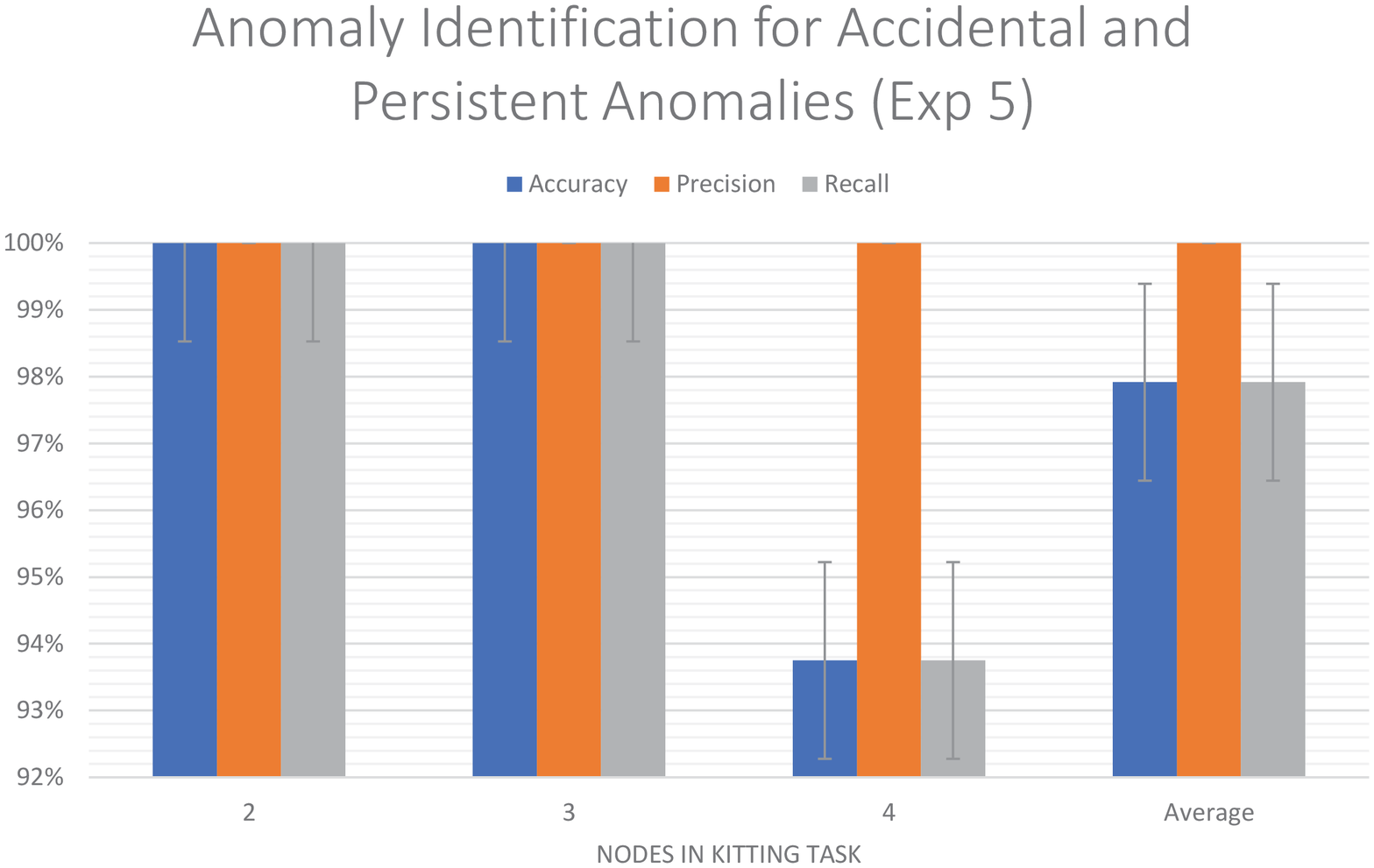}}
    \hspace{0.01cm}
    \subfigure{
    	\label{fig:exp5_AC}
        \includegraphics[width=0.475\linewidth]{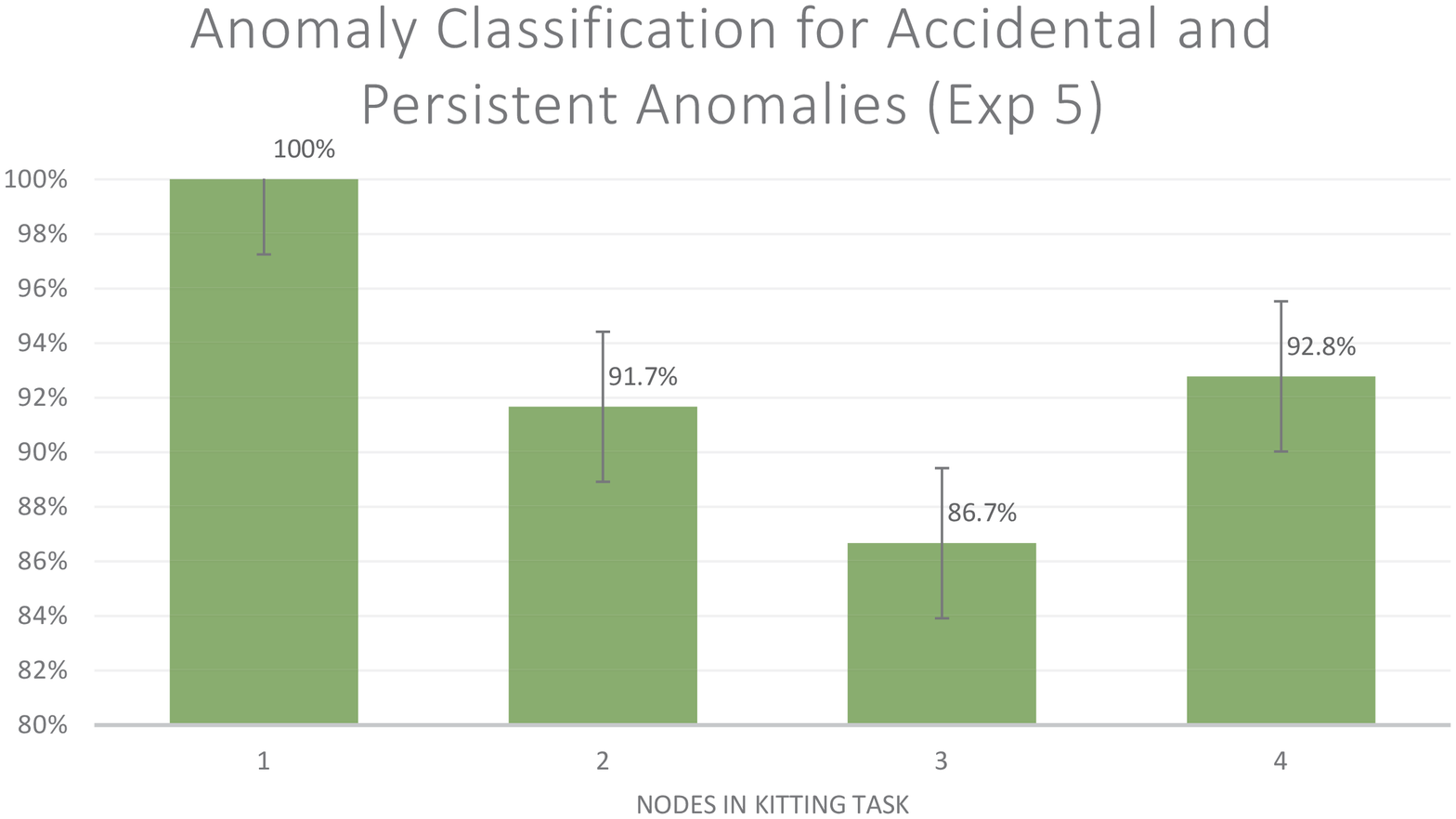}} 
    \caption{Accuracy, precision and recall metrics for the anomaly identification system and accuracy metrics for the classification system on a per-(sub)experiment basis for merged accidental and persistent anomalies (left and right respectively).} 
    \label{fig:exp5_prec_recall}
\end{figure*}

\begin{figure}[b]
	\centering
         \includegraphics[width=\linewidth]{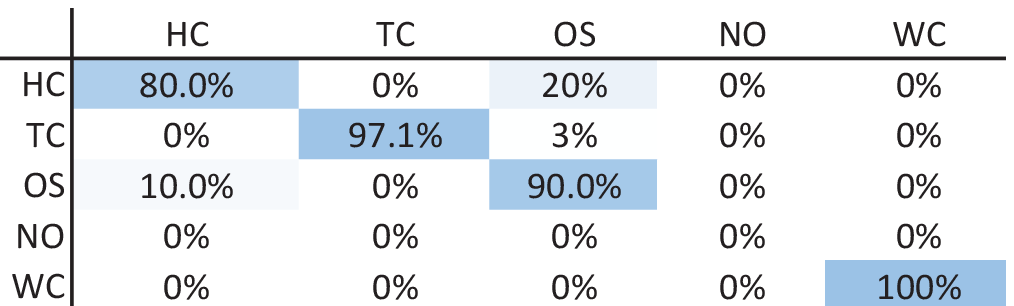}
    \caption{An anomaly classification confusion matrix for accidental and persistent anomalies HC, TC, OS, and WC in Exp. 5.}
    \label{fig:exp5_confusion}
\end{figure}

Experiment \theExperiment, analyzes the robustness of the system when re-enactment and adaptations are both integrated and present in the system. It is important to verify that re-enactment policies are not detrimental to adaptive policies and vice-versa. 
For this experiment, we integrate the accidental and persistent anomalies of experiments 3 \& 4, and similarly use the re-enactments and adaptations already learned. Anomaly identification and classification metrics are presented as before under both classification modalities. The sequence of anomalies and recovery policies present in the system are delineated in Table \ref{tbl:RE_AD_flow}, where we refer to re-enactments as ``RE'' and adaptations as ``AD''.
\begin{table}[h]
\centering
\caption{Sequence of induced accidental and persistent anomalies into the system along with triggered re-enactment (RE) and adaptive (AD) policies during the Kitting experiment.}
\label{tbl:RE_AD_flow}
\begin{tabular}{lcc}
Node                    & Anomaly Type  & Recovery Type\\
\midrule
2                       & TC       & RE       \\
3                       & HC 	   & RE       \\
3                       & WC       & AD       \\
4                       & TC       & AD       \\
\end{tabular}
\end{table}
For this experiment, 2 objects were selected at random and 10 test trials were conducted for each object. A total of 20 trials were run for each modality. Anomaly identification results across nodes can be seen in Fig. \ref{fig:exp5_AD} while anomaly classification accuracy can be seen in Fig. \ref{fig:exp5_AC}.
The anomaly confusion matrix is shown as a figure in Fig. \ref{fig:exp5_confusion}.
Corresponding success rates for modalities (i) and (ii) are summarized in Table \ref{tbl:RE_AD_success_rate}. Notation in Table \ref{tbl:RE_AD_success_rate} has been abbreviated as follows: for a TC that occurs at node 2 followed by a re-enactment, the notation we use is: @2TC-RE, hence @ indicates the node phase, followed by the two digit anomaly, followed by a dash to indicate the type of recovery. Tables \ref{tbl:RE_AD_success_rate_w_slip} and \ref{tbl:recovery_on_recovery} follow the same notation.
\begin{table}[b]
	\centering
    \caption{Success rate for combined Re-Enactment and Adaptive recoveries across 2 objects under different classification modalities. Anomaly and recovery are presented under the following notation: node location for anomaly occurrence denoted with @; followed by anomaly type, and recovery policy indicated after (-). Additionally, manipulation system errors contribution as a percentage of total failures is enclosed in parenthesis.}
	\label{tbl:RE_AD_success_rate}
      \begin{tabular}{>{\small}ll@{}}
      \toprule
       @2TC-RE @3HC-RE @3WC-AD @4TC-AD         		& Success Rate \\ 
      \toprule
      Modality (i): Perfect Classification			& 90.0 (10.0)\%\\
      Modality (ii): Imperfect Classification		& 80.0 (10.0)\% 
      \end{tabular}
\end{table}
\\ \\
\textbf{Results}\\
We now summarize the results for experiment 5. For anomaly identification a total of 72 trials were tested. We had an average accuracy and recall of 97.9\% and a perfect precision. For nodes 2 and 3 anomaly identification was done perfectly for the three metrics. It was node 4 that was more challenging with an accuracy and recall of 93.8\% and perfect precision. For anomaly classification, 71 trials were tested with an average accuracy of 92.8\%. As with anomaly identification, it was also node 4 that was the most challenging to classify followed by node 3 with an accuracy of 86.7\% and 91.7\% respectively. Note that by the time the robot reaches node 4 it has undergone 3 different anomalies and is undergoing one more and the system has also experienced two re-enactments and an adaptation. As discussed earlier, a high degree of variability in the sensory-motor signals (compared to training) begins to enter the system as more recoveries take place and change gripping poses, dynamics and inertia, and the interaction with the objects.

With regards to success rate, under classification modality (i) the success rate was 90.0\% and under modality (ii) the rate was 80\%. Fatalities occurred during the wall collision where the collision caused an object slip that displaced the object beyond the camera's field of view impeding any further attempts to re-pick. Under imperfect classification, we experienced a misclassification of HC as OS. The robot attempted to re-enact a pick. However, the object's pose was too high and no IK solutions existed. On another occasion a WC got misclassified as TC repeatedly, we aborted after 3 attempts. Specific experimental outcomes can be found as comments for this experiment in Extension 3, under the ``Exp 5'' tab in Excel.

The Wall collisions experienced in this experiment, afforded a new phenomenon. Namely, how the generation of one anomaly leads to the trigger of a subsequent anomaly. In Table \ref{tbl:RE_AD_flow}, note that an HC is induced in node 3. This same HC can trigger an OS in the task. For this reason, we further studied the system's ability to recover from a subsequent OS anomaly. As before, 10 trials were tested for the same 2 objects under both classification modalities with results shown in Table \ref{tbl:RE_AD_success_rate_w_slip}.
\begin{table}[h]
	\centering
\caption{Success rate for combined Re-Enactment and Adaptive recoveries in the presence of a subsequently generated anomaly across 2 objects under different classification modalities. Generated anomaly is denoted with($\rightarrow$). Manipulation system errors enclosed in parenthesis as a percentage of failure contribution.}
\label{tbl:RE_AD_success_rate_w_slip}
\begin{tabularx}{\linewidth}{@{}>{\small}l@{}l@{}}
    \toprule
     @2TC-RE @3HC-RE @3WC-AD$\rightarrow$OS @4TC-AD & \mbox{ }Success Rate \\ 
    \toprule
    Modality (i): Perfect Classification		& 90.0 (10.0)\% \\
    Modality (ii): Imperfect Classification		& 70.0\% \\
    \end{tabularx}
\end{table}
Under perfect classification 90.0\% success rates were also achieved. The fatality occurred when the wall collision displaced the packaging box in a way that precluded further placing of objects in the box. For imperfect classification 70.0\% success rates were achieved. In this experiment, during node 3, when an OS occurred, the system misclassified as a HC and triggered a re-enactment of the same node. Later the system triggers an NO object flag; however, because we had not previously trained a re-enactment at node 3 (only for node 2) the system halted. Experimental details can be found as comments  can also be found under Extension 3.

\stepcounter{Experiment}
\subsection*{Experiment \theExperiment: Recovering from Anomalies that Happen during Recovery}\label{subsec:exp_recovery}
The final experiment analyzes the robustness of the system in identifying and recovering from anomalies (accidental and persistent) that occur during an already executing recovery skill. It is imperative that the system performs reliably even during recovery actions. In this experiment, we test two situations: 
\begin{enumerate}[label=\roman*.,noitemsep]
\item a persistent anomaly induced during an adaptation. 
\item an accidental anomaly induced during an adaptation. 
\end{enumerate}
These two conditions will be referred to as ``Adaptation over Adaptation'' (AOA) and ``Re-enactment over Adaptation'' (ROA) respectively. Experiments are run under our two aforelisted classification modalities. Each experiment is executed for one object chosen at random and repeated 10 times. Details are shown in Table \ref{tbl:recovery_on_recovery}.
\begin{table}[h!]
	\centering
\caption{Conditions under which anomalies are induced during an adaptation recovery. Anomaly and recovery are presented under the following notation: node location for anomaly occurrence denoted with @; followed by anomaly type, and recovery policy indicated after (-). Also, ($\rightarrow$) indicates a subsequently caused anomaly. For AOA: AD1 and AD2 describe 1st and 2nd adaptations. For ROA: RE refers to re-enactment.}
\label{tbl:recovery_on_recovery}
\begin{tabularx}{\linewidth}{p{2.5in}p{0.5in}}
    \toprule
    Events								& Situation\\ 
    \toprule
    @2TC-AD1, @2TC-AD2					& AOA\\
    @3WC-AD1, @HC$\rightarrow$OS-RE		& ROA\\
    \end{tabularx}
\end{table}

\begin{figure*}[tb]
	\centering
    \subfigure{
    	\label{fig:a_a_001}
        \includegraphics[width=0.22\linewidth]{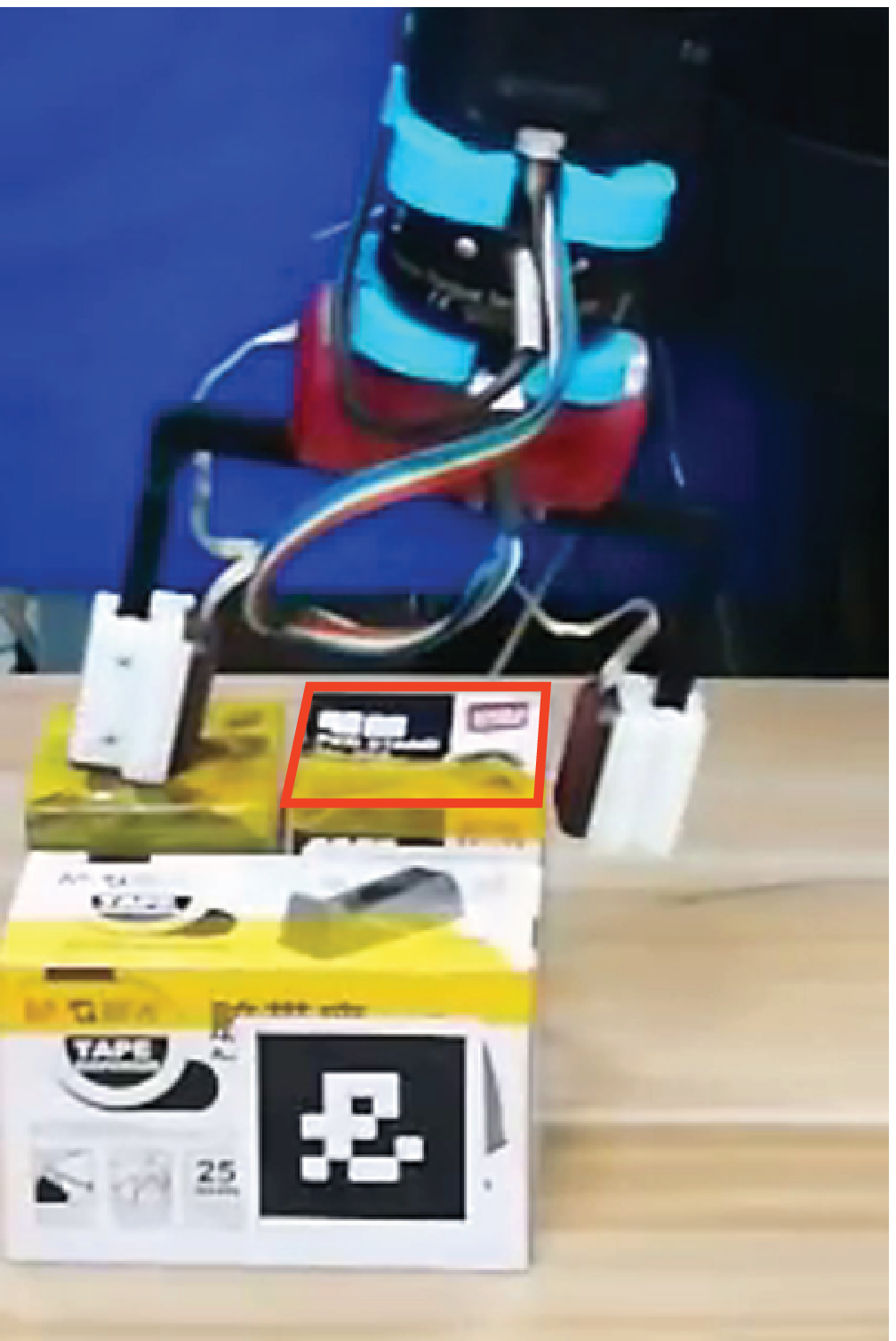}}
    \hspace{0.01cm}
    \subfigure{
    	\label{fig:a_a_002}
        \includegraphics[width=0.22\linewidth]{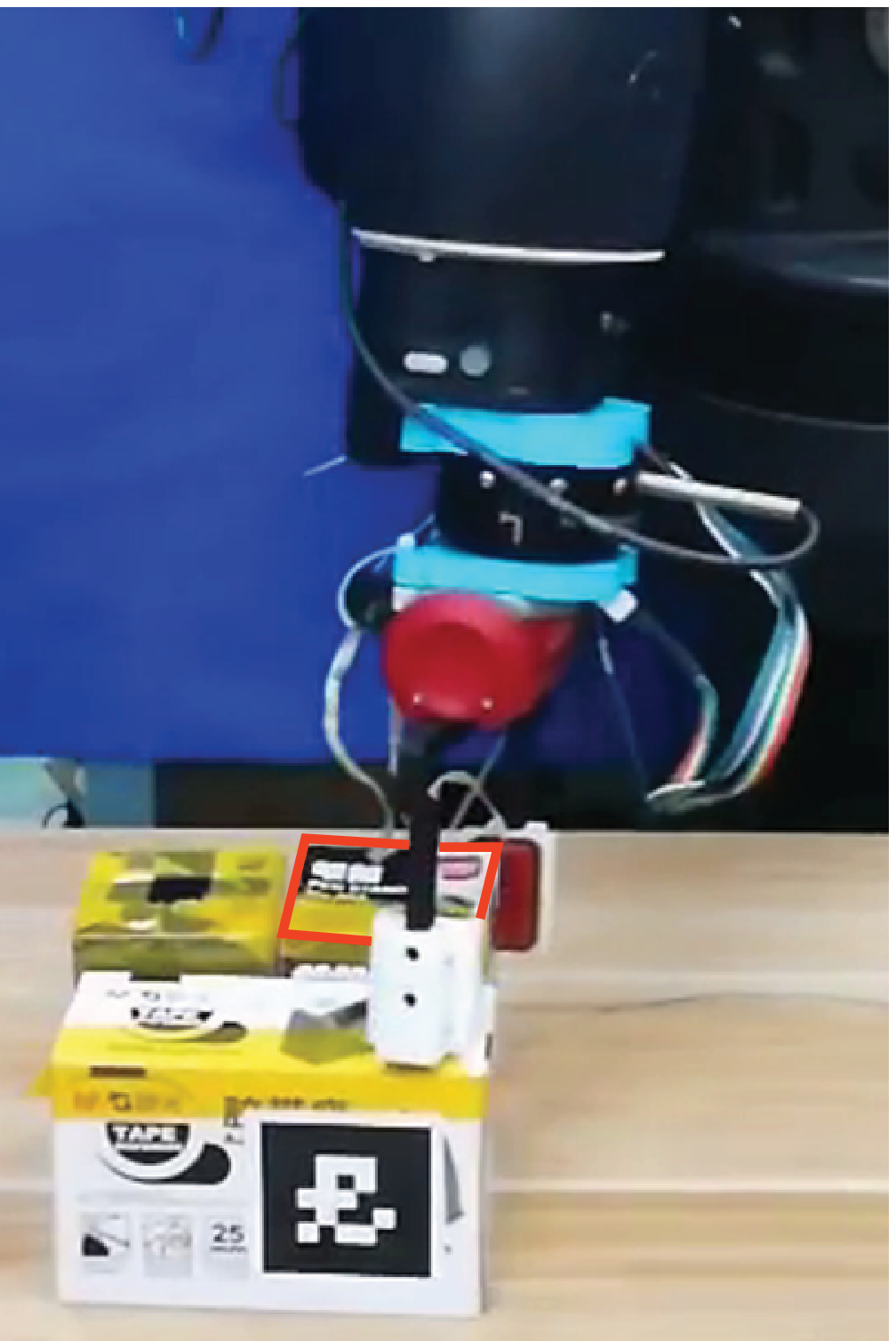}} 
        \hspace{0.01cm}
    \subfigure{
    	\label{fig:a_a_003}
        \includegraphics[width=0.22\linewidth]{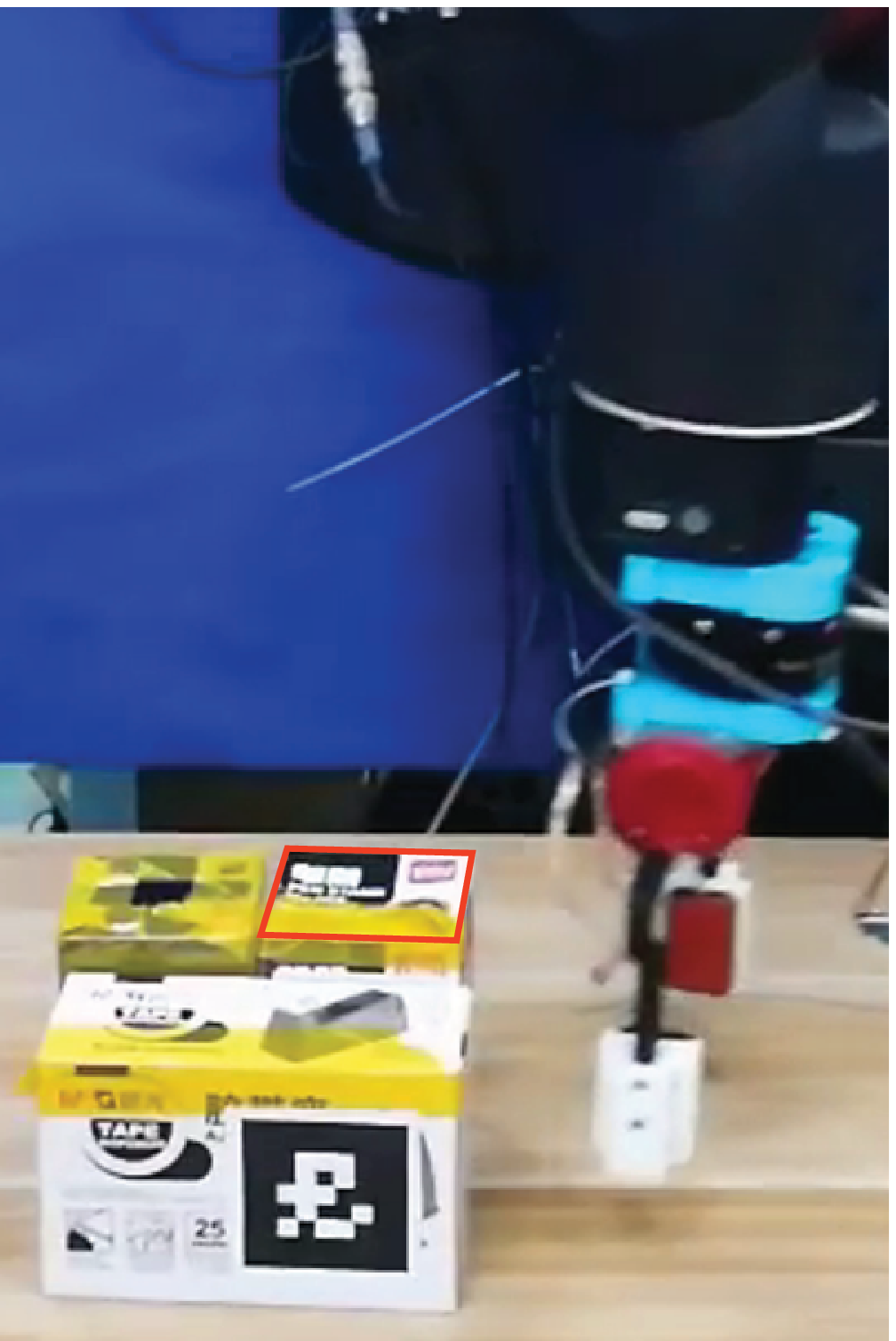}} 
        \hspace{0.01cm}
    \subfigure{
    	\label{fig:a_a_004}
        \includegraphics[width=0.22\linewidth]{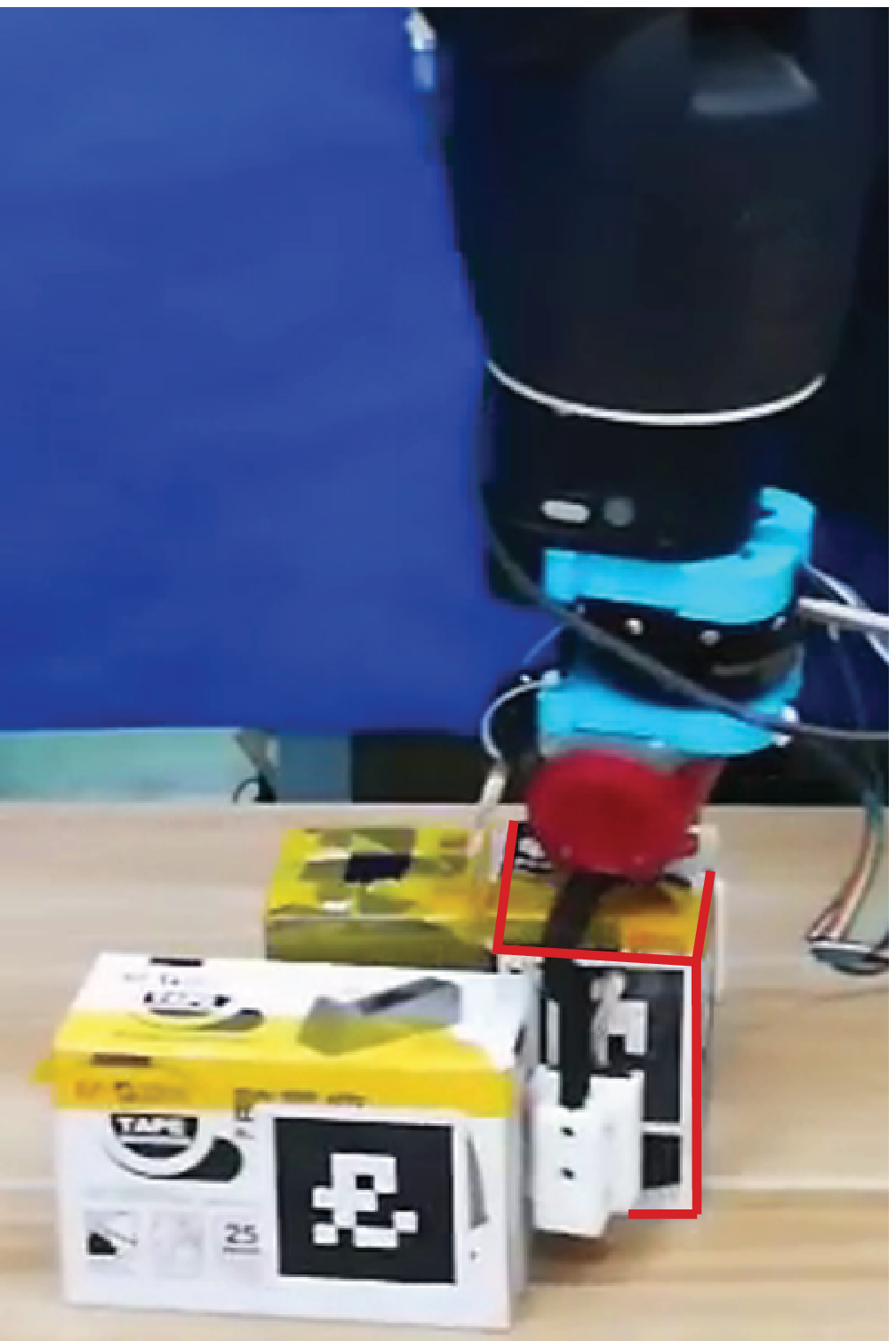}} 
    \caption{The scene shows three objects. Two smaller boxes towards the robot and one wider box away from the robot. Originally, a pick skill looks to grab the object delineated in red. However, a persistent anomaly occurs in Fig. \ref{fig:a_a_001} as one of the robot fingers collides with one of the adjacent smaller boxes. An adaptation is taught as described in Exp. 4a.1. That recovery behavior now faces a new persistent anomaly as seen in Fig. \ref{fig:a_a_002}. A new wider box was also placed nearby and now causes a new collision with the fingers. Fig's. \ref{fig:a_a_003} \& \ref{fig:a_a_004} show the implementation of a newly taught adaptation. As part of the whole process, the system is able to learn a new model of the adaptation as a nominal skill and deviations from its norm can be flagged as anomalous. The framework enables endless extensions to the graph. Given that this is a persistent anomaly, a new node is introduced to the graph. The user demonstrates a horizontal-sliding hand approach \textit{v.s.} a vertical one to resolve the new condition.}
    \label{fig:TC_AD_TC_AD}
\end{figure*}

For (i) we use the same persistent anomaly and adaptation of Exp. 4a.1. Namely, during pick, one finger collides with the placement of an adjacent object. The original adaptation rotates the robot wrist about the approach axis by $\pi/2 \mbox{ }rad$ (see Fig. \ref{fig:adapt_fingers}). In this experiment, we consider the placement of an additional object at the position where the already adapted grip fingers would descend. This in turn, would cause a new persistent tool collision. In this scenario, a new adaptation is needed. The human demonstrator decides to teach a sliding approach, whose direction of motion is parallel to the tangent of the table plane, until the fingers are centered on the object, at which point a pick behavior ensues. The adaptation is illustrated in Fig. \ref{fig:TC_AD_TC_AD} and  can also be seen in the video Extension 1. For (ii) we combine the wall collision adaptation of Exp. 4a.2 with the phenomena experienced in Exp. 5 where an HC during move-to-place causes a subsequent OS that the system recognizes and one that is resolved via a pick re-enactment. In this case, we induce a human collision that results in a subsequent slip whilst the system is resolving a wall collision through a lifting adaptation. 
\\\\ \textbf{Results}\\
For anomaly identification, a total of 20 trials were used for testing (10 and 10 for experiments AOA and ROA) and had an average accuracy of 100\% and 90.0\% for AOA and ROA respectively. Precision had the same performance and recall was perfect. 

\begin{figure}[tb]
	\centering
         \includegraphics[width=\linewidth]{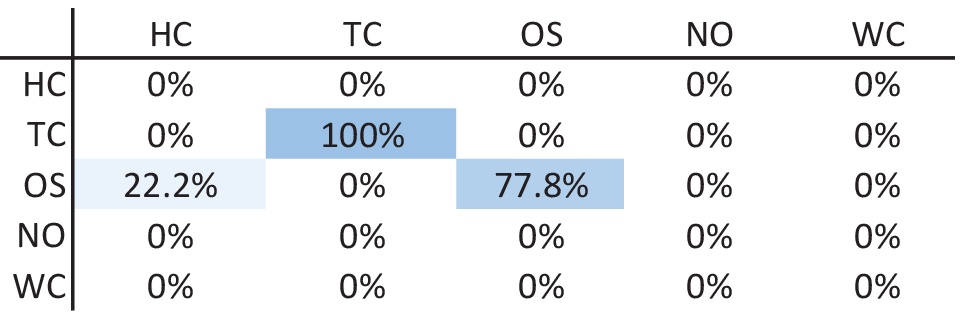}
    \caption{An anomaly classification confusion matrix for persistent anomalies TC and OS in Exp 6.}
    \label{fig:exp6_confusion}
\end{figure}

For anomaly classification, a total of 19 trials were used for testing (10 and 9 for experiments AOA and ROA) and had an average accuracy of 100\% and 77.78\% for AOA and ROA respectively.
A confusion matrix was also computed and shown as a figure in Fig. \ref{fig:exp6_confusion}. TC and WC were the target classes and the resulting HC statistics were due to miss-classification.

As for success rates, each of the two situations under both classification modalities are shown in Table \ref{tbl:Anomaly_Anomaly_success_rate}.
\begin{table}[b]
  \centering
  \caption{Success rate for anomalies and adaptations that occur during the original execution of a recovery policy. One object and two classification modalities are used to report performance metrics. Manipulation system errors enclosed in parenthesis as a percentage of failure contribution.}
  \label{tbl:Anomaly_Anomaly_success_rate}
    \begin{tabularx}{\linewidth}{p{1.25in}p{0.8in}p{0.8in}}
    \toprule
    Situation		& Perfect    		& Imperfect  			\\ 
    \toprule
    AOA				& 80.0 (20.0)\% 	& 	90.0 (10.0)\%		\\
    ROA				& 100\% 			& 	70.0\%				\\
    \midrule
    Total			& 90.0\%			& 	80.0 (10.0)\%	
    \end{tabularx}
\end{table}
For Adaptations-over-Adaptations, the system success rates were 80.0\% and 90.0\% for both classification modalities respectively. For AOA with perfect and imperfect classification, a total of three failures occurred as follows: during the 2nd adaptation attempt to grasp the block, the approach pose was inaccurate. Normally, our fingers open when a pre-pick motion has terminated. The approach trajectory had some imprecision and let to the fingers making contact with the block and tip it (instead of a sliding along the block to reach an optimal pick pose). After the tip, the block was displaced beyond the field-of-view of the camera. At this point the system continued to \textit{correctly trigger an NO flag}, however on re-enactment the pose of the object was unavailable thus holding-up the execution of the re-enactment. This could be prevent by a better implementation of the manipulation skills taught to pick the object. In retrospect, we never envisioned that training the pick in this way would be problematic. It is not clear if end-to-end training would not suffer from similar problems from inception. Clearly, the adaptations could be re-trained or improved to address the issue under any manipulation scheme. The question remains which approach would be more robust to previously unseen situations.

For Re-enactments-over-Adaptations, the system success rates were 100\% and 70.0\% for both classification modalities respectively. The latter was caused by 1 false-negative in anomaly identification, 1 false-positive in node 3, and the same system limitation previously mentioned for AOA also occurred once here.
If we look at the combined contribution of both situations for a given modality we have 90.0\% for perfect classification and 80.0\% for modality 2. 
\stepcounter{Experiment}
\subsection*{Experiment \theExperiment: Anomaly Classification Reactivity}\label{subsubsec:exps_anom_class_reactivity}
In this experiment, we analyze if anomaly classification accuracy varies as a function of the time window we use to capture multi-modal signal observations before and after the anomaly identification flag has been issued. We wish to learn the top limits in reactivity of the algorithm. That is, how quickly can we classify without sacrificing important levels of accuracy. As originally stated in Sec. \ref{subsec:overview_anomaly_collection}, we use a standard windows of $\pm$ 2 seconds to capture multi-modal signal observations before and after an anomaly has been identified. 

\begin{figure}[tb]
	\centering
         \includegraphics[width=\linewidth]{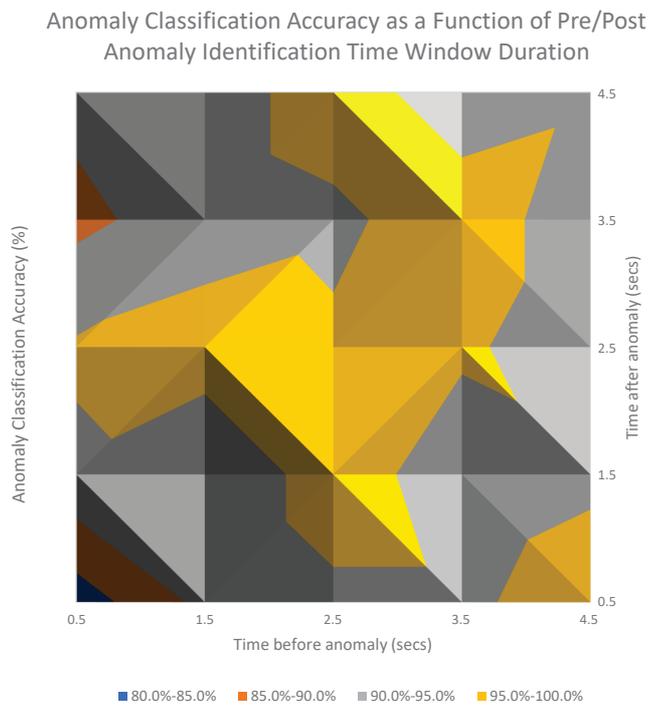}
    \caption{A contour plot of anomaly classification accuracy as a function of pre/post anomaly time window duration. The bottom axis represents the time duration for capturing signals before the anomaly trigger. The right axis represents the time duration for capturing signals after the anomaly trigger. The contour plot presents accuracy regions in groupings of 5 percentile points. The lower left corner indicates the smallest time windows, the top right corner indicates the longest time windows. Classification accuracy seems to be the highest (95\% and above) in an approximate golden central radius, with another outer ring in gray holding the next percentile accuracy grouping (90-95\%). For the smallest window, $\pm$ 0.5 secs, the classification accuracy ranges in the (80-85\%) grouping.}
    \label{fig:exp7_AC_reactivity}
\end{figure}

Fig. \ref{fig:exp7_AC_reactivity} shows a contour map of anomaly classification accuracy as a function of pre and post anomaly identification time duration. The figure contains accuracy regions in groupings of 5 percentile points, where the lower left corner indicates the smallest range of time windows, whilst the top right corner indicates the longest range time windows. The anomaly classification data in this experiment was setup in the same way as in Exp. 2. The final anomaly classification accuracy is computed as the average of the true-positive confusion matrix rates. Finally, note that reactivity measurements for anomaly identification were originally presented in \citep*{2018ROMAN-Luo-RobustVersatileEventDet} and concluded that we could identify anomalies on average consuming 1.84\% of the duration of skills. 
\\\\ \textbf{Results}\\
According to Fig. \ref{fig:exp7_AC_reactivity}, classification accuracy seems to be the highest (95\% and above) in an approximate golden central radius, with another outer ring in gray holding the next percentile accuracy grouping (90-95\%). For the smallest window combination, the lower left corner, the classification accuracy ranges in the (80-85\%) grouping. Recall from Exp. 2 that our overall anomaly classification accuracy for the standard $\pm$ 2 second window was of 96.15\%. The contour patterns seen in our experiment indicates that in general there tends to be quite similar performance in most of the studied regions. Only the region from 0.5-1.0 seconds seems to register a symmetrical drop in performance across both axis from the 90-95\% range to the 80-90\% range. Such information indicates that the main structural signatures of anomalies require slightly more than one second, given our classification algorithm in this kind of task, to provide accuracies above 90\%. 
Note that the Extension 1 video uses the standard time window capture of $\pm$ 2 seconds.
\subsection{Summary}\label{subsec:experimental_summary}
\begin{figure}[t]
	\centering
         \includegraphics[width=\linewidth]{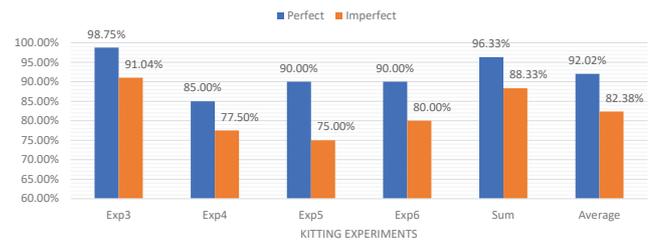}
    \caption{Recovery policy success rate across experiments along with final rate across all experiments for both classification modalities.}
    \label{fig:success_rate_across_experiments}
\end{figure}

In this last section we summarize and analyze the performance of the recovery policies. Fig \ref{fig:success_rate_across_experiments} shows the success rate across experiments along with the final percentage as a total sum across all experiments. 
When we consider classification modality (i), we can isolate the recovery critic performance. When considering counts across all experiments, our system was able to successfully recover 96.33\% of the time. This result is not the experiment's average and reflects the more heavily weighted results of Exp. 3 where we had 98.75\% success rate for re-enactments across 480 trials (across nodes, objects, and users). If we consider the average performance, we still obtain a very strong 92.02\%. This result reflects a result that we have commented on already; namely, that our work shows that as a manipulation task experiences a larger degree of recoveries, more variability enters the system rendering further introspection and classification more challenging (we recovered 85\% of the time in Exp. 4). 
Nonetheless, we still recovered on nine out of ten times across users, objects, anomaly types, and nodes in the graph, hence showing very strong performance overall. 

When we consider classification modality (ii), we are considering the entire system and the effects of not only the recovery critic, but also those of anomaly identification and anomaly classification. These results tell about the effectiveness of a highly integrated introspection and recovery system (along with a manipulation and visualization aspects of the framework). When consider all counts across experiment we recovered 88.33\% of the time and we consider the averaged result 82.38\% of the time. Hence, the integration of the complete system, diminishes the performance of the recovery system, by slightly less then 10\% points. 
Again, within comments we emphasized that the loss in performance was mainly experienced in Exp. 4 and 5 where a large number of anomalies were induced. This will often not happen in practice. Exp. 4a might be a more likely event, where 95\% recovery was achieved under imperfect conditions in our work. 
Exp. 5 contained our worst performance with successful recoveries 75\% of the time. This may not be a bad result after all. Recovering more than seven times out of 10 with unexpected scenarios, in our estimation, is not bad for current robotic performance in unstructured environments. Furthermore, in Sec. \ref{sec:discussion}, we comment in detail specific directions in which we can significantly improve and expect better results. All experimental data is contained in Extension 2, results analysis can be found in Extensions 3 and 4, and code in Extension 5. We expect the community to use the current work and results as future baselines and improve performance further. 
\section{Discussion} \label{sec:discussion}
Our comprehensive experimental results showed that our tightly-integrated, graph-based online motion-generation, introspection, and incremental recovery system worked accurately and robustly for a wide range of anomalous situations in an unstructured co-bot scenario where a human and a robot collaborated to complete kitting tasks. To the best of the author's knowledge, this is the first study where the recovery ability of a robot is examined in the presence of anomalies in manipulation in unstructured environments. In our study, we demonstrated that we could not only identify anomalies reliably (overall accuracy of 93.09\%) but also classify them in an online fashion (overall accuracy of 96.15\%). And that given simple task-level recovery policies, we could also recover consistently and reliably most of the time. The tight integration achieved in this work enabled robots to continue functioning, more than 82\% across all our anomaly scenarios, and 95\% in more typical scenarios like Exp. 4a. Even when anomalies occurred during recoveries themselves, we recovered with 80\% of effectiveness. Hence, the combination of anomaly identification, with global classification and simple but contextual task-level policies reliably showed broad robustness in being able to recover at all stages of the task, across all anomaly conditions, across different users and objects thus extending the autonomy of the system in significant ways. While the system has a number of weaknesses we will soon address, this system with simple observation capabilities of the world may serve robotics systems were sensors are limited but desire more robustness in unstructured environments. 

A couple of unexpected but welcome results are also discussed. First, the robustness results of the anomaly classification system and the recovery critic were somewhat unexpected. The sHDP-VAR-HMM model displayed a strong ability in generating good models that worked across different phases of the task and identified anomaly categories that contain important variations within. The limits of the model seemed to have shown up in Exp. 5 at node 3, when the most strenuous conditions were presented. Even there the classification system had an 86.7\% accuracy. In our hand-engineered features, we attempted to abstract structure from the data instead of only keeping raw-observations. Such that, if signal patterns that were similar occurred at dissimilar temporal positions during the observation window, they would still possess similar representations. Structure was abstracted by integrating the norm of each of the modalities in our feature set.

The second unexpected emergent result occurred when we presented results for classification modality (ii) and saw that the combined (AD/AC/REC) system at times had better performance than under modality (i) where we had perfect classification (see Exp. 3, node 3, in Fig. \ref{fig:exp3_overall_success_rate} and Exp. 4a.2). There we learned that \textit{many anomaly misclassifications did not result in unsuccessful task completions}. We learned in fact that the system could self-heal. Even when a misclassification was originally present and an inappropriate recovery policy enacted, the system self-corrected at a later time step by correctly understanding its anomalous state and later triggering the correct recovery policy. 

We believe this work has broad applicability. It's graph based structure with internal modules for motion generation and introspection, and a supervisory recovery critic, allow the system to leverage any class of motion generation algorithms including attractor-based, probabilistic, and deep end-to-end approaches (better introspection techniques can be leveraged as well). The bottom-line is that even as motion generation techniques become increasingly robust to disturbances \citep*{2016IJML-Abeel-End2EndTraining,2016ISER-Levine-LearnHandEyeCalib_BotGrasping_DL,2018ICRA-Haarnoja-ComposableDRL_BotManip}; failure is still a frequent occurrence when uncertainty in the environment surpasses the modeling ability of the system. Thus, our framework can enhance the long-term autonomy and robustness of systems that use various motion-generation approaches. 

Additionally, the deep system integration presented in the paper allowed for a comprehensive  study of the dynamics between an introspection system and an accompanying recovery-critic. We believe this is the first study of its kind, where an explicit and detailed study of the anomaly-recovery relationship is presented. We have open-sourced the code, dataset, and result analysis (see Extensions 5, 2, and 3/4 respectively) to promote and facilitate further examination of the topic. We hope others can build on our work and use the current results to further improve performance. There is still much improvement ahead and we attempt to discuss some of the main issues next. 

\subsection{Limitations, Comparisons, and Future Work}\label{subsec:discussion_limitations}
An important limitation in our work is the fact that the kitting experiment was not conducted under real factory conditions. Thus the verifiability of the work in real-world applications is unclear and further testing in real-factory conditions is necessary. The kitting experiment provides a proof-of-concept and the authors would like to extend their work to actual scenarios through corporate partners.

With regards to motion generation, we see the need for the adaptations of motion generation skills when objects are varied. While adaptations often transferred to other objects, Exp. 4c taught us that when the shape properties of an object different significantly from the object shape that was used to train motion skills, the system is susceptible to anomalies such as collision due to the lack of adaptation (end-to-end training motion generation might resolve this as it uses visual input to drive its behavior). Such adaptation is natural in humans to achieve safety \citep*{2016Nature-babivc-HumanAdapt_WholeBodyMotion}; in robotics attractor dynamics have also been used to avoid collisions \citep*{2010IROS-Haddadin-RT_MotGen_VarAttracDyn} although such dynamics have not explicitly considered object morphology in its computation. This is left as possible future work. 

Another aspect related to motion generation would be using latent state data from the anomalies to produce low-level feedback signals that could provide more immediate reactivity. The challenge of transferring high-level knowledge to useful low-level feedback still remains an open challenge. More interestingly would be the ability to recognize not just an anomaly but the onset of an anomaly and trigger feedback that rather than recovering does preventing instead. 

With regards to anomaly identification, the work of Park \et \citep*{2018AutBot-Park-MultimodAnomDet_AssistiveBots,2017IROS-Park-MultiModalAnomalyClassifFeeding,2016ICRA-Park-MultiModalMonitoringAnomalyDet_RobotManip} is the most closely related to our work. 
For Park \et, there are a couple of comparison points to be made. The first point relates to the way anomaly data is compartmentalized. Their system applied HMMs to identify anomalies for ensembles of either: a specific robot skill with a specific object, or a specific robot skill with a specific person. Such specificity makes it easier to identify anomalies but it also increases the number of classes to be trained. Evidently, models that can accurately discriminate across broader datasets (such as being trained with a multiplicity of objects or users) is desirable. In our work, our anomaly identification (and classification) was trained to identify anomalies across different task nodes, different objects, and different users (where relevant). Thus, a broader training domain was considered in our work.It is difficult to perform a direct comparison with Park. \ets work given that the the task, robot system, and environment are different. A broad comparison is only possible. In their work, they obtained an average anomaly identification accuracy across 5 tasks of 86.87\%. In our work the anomaly identification across nodes (also for 5 tasks) was 93.09\% (see Fig. \ref{fig:exp1_AD_summary_all_exps} in Exp. 1). 

With regards to anomaly classification our system seems to outperform the state of the art. The work of Park in \citep*{2017IROS-Park-MultiModalAnomalyClassifFeeding} and the work of Di Lello \et \citep*{2013IROS-DiLello-BayesianContFaultDetection} most closely resemble our work. 
In Park \ets work, their multi-perceptron classifier classified 12 common anomalies with 90\% accuracy. Furthermore, the paper also includes experiments where the robot feeds a real person with quadriplegia. In this work, they conducted anomaly identification and classification (they also classified the cause of the anomaly) and had 86\% and 90\% accuracy, resulting in a combined 88\% effectiveness for the system. So with regards to anomaly classification, we still outperformed the accuracy marker, nonetheless the number of cases they considered was larger (12 instead of 5). With regards to the combined system, our (AD/AC) overall performance was of 94.62\%, about 6\% points higher than their, but again for a smaller number of anomaly cases. 
In Di Lello \ets work, they use a simple non-parametric Bayesian model, namely the sHDP-HMM with Gaussian observations and Gibb's sampling to classify anomalies. In their work, they achieved an average classification accuracy of 87.5\% over four anomaly classes in an alignment skill with 4 obstructing objects. Our performance was between 6-8\% points higher: 96.15\% across nodes (Fig. \ref{fig:exp2_AC_summary_all_exps} in Exp. 2) and 94.4\% was the confusion matrix average in Exp. 2 (Fig. \ref{fig:exp2_AC_confusion_summary_all_exps}). Again, comparisons are difficult. Their experimentation consisted of single anomaly scenarios that did not change over time. Our scenarios included a wide range of anomalies, from one to multiple, occurring at different phases of the task with different objects and users. So, given that our anomaly experimentation was considerably more complex.

With regards to reactivity, Di Lello \et \citep*{2013IROS-DiLello-BayesianContFaultDetection} only presents a simple statement declaring that his system would have degraded anomaly classification performance if the decision had to be made before 0.65 seconds. For Park \et, they studied how fast and how well they could classify one of four anomalous signals if they changed the signal amplitude. More specifically they measured detection delay in seconds along with the true positive rate as a function of detection magnitude. They found that small amplitudes, less than 10\% could take them as much as three seconds to identify but with low true positive rates ranging less than 20\%. Signals which maintained the original amplitudes were identified in around 1 second with about 80\% accuracy. In our case, Fig. \ref{fig:exp7_AC_reactivity}, revealed that for a window of $pm$ 2 seconds, our anomaly classification (for five classes) was 96.15\%. If the window after an anomaly is triggered is brought to 1 second, our classification accuracy ranges slightly above 90\%. 

The comparison with Park \et work in \citep*{2018AutBot-Park-MultimodAnomDet_AssistiveBots} may indicate that for anomaly experiments where simulated data is not yet reliable and where real-robot (or cobot) experiments are conducted and produce a limited number of trials, then non-parametric Bayesian models with specialized variational inference algorithms models are very competitive in performing anomaly identification and classification and with very good reaction rates.

We would like to note the time and human cost that it took to gather the anomaly classification data in unstructured environments for this task. The process was arduous as manual induction was required to test anomalies. Labeling the anomalies was also problematic as the anomalies took place in a laboratory settings and may not be reflective of a true factory-floor or warehouse scenario. Automating the anomaly label collection process through simulation or a farm of robots as in \citep*{2016ISER-Levine-LearnHandEyeCalib_BotGrasping_DL}) is possible, though the algorithm by which anomaly induction takes place should be examined to understand whether it approximates real-life conditions. Another interesting possibility is the use of synthetically generated anomaly data. Synthetically generated data is becoming more common place \citep*{2014CORR-Radovanov-ComparisonBlockBootstrapMeth,2017ICDM-Forestier-GenSynthTimeSeriesAugmentSparseDatasets,2009JSS-Vinod-MaxEntropyBootstrapTimeSeries,2016W-AALTD-Le-DataAug_TimeSeries_CNN}, examples include synthetic voices, images, or depth representations. However, when it comes to anomaly data, the use of synthetic data seems more challenging as the structure of anomalous data can have important variations as discussed in this paper. It would be interesting to investigate the minimal amount of nominal data needed from which synthetic data could be generated with sufficient accuracy to properly introspect anomalies. If feasible, it would enable the learning of anomalies in an incremental fashion, similar to the way biological systems can learn from one mistake and apply the knowledge to a new scenario. Incremental learning helps classification especially when we cannot control neither the frequency or type of occurrence. It would also be desirable to continually update our models with the new experiences. More so, consider leveraging learning across similar robots that might independently face unique situations in different environments. Transfer learning of this sort has been an area of growing interest recently \citep*{2017ICRA-Devin-NNPolicies_multiTask_transfer}. Incremental learning would also open questions about how to discriminate the right level of granularity for anomaly classification labels (are all collisions with a human the same type of collision? Should such collisions be subdivided? How to determine that?) in such systems and how such discrimination would compare to that of a human. Anomaly clustering has a direct impact in recovery policies as different anomaly classifications pair-up with unique recoveries. We also foresee that as the ability to generate more faithful synthetic data becomes available, deep networks will may play more significant roles in anomaly identification and classification."
 
One more future line of research in anomaly classification that stems from this work is the ability to simultaneously identify multiple anomalies. Often times in our experimentation human collisions resulted in object slips, this raised the possibility of having two co-existing anomalies. Based on our current anomaly discrimination approach, we select the class whose likelihood is maximal. We lack an underlying structure that understands that either two anomalies are happening simultaneously or are chained to each other back-to-back. We wish to explore this as a future line of work. 

With regards to re-enactment policies on a task-planning level, the multinomial distribution is admittedly simplistic. It is an indirect process of capturing decision policies. Furthermore, while we try to reduce re-teaching by having adaptation nodes inherit re-enactment policies from their parent node; there are times anomalies will occur for the first time in later nodes for which no policy exists. This requires user intervention to train the system as happened in Exp. 5 for imperfect classification where the system halted its performance because no re-enactment policy existed for the NO class in a particular node. We are interested in looking for automated policy learning solutions that evolve over time. 

With regards to adaption policies, we do not yet model the spatial relations amongst the actors of interest; namely, the robot (end-effector), active objects (like objects to be gripped and the packaging box), and the world (support surfaces like tables and floor). These relationships provide important context for decision making and are recently attracting more attention \citep*{2018ICRA-Jund_Burgard-GenSpatReltnswEnd2EndLearn,2018ICRA-Adjali-HL_MLN_SpatialContxtDisambig,2018ICRA-Aly-TwdsUndObjOrientActions,2018ICRA-Gong-TempSpatInvSemantics_4_CommunicHum}. Without spatial relation understanding, the solutions learned in Exp. 6 will not extend to situations where the spatial relations are different from those in training. Not all experiments would fail without spatial relations context however. The HC, OS, and NO anomalies do not seem to explicitly depend on spatial context and may likely be resolved as-is in new situations. In effect, despite the lack of explicit spatial relationship modeling, our recovery policies often overcame external disturbances that might have otherwise terminated the task and endowed the system with longer operational horizons. By learning context relations, adaptations would do more than replay a learned behavior, they would in fact restore the complete and original state of the system before the anomalous condition. The larger overall challenge remains in learning how to integrate real-time reasoning and apply it to a learned skill, how to explicitly consider the spatial and functional relations between objects, the robot, and the world. It is possible that by theoretically grouping anomaly-recovery pairs into groups that do need functional-spatial reasoning and groups that do not \citep*{2013IJRR-Koppula-LearnHumActivObjAffordances_RGBD,2016TroPat_ML-Kopula_Saxena-AnticipatHumanAct_ObjAffordances}. In \citep*{2016IROS-Paulius_Sun-FOON,2018Arxiv-Jelodar-IdObjStates_Cooking}, for example FOON graphs and object affordances are introduced and might be particularly useful to resolve spatial and reasoning problems. Resolving this issue will be a consideration for future work. Notwithstanding, the work as-is with its limitations, might be useful in extending the autonomy of robots with limited sensor and/or computational capabilities. 

Finally, one last comment involves the application of our work to multi-task scenarios and human-robot interaction (HRI). To further extend long-term autonomy horizons this work should be tested not just in isolated single tasks but in longer-term multi-task scenarios that can further test the effectiveness of the proposed approaches. Additionally, it would be interesting to consider more complex graph topologies in HRI, such as a dual-graph framework that synchronizes both human and robot activity and enables mutual introspection and recovery under explicit collaboration. We plan to extend our work to include hand-over tasks from humans to the robot instead of placing objects directly in the collection bin.
\section{Conclusion} \label{sec:conclusion}
This work presented a tightly-integrated, graph-based online motion-generation, introspection, and incremental recovery system for manipulation tasks in loosely structured co-bot scenarios. Failures are and will continue to be a reality in robotics despite increasingly powerful motion-generation algorithms. Dealing with them explicitly has been the focus of this work. Recovery and introspection had robust performance. Importantly however we learned that the recovery ability of the system grows in difficulty with an increased number of adaptations as variations in sensory-motor signals increase as more recoveries are attempted. The system also showed signs of self-repair. On occasion, after an anomaly misclassification and improper recovery policy enactment, the system would correct its introspection and emit successful recovery policy that complete the task. Ultimately the system presented in this work significantly extended the autonomy and resilience of the robot and has broad applicability to all manipulation domains that suffer from uncertainties in unstructured environments: making industrial and service robots prime candidates for this technology.
\begin{acks} \label{sec:Acknowledgements}
We would like to thank Prof. Vincent Duchaine and his team from the Department of Automated Manufacturing Engineering at Quebec University for his kind support in donating the multimodal tactile sensor used in this work \citep*{2017ICRA-maslyczyk-CapacitiveSensor}. 
\end{acks}
\begin{funding} \label{sec:funding}
This work is supported by Major Project of the Guangdong Province Department for Science and Technology [grant numbers 2014B090919002, 2016B0911006], by the NSFC-Guangdong Joint Fund [grant number U1401240], by the National Science Foundation of China [grant number 61750110521], and by the Frontier and Key Technology Innovation Special Funds of Guangdong [grant numbers 2014B090919002, 2016B090911002, 2017B050506008].
\end{funding}
\theendnotes
\bibliographystyle{SageH}
\bibliography{Xbib}
\begin{appendices}
\section{Graph Structure} \label{appendix:graph}

\subsection{Nodes}
In principle, a node specifies a motion-generation model and an associated goal. Two type of nodes are specified in our system: nominal and adaptive nodes. 

\subsubsection{Nominal nodes}
Nominal nodes are implemented as ROS-SMACH states whose class definition contains member functions. In the specific case of DMPs, these are labeled as ``get\_dmp\_model'' and ``get\_pose\_goal'' for model and goal retrieval respectively. There is an additional attribute of integer type in the class definition acting as the ID of the nominal node.

\subsubsection{Adaptive Nodes}
Adaptive nodes are not implemented as a specific entity but rather as two procedures. 

The first procedure concerns when and how to create an adaptive node. Since a new type of adaptation can only be brought into our system via human demonstration, we create a new adaptive node after a human demonstration has curred. The new adaptive node simply contains a unique integer as its ID and a DMP model trained from that human demonstration.

The second procedure concerns how to determine the goal for an adaptive node. If we were to use the last frame of a human demonstration as the goal, it would result in an adaptive node having little or no generalization ability due to the fixed structure. We thus propose that the goal of an adaption is a linear transformation with respect to the previous goal of the system. This linear transformation can be retrieved by computing the transformation matrix from the previous goal to the last frame of human demonstration. This information is then saved alongside the model of the adaptive node. At runtime, we can determine the skill goal of an adaptive node by applying the saved linear transformation on the previous goal of the system. 

\subsection{Node Transitions}
\subsubsection{Transitions across Nominal Nodes}
Since nominal nodes are implemented as SMACH states, we inherit SMACH's state transition paradigm as our node transition paradigm. In the ROS-SMACH state definition, the member function named ``determine\_successor'' is called by our system to determine a nominal node's successor. 

\subsubsection{Transitions among Adaptive Nodes}
Since an adaptive node are entered only after an anomaly has occurred, we create a mapping from anomalies to their corresponding adaptive nodes. A key aspect of the mapping is a ``compound key'' composed of the ID of the node in which the anomaly happened and the anomaly type. For example, a key could be ``nominal\_node\_(4)\_anomaly\_type\_(tool\_collision)''.

After an adaptive node terminates its motion, we must consider the successor node. The system assumes that adaptive nodes, perform recovery for a nominal node that previously failed and that must arrive at the next phase or milestone of the task. In this sense, when the adaptive node terminates, it signals that a nominal state into the next phase has been attained. In this case, the originally nominal node that experienced an anomalous condition should now regain its control in determining its successor such that the original task control flow could continue as if no anomaly happened at all. 
\section{Kitting Anomaly Dataset}\label{appendix:dataset}
This section presents details of a dataset which captures sensory-motor and video data regarding the Kitting experiment under anomalous scenarios as outlined in this paper. The dataset consists of 538 rosbags. 85 of those rosbags are paired with RGB video that was captured by an external camera placed directly in front of the robot. The size of the 538 rosbags is of 37GB whilst the size of all videos is of 3.1GB. The dataset is found as Extension 2 in the paper as well as in \citep*{2018IJRR-supplement}.
\subsection{Data Description} 
The main content of our dataset is the sensory-motor recordings of the robot manipulator's experience while performing the manipulation task. Specifically for the Rethink Baxter robot, we use the following data modalities:
\begin{itemize}
  \item the right endpoint state: contains end-effector pose, twist, and a wrench defined from the joint torques (not used).
  \item the stamped wrench: obtained from a Robotiq FT 180 force-torque sensor installed on the right wrist (see Fig. \ref{fig:experimental_setup}).
  \item tactile data: obtained from a custom designed tactile sensor (see Sec. \ref{sec:Acknowledgements}).
\end{itemize}
When anomalies are triggered, we also record: (i) the time-stamp at which the anomaly is flagged as well as the anomaly classification label.
\subsection{Recording methodology}\label{sec:howisthedatarecorded}
All sensory-motor signals exist as ROS topics in our system and as such recorded as ROS bags offline. When an anomaly is identified, we signal this event by sending a timestamped ROS message to a pre-defined topic that is also recorded as a rosbag. Anomaly classification labels are recorded in a text file in a line-by-line basis. 

Mapping from data modalities to ROS topics is as follows:
\begin{itemize}
\item 
	\begin{description}
		\item[Baxter right endpoint state] \hfill \\ /robot/limb/right/endpoint\_state
    \end{description}
\item
	\begin{description}
		\item[Robotiq force sensor FT 180] \hfill \\ /robotiq\_force\_torque\_wrench
	\end{description}
\item
	\begin{description}
		\item[Robotiq tactile sensor] \hfill \\ /TactileSensor4/Accelerometer, \hfill \\ /TactileSensor4/Dynamic, \hfill \\ /TactileSensor4/EulerAngle, \hfill \\ /TactileSensor4/Gyroscope, \hfill \\ /TactileSensor4/Magnetometer, \hfill \\ /TactileSensor4/StaticData
	\end{description}
\end{itemize}
\subsection{Data Organization}
The dataset is composed of folders that use the format: ''experiment\_at\_[time]''. Each folder represents a test trial in the kitting experiment. Within a given folder, there will be a rosbag ''record.bag'' and a text file ''anomaly\_labels.txt''. Each of these contain the rosbag topics mentioned in Sec. \ref{sec:howisthedatarecorded} and the recorded labels for the given experiment.
\subsection{Anomaly Data Extraction}
To extract anomaly data, one should first focus on the topic ''/anomaly\_detection\_signal'' whose messages are effectively timestamps indicating when anomalies were identified. It's worth noting that a burst of anomaly timestamps might have been published to this topic for one anomaly. Therefore timestamps that are adjacent in time should be ignored. We recommend ignoring a timestamp if its distance to its precursor is less than 1 second. After anomaly timestamps are extracted, labels in the accompanied ''anomaly\_labels.txt'' can be paired accordingly. 

We have tried to clear the dataset of any corrupted trials. However, if the number of anomaly timestamps does not equal to the number of labels, that experiment should be discarded. 
\section{Notation Table}\label{app:notation_table}
\begin{table}[h]
  \centering
  \caption{Summary of graph and DMP notation.}
  \label{tbl:notation_table1}
    \begin{tabularx}{\linewidth}{p{0.35in}p{3.5in}}
      \toprule
      Notation					& \qquad\quad Description 	\\ 
      \bottomrule 				&							\\
      \textbf{\textit{Graph}}	& 							\\
      \midrule
      \GC						& Graph for a given task	\\
      \BC						& Behavior in a given task 	\\
      \NC						& Behaviors are represented by nodes in the graph\\
      \TC						& Transitions in a graph 	\\
      $\TC_{s,t}$               & Node transitions from $\NC_s$ to $\NC_t$\\
      $\mathcal{N}_{ij}$		& 1st branch level node		\\
      $\mathcal{N}_{ijk}$		& 2nd branch level node		\\
      \SC						& Skill generation modules	\\
      \VC						& Visual goal processing modules \\
      \MC						& Introspection modules 	\\
      \FC						& A given anomaly 			\\
      \RC						& A recovery action 		\\
      $\RC_R$					& A re-enactment recovery type \\
      $\RC_A$					& An adaptive recovery type	\\
      \\
\end{tabularx}
\end{table}
\begin{table}[h]
\begin{tabularx}{\linewidth}{p{0.35in}p{3.5in}}
      \textbf{\textit{DMPs}}	&									\\
      \midrule														\\
      $K$                       & Spring constant of PD control		\\
      $D$                       & Damping constant of PD control	\\
      $x, g, v$                 & Position, position goal, \& velocity \\
      $s$                       & Spatial scaling constant			\\
      $\tau$                    & Temporal scaling constant			\\
      $\alpha$					& Arbitrary scaling term			\\
      $f(\cdot)$                & The forcing term					\\
      $\psi(\cdot)$             & The basis function				\\
      $\omega$					& Weighting of basis functions		\\
      \\
  \end{tabularx}
\end{table}
\begin{table}[h]
    \begin{tabularx}{\linewidth}{p{0.35in}p{3.5in}}
      \textbf{\textit{HMM}}	&												\\
      \midrule																\\
      $z_t$                     & Latent state at time $t$ 					\\
      $\XC_n$                   & The \textit{nth} training example sequence \\
 	  $x_t$                     & Observation at time instant $t$ 			\\
      $b(z_t)$                  & Mode specific emission distribution 		\\
      $\boldsymbol{\theta}$     & Set of dynamic parameters of state $k$ 	\\
      $\pi_0$					& Initial mode distribution 				\\
      $\pi_{jk}$                & Transition probability from state $k$ to $j$ \\
      $\Pi$ 					& A given HMM model 						\\
      \\
\end{tabularx}
\end{table}
\begin{table}[h]
\begin{tabularx}{\linewidth}{p{0.4in}p{3.45in}}
      \multicolumn{2}{l}{\textbf{\textit{sHDP-HMM}}}						\\
      \midrule
      $G_0$						& Base probability measure	 				\\
      $G_j$						& HMM transition probability measure		\\
      
      $\alpha,\gamma$           & DP concentration parameters				\\
      $H$						& Continuous base distribution				\\
      $\beta_k$					& Weights used to compute $G_0$				\\
      $\kappa$                  & The sticky parameter of transition distribution \\
      $GEM(\gamma)$				& Distribution to define stick-breaking process \\
      \\
\end{tabularx}      
\end{table}
\begin{table}[t!]
\begin{tabularx}{\linewidth}{p{0.5in}p{3.35in}}
 	\textbf{\textit{VAR}}		&												\\
    \midrule
	$e_t$						& Additive white noise at time $t$ and mode $z_t$ \\
    $\boldsymbol{\Sigma}$		& White noise covariance matrix for mode $z_t$ 	\\    
    $\boldsymbol{A}$			& Time-invariant regression matrix at $z_t$ 	\\
    $\Delta$, $\nu$				& Covariance $\Delta$ \& degrees of freedom $\nu$ in $\mathcal{IW}$ \\
    $\mathcal{IW}(\nu,\Delta)$  & Inverse wishart 								\\
    $\boldsymbol{K}$			& Covariance across matrix columns				\\
	\\
    \multicolumn{2}{l}{\textbf{\textit{Anomaly Identification}}}			\\
	\midrule																
    $\nabla L$                      & Natural log of HMM filtered belief state \\
    \end{tabularx}
    \vspace*{8in} 
\end{table}
\end{appendices}
\end{document}